\documentclass{article} % For LaTeX2e
\usepackage[final]{colm2026_conference}

\usepackage{microtype}
\usepackage{hyperref}
\usepackage{url}
\usepackage{booktabs}
\usepackage{paralist}
\usepackage{graphicx}
\usepackage{subcaption}
\usepackage[most]{tcolorbox}

\definecolor{promptbg}{RGB}{245,248,250}   % very light blue-gray
\definecolor{promptframe}{RGB}{200,210,220} % soft gray-blue frame

\newtcolorbox{prompttemplate}[2][]{
  title={Template~#2},
  colback=promptbg,
  colframe=promptframe,
  coltitle=black,
  fonttitle=\bfseries,
  breakable,
  boxrule=0.6pt,
  arc=2mm,
  left=4pt,
  right=4pt,
  top=4pt,
  bottom=4pt,
  #1
}
\newif\ifshowannotations
\showannotationstrue % or \showannotationsfalse
\showannotationsfalse
\ifshowannotations
  \newcommand{\authorNote}[4]{\noindent\textbf{\textcolor{#1}{[$^{#2}_{#3}$: \textit{#4}]}}}
\else
  \newcommand{\authorNote}[4]{}
\fi

\usepackage{booktabs}
\usepackage{multirow}
\usepackage[table,dvipsnames]{xcolor}
\usepackage[normalem]{ulem}
\usepackage{wrapfig}
\useunder{\uline}{\ul}{}

\usepackage{lineno}

\definecolor{darkblue}{rgb}{0, 0, 0.5}
\hypersetup{colorlinks=true, citecolor=darkblue, linkcolor=darkblue, urlcolor=darkblue}

\usepackage[noabbrev,capitalize]{cleveref} % http://tug.ctan.org/tex-archive/macros/latex/contrib/cleveref/cleveref.pdf

\crefname{page}{page}{pages}
\crefname{footnote}{footnote}{footnotes}   % "footnote" is lowercased, overriding capitalize option
\crefname{equation}{equation}{equations}   % "equation" is lowercased, overriding capitalize option; note that \labelcref drops this word if you want to say something like "the divergence (3)"

\creflabelformat{equation}{#2\textup{#1}#3}

\crefname{line}{line}{lines}               % "line" is lowercased, overriding capitalize option
\crefrangeformat{line}{lines #3#1#4--#5#2#6}
\crefname{lstlsting}{Listing}{Listings}   
\crefname{section}{\S}{\S\S}
\Crefname{section}{\S}{\S\S}    % must define start-of-sentence version explicitly since \S isn't a letter
\crefformat{section}{#2\S#1#3}  % remove space between section symbol and the number, and include \S in the hyperlink
\Crefformat{section}{#2\S#1#3}
\crefrangeformat{section}{\S\S#3#1#4--#5#2#6}
\Crefrangeformat{section}{\S\S#3#1#4--#5#2#6}
\crefmultiformat{section}{\S#2#1#3}{ and~\S#2#1#3}{, \S#2#1#3}{ and~\S#2#1#3}
\Crefmultiformat{section}{\S#2#1#3}{ and~\S#2#1#3}{, \S#2#1#3}{ and~\S#2#1#3}
\crefrangemultiformat{section}{\S\S#3#1#4--#5#2#6}{ and~\S\S#3#1#4--#5#2#6}{, \S\S#3#1#4--#5#2#6}{ and~\S\S#3#1#4--#5#2#6}
\Crefrangemultiformat{section}{\S\S#3#1#4--#5#2#6}{ and~\S\S#3#1#4--#5#2#6}{, \S\S#3#1#4--#5#2#6}{ and~\S\S#3#1#4--#5#2#6}
\def\eqref#1{equation~\ref{#1}}
\usepackage{xspace}

\newcommand{\calibrate}{tune\xspace}
\newcommand{\calibrated}{tuned\xspace}
\newcommand{\calibration}{tuning\xspace}
\newcommand{\calibrating}{tuning\xspace}
\newcommand{\patchscopes}{Tokens2Words\xspace}
\newcommand{\swpatchscopes}{FragMend\xspace}
\newcommand{\embedding}{embedding\xspace}
\newcommand{\embeddings}{embeddings\xspace}
\newcommand{\Embedding}{Embedding\xspace}
\newcommand{\Embeddings}{Embeddings\xspace}
\usepackage{tcolorbox}
\usepackage{float}

\tcbuselibrary{breakable, skins}

\newtcolorbox[auto counter, number within=section, 
              list inside=prompt, list type=figure]{promptbox}[2][]{
  enhanced,
  breakable,
  colback=black!3!white,
  colframe=black!60!black,
  fonttitle=\bfseries,
  title={Prompt Template~\thetcbcounter: #2},
  label={prompt:\thetcbcounter},
  #1
}
\title{Defragmenting Language Models: \\An Interpretability-based Approach for Vocabulary Expansion}

\author{Maitrey Mehta$^1$, Nishant Subramani$^2$, Zhichao Xu$^1$, Ashim Gupta$^1$\& Vivek Srikumar$^1$ \\
$^1$Kahlert School of Computing, University of Utah\\
$^2$Language Technologies Institute, Carnegie Mellon University \\
\texttt{\{maitrey,svivek\}@cs.utah.edu}
}
\begin{document}
\ifcolmsubmission
\linenumbers
\fi

\maketitle
\begin{abstract}
All languages are equal; when it comes to tokenization, some are more equal than others. 
Tokens are the hidden currency that dictates the cost and latency of contemporary LLM access. 
However, many languages written in non-Latin scripts observe a poor exchange rate: LLMs take several multiples of tokens to encode the same information in many languages as they do for English. 
We observe that this issue, known as \emph{token over-fragmentation}, persists in modern open-weight LLMs. 
The standard remedy is vocabulary expansion that adds target language items missing from the model's vocabulary.
We focus on two core decisions in the vocabulary expansion process: \textit{What items should we add?} and \textit{How should we initialize their corresponding input and output embeddings?} We comprehensively study and advance the new research direction of interpretability-based vocabulary expansion along these two core decisions, making four key contributions: 
First, we question the conventional use of frequency-based methods to choose candidate vocabulary items to add~(a decision long treated as settled), and show that interpretability-based methods offer a superior performance-token efficiency tradeoff. 
Second, we strengthen the case for interpretability-based embedding initialization by showing large gains~($\sim$20 pts) over baseline initialization methods for several languages written in non-Latin scripts. 
Third, we identify the phenomenon of ``\textit{subword detokenization}," where models progressively merge fragmented subword tokens into larger subwords across layers. 
Fourth, we propose \swpatchscopes, a method that leverages this phenomenon to push the efficiency ceiling of interpretability-based expansion, and validate its effectiveness. 
\end{abstract}

\section{Introduction}
\label{sec:intro}
\begin{figure*}[h]
    \begin{minipage}[]{0.49\textwidth}
        \centering
        \includegraphics[width=\linewidth]{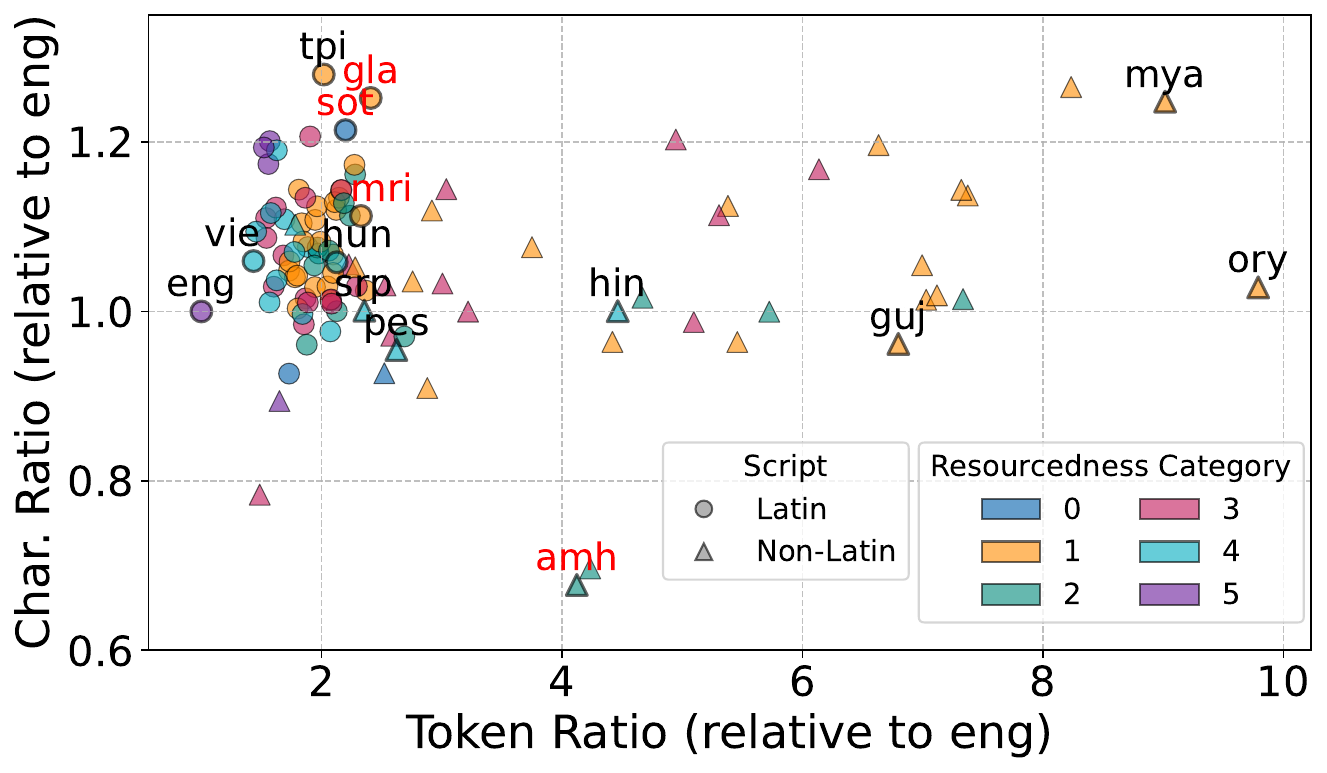}
        \subcaption{}
        \label{fig:token_ratio_qwen3_30b}
    
        \includegraphics[width=\linewidth]{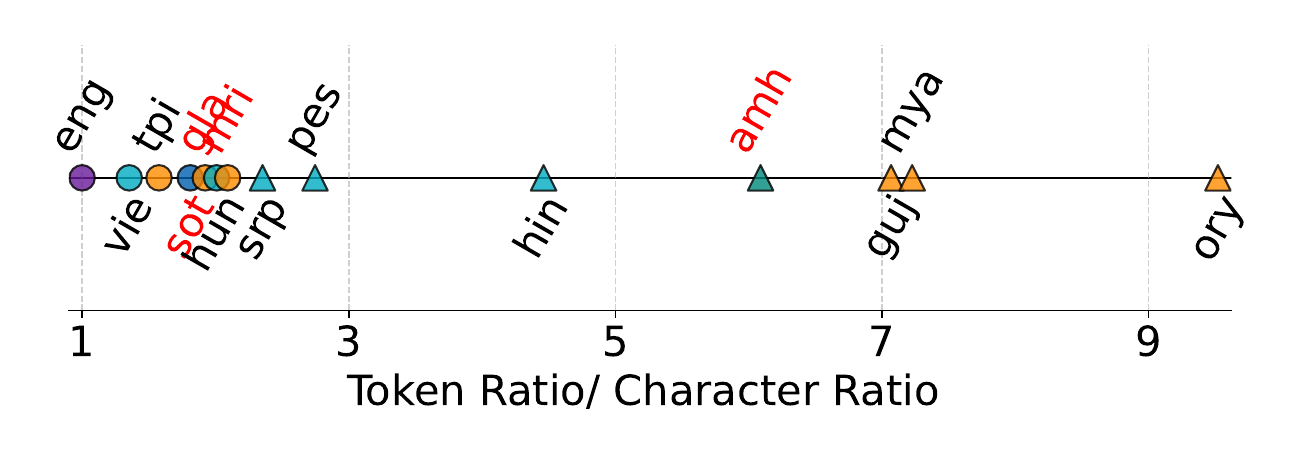}
        \subcaption{}
        \label{fig:x_by_y_tokratio2charac}
    \end{minipage}
    \hfill
    \begin{minipage}[]{0.5\textwidth}
        \centering
        \includegraphics[height=\textheight, keepaspectratio, width=\linewidth]{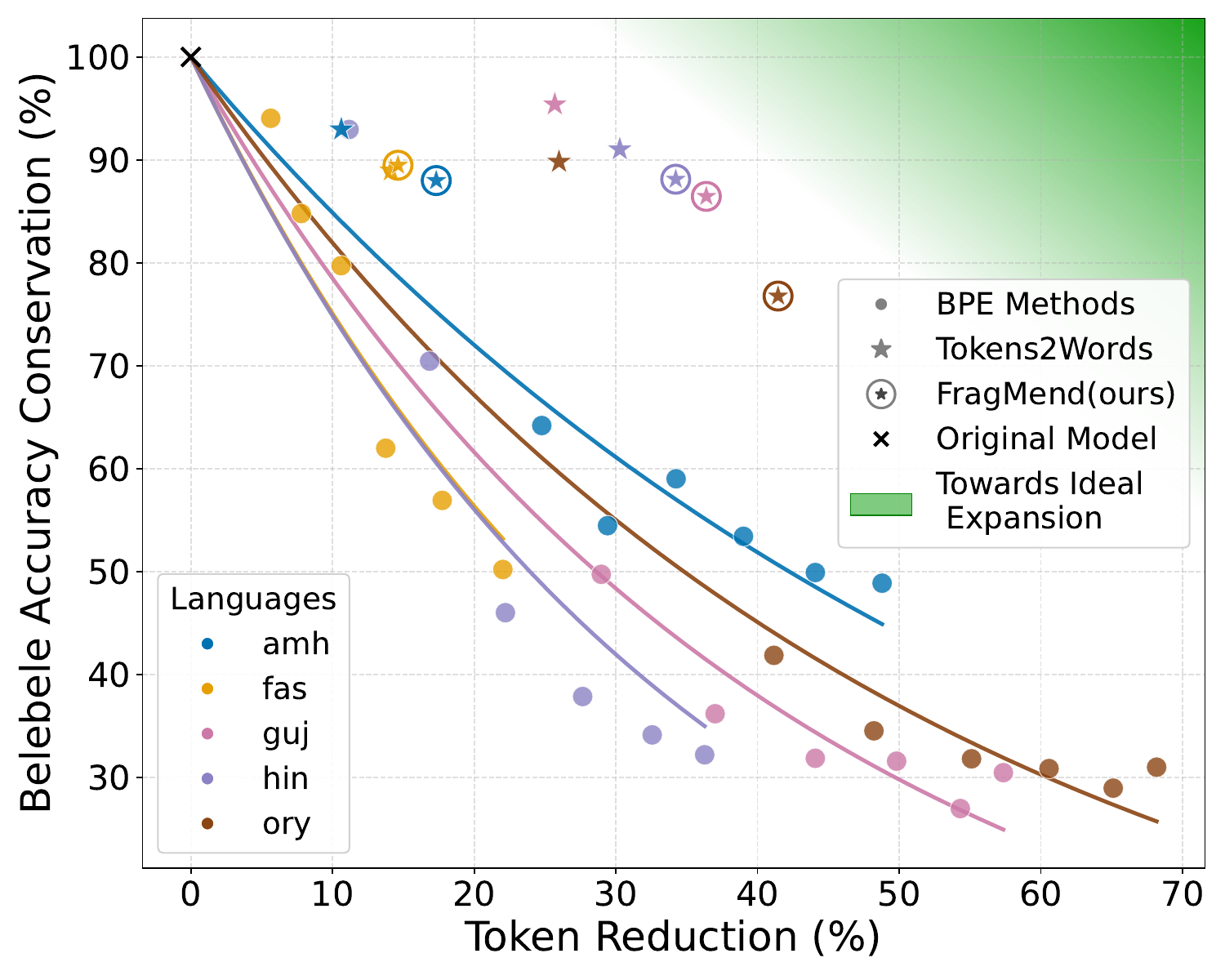}
        \subcaption{}
        \label{fig:pareto_qwen3_30b}
    \end{minipage}
    \caption{(a) Token over-fragmentation across 111 languages for Qwen3, measured using the ratio of tokens needed to encode the same information~(SIB200;~\citealp{adelani-etal-2024-sib}) in a language to that in English~(\textit{x-axis}). Several languages, predominantly low-resource~(as indicated by a lower \textit{resourcedness category}) ones that use a non-Latin script, require several multiples of tokens despite using similar numbers of UTF-8 characters (\cref{sec:motivation}). Languages in red are not supported by the model. (b) Token ratio scales disproportionately with character ratio. (c) Interpretability-based methods, like \patchscopes and \swpatchscopes, push the Pareto front achieved with the conventional frequency-based vocabulary expansion methods like BPE to counter over-fragmentation~(\cref{sec:challenging_bpe}). Points towards higher token reduction and accuracy conservation~(\textit{top right} area) are more desirable.}
    
\end{figure*}

Subword-based tokenization is a prevalent strategy in modern Large Language Models~(LLMs) that relies on a fixed-size vocabulary to encode model inputs and outputs. However, data-driven approaches to selecting vocabulary items, like Byte-Pair Encoding~\citep[BPE;][]{sennrich-etal-2016-neural} and Unigram~\citep{kudo-2018-subword},
often introduce unwanted disparities across languages. In particular, languages under-represented in the pre-training data used for vocabulary construction are known to require several multiples of tokens as other languages to encode the same information~(as shown in~\cref{fig:token_ratio_qwen3_30b}) --- a phenomenon known as \emph{token over-fragmentation}~\citep{rust-etal-2021-good, ahia-etal-2023-languages, petrov2023language}.

Addressing token over-fragmentation is increasingly pressing, as it directly affects inference speed and API costs for the speakers of impacted languages. 
Since tokenizer design is largely fixed before model training~\citep{abagyan2025tokenizerruleallemergent}, re-training the model from scratch with an equitable tokenizer is computationally infeasible. Prior work instead explores `\textit{vocabulary expansion}', a practical solution of expanding the model vocabulary with new target language items, retrofitting their \embeddings\footnote{We use `\textit{\embeddings}' to collectively denote both input and output~(or, LM head) embeddings.}, and \calibrating the model to align with the new vocabulary items~\citep[etc.]{wang-etal-2020-extending,gee-etal-2022-fast,dobler-de-melo-2023-focus}.

Vocabulary expansion thus reduces to answering three core questions~\citep{yamaguchi2026howcan}: \begin{inparaenum} [a)]
    \item \textit{Which vocabulary items should be added to the tokenizer?}  (\texttt{Stage 1}: Item Selection),
    \item \textit{How should the \embeddings for new vocabulary items be initialized?} (\texttt{Stage 2}: \Embedding Initialization), and
    \item \textit{Which model parameters should be \calibrated, and how?} (\texttt{Stage 3}: \Embedding Tuning).
\end{inparaenum}  

In this work, we focus on the first two questions, treating the comparatively well-studied \calibration step~\citep{kim2024efficienteffectivevocabularyexpansion,yamaguchi2026howcan} as secondary, and adopting a fixed \calibration strategy across all experiments. 

Surprisingly, the choice of \emph{which} items to add has received far less attention than how to initialize them. Prior work~\citep[\textit{inter alia}]{cui2023efficient, yamaguchi2026howcan} uses frequency heuristics such as BPE to determine the vocabulary items to be added~\citep{gage1994anewalgo}, a design choice we carefully scrutinize in this work. 
Consequently, the related \embeddings are initialized by simple heuristics~\citep{gee-etal-2022-fast, dobler-de-melo-2023-focus} or hyper-networks~\citep{minixhofer2024zero,ozeren-etal-2025-hyperofa}. More recently, interpretability-based vocabulary expansion, which uses model activations~\citep{kaplan2025from, dobler2026token} to inform both item selection and \embedding initialization, has emerged as a new direction, but remains largely under-explored in the multilingual setting we consider here.

In this work, we address this gap by focusing on \patchscopes~\citep{kaplan2025from} as a representative interpretability-based method. We analyze its approach to item selection and embedding initialization, and compare it with conventional methods mentioned above in the multilingual context. 
We find that \patchscopes has an efficiency ceiling that limits its token efficiency\footnote{Token efficiency indicates the number of tokens required to encode a piece of information.
Fewer tokens imply higher efficiency. More such definitions for the reader's aid are in App~\ref{sec:definitions}.} benefits. We propose \swpatchscopes, an extension that raises this efficiency ceiling by operationalizing our observation of `\textit{subword detokenization}.` With this work, we advocate for interpretability-based vocabulary expansion approaches that require only a small unlabeled corpus, no model retraining, and yet yield meaningful tokenization improvements for communities that bear a large cost premium relative to English speakers. 

In summary, our main contributions follow a four-part cascading structure: 
\begin{inparaenum}[(1)]
    \item We show that interpretability-based vocabulary expansion techniques outperform conventional frequency-based heuristics when selecting items to be added~(\texttt{Stage 1}). Crucially, these techniques push the token efficiency-performance Pareto front set by frequency-based methods, indicating a better tradeoff~(\cref{fig:pareto_qwen3_30b}). 
    \item We present a comprehensive analysis of \patchscopes-based \embedding initialization~(\texttt{Stage 2}). We show that this method outperforms many heuristic-based initialization strategies across several languages, including low-resource languages written in non-Latin scripts that suffer from severe token over-fragmentation. Together, these two points provide strong evidence for the merit of interpretability-based vocabulary expansion. However, we find that \patchscopes possesses a low ceiling on efficiency gains. 
    \item We demonstrate the presence of `\textit{subword detokenization}', where models reconstruct whole words through the progressive merging of subwords, forming many intermediate subwords. Our observation suggests that detokenization generalizes effectively beyond the full-word setting proposed by \citet{kaplan2025from}, hinting that LLMs maintain a richer internal lexicon than their vocabulary implies. 
    \item  We operationalize `\textit{subword detokenization},' proposing a new vocabulary expansion technique called \swpatchscopes. Experiments indicate that \swpatchscopes provides substantial token efficiency gains with minor performance tradeoffs relative to \patchscopes for many non-Latin-script languages.\footnote{Code available at \url{https://github.com/utahnlp/fragmend}}  
\end{inparaenum}

\section{Token Over-fragmentation: The Disparity}
\label{sec:motivation}

\textit{Token over-fragmentation}~(or, \textit{excessive fragmentation}) is the phenomenon observed in LLMs where certain languages, especially ones with non-Latin writing systems, are encoded in a disproportionately large number of subword tokens~\citep{ahia-etal-2023-languages, petrov2023language}. 

\textbf{Analysis.} In \cref{fig:token_ratio_qwen3_30b}, we capture token over-fragmentation across 111 languages using the Qwen3-30B-A3B~\citep{qwen3technicalreport} tokenizer on the parallel SIB200 dataset~\citep{adelani-etal-2024-sib}. 
For each language, we compute the ratio of subword tokens~(and characters\footnote{We use character counts as a tokenizer-agnostic proxy to contextualize token usage across scripts.}) relative to their English counterparts. Values greater than 1 indicate that a language requires more tokens~(or characters) than English to encode the same content~\citep{kanjirangat-etal-2025-tokenization}. 
An equitable tokenizer would place all points for all languages on or close to the diagonal $y=x$. In other words, we would expect the token ratio to increase proportionally with text length. Instead, we see that all the points lie below the diagonal~($x/y>1$, \cref{fig:x_by_y_tokratio2charac}), indicating that tokenizers use disproportionately more tokens than necessary.

Over-fragmentation is most acute in languages that use a non-Latin script~(shown as triangles in Figures \ref{fig:token_ratio_qwen3_30b} and \ref{fig:x_by_y_tokratio2charac}) like Odia (\texttt{ory}), which requires about ten times as many subword tokens as English despite having similar character counts. 
Analyzing under the language resource lens via \citet{joshi-etal-2020-state}'s taxonomy, lower resourced languages~(categories 0--2) have higher fragmentation; though script is a stronger predictor. For instance, a high-resource non-Latin-script language like Hindi~(\texttt{hin}), still requires over four times the number of tokens as English, far worse than many lower-resourced Latin-script languages. Perhaps most surprisingly, fragmentation is far more severe for some languages \textit{officially supported} by the model~(e.g., Odia, Burmese~(\texttt{mya})) than for some unsupported languages~(e.g., M\={a}ori~(\texttt{mri}) and Amharic~(\texttt{amh}), marked in red).

\textbf{Why does it happen?} Tokenizers developed using frequency-based methods like BPE~\citep{sennrich-etal-2016-neural} and Unigram~\citep{kudo-2018-subword} inherit the biases of their training data. Training datasets tend to be heavily skewed towards higher-resourced languages written in the Latin script~\citep{xue-etal-2021-mt5}. Also, Latin remains by far the most prevalent script, serving over 500 languages~\citep{mayhew2016howmany}, and is only expanding as the popularity of romanization in non-Latin-script languages increases~\citep{sengupta2024social}. As a result, tokenizer vocabularies tend to be dominated by Latin subwords, unless explicitly rebalanced. Moreover, byte fallback in standard subword tokenizers encodes poorly represented scripts as byte strings, further fragmenting individual characters: a single Chinese character uses three bytes versus one byte for a Latin character~\citep{limisiewicz-etal-2024-myte}.

\textbf{Impact on Real-World Users.} 
Input and output tokens form the interface a user has with a model, so an over-fragmented language can lead to increased API costs~\citep{ahia-etal-2023-languages, churchill2026reducing}, higher latency~\citep{sun-etal-2023-multi}, and poor downstream performance~\citep{ali-etal-2024-tokenizer,kanjirangat-etal-2025-tokenization} for people from that language community. Tokenization may seem like a technical footnote, but it is a silent tax for many.
\section{Background: Interpretability and Vocabulary Expansion}
\label{sec:background}
In this section, we summarize \patchscopes~\citep{kaplan2025from}, one of the first interpretability-based vocabulary expansion methods, and its use for item selection~(\texttt{Stage 1}) and \embedding initialization~(\texttt{Stage 2}). In this work, we use \patchscopes as the representative interpretability-based method to show the promise of this research direction for the two stages of vocabulary expansion. \patchscopes builds on the phenomenon of word-level `\textit{detokenization}', which posits that models maintain an inner lexicon far larger than their vocabulary and learn to reconstruct meaningful representations for concepts that span multiple subword tokens, echoing prior findings~\citep[e.g., ][]{gurnee2023finding}. 

\subsection{\texttt{Stage 1}: Selecting Vocabulary Items to Add }
\label{sec:background_addition}
\citet{kaplan2025from} employ the Patchscopes framework~\citep{ghandeharioun2024patchscopes} to detect word-level \textit{detokenization}. For every unique pre-tokenized word $w\notin V$ and $w\in C_{train}$, where $V$ is the model vocabulary and  $C_{train}$ is some training corpus, Patchscopes is applied using a candidate LLM~(parameterized by $\theta$) to determine if $w$ is successfully detokenized and, consequently, is a part of the model's inner lexicon. Suppose $w$ is tokenized into a set of subword tokens $\{t_1,t_2,\ldots,t_n\}$ using the model tokenizer $T$. 
In the first forward  pass, the Patchscopes framework computes the hidden activations~($h_{t_i}^{l}$) for all subwords, $t_i\in T(w)$, at every layer $l$. 
For the second set of independent forward passes, each activation $h_{t_i}^{l}$ is patched in place of the input embedding of a placeholder in a repetition-seeking prompt template, $P$\footnote{Example prompt template: ``In English: $<$PH$>$, $<$PH$>$, $<$PH$>$'', where the placeholder token, $<$PH$>\in V$.}: $y_{t_i}^{l} = LLM_{\theta}(f(P,h_{t_i}^l))$, where $f$ is the activation patching function that replaces the input embeddings of all placeholder tokens with the representation $h_{t_i}^{l}$. 
The word $w$ is considered successfully detokenized~($S_w=\top$) if $w$ appears as a prefix in any output~($y_{t_i}^{l}$) across all subwords and layers: 
$S_w = \bigvee_{i=1}^{n} \bigvee_{l=1}^{L} Match(y_{t_i}^{l}, w)$, 
where $L$ is the number of layers in the model. In their vocabulary expansion experiments, \citet{kaplan2025from} only restrict to the activations corresponding to the last token~($i=n$). All successfully detokenized words are then added to the model vocabulary, forming the expanded vocabulary, $\overline{V} := V \cup \{w \in C_{train} \mid S_w = \top\}$.\footnote{\textbf{Caveat:} The number of addable items is bounded by the number of unique words in $C_{train}$, limiting tokenizer efficiency---a limitation our proposed method~(\cref{sec:our_method}) directly addresses.}

\subsection{\texttt{Stage 2}: Initializing \Embeddings for the Added Items}
\label{sec:background_initialization}
The input and output embedding matrices are expanded to accommodate the extended vocabulary. 
\Embeddings for original vocabulary items remain unchanged. For newly added items, $w^* \in \overline{V}-V$, \embeddings are initialized by multiplying learned layer-dependent mappings $T_{l,E}$ and $T_{l,U}$ with the output hidden states of the last subword token of $w^*$ at the earliest layer where detokenization was successful ($l_{min}$).
$T_{l,E}$ and $T_{l,U}$ are learned by fitting an orthogonal Procrustes transformation~\citep{Schonemann_1966} from the hidden activations at layer $l$ to the corresponding \embeddings of the original vocabulary items. 
New \embeddings are then fine-tuned using language-adaptive pre-training~\citep[LAPT;][]{chau-etal-2020-parsing}.
\section{Subword Detokenization and \swpatchscopes}
\label{sec:our_method}
\patchscopes limits the candidate items to the unique full words occurring in the training corpus~(\cref{sec:background}), which presents two limitations. 
First, this constraint imposes a low ceiling on the number of words that can be added. Second, the candidate set may be dominated by heavy-tailed words that contribute little to reducing token over-fragmentation, if added to the expanded vocabulary. 
To address both limitations, we explore the possibility that models possess a latent subword vocabulary, a phenomenon we call ``\textit{subword detokenization}''. We introduce \swpatchscopes that leverages this phenomenon to add subword tokens as vocabulary items, increasing exposure to items that may never occur as words in the training corpus. 

In \cref{sec:background_addition}, we describe the success conditions for the detokenization of full words. We ask: \textit{Can we successfully detokenize subwords that do not occur in the model vocabulary?} We define candidate subwords as all unique token affixes~(prefixes, infixes, suffixes) extracted from every unique word in $C_{train}$. 
We then apply the procedure from \cref{sec:background_addition}~(first and second forward passes) for every unique $w$ in $C_{train}$ and extend the detokenization success condition to the subword setting, i.e., for an affix $t_{i:j}$, it is defined as: $S_{t_{i:j}} = \bigvee_{l=1}^{L} Match(y_{t_j}^{l}, t_{i:j})$.

\begin{figure*}[t]
    \centering
    \begin{subfigure}{0.44\textwidth}
        \centering
        \includegraphics[trim={1cm 0cm 0cm 0cm}, width=\linewidth, height=5.5cm]{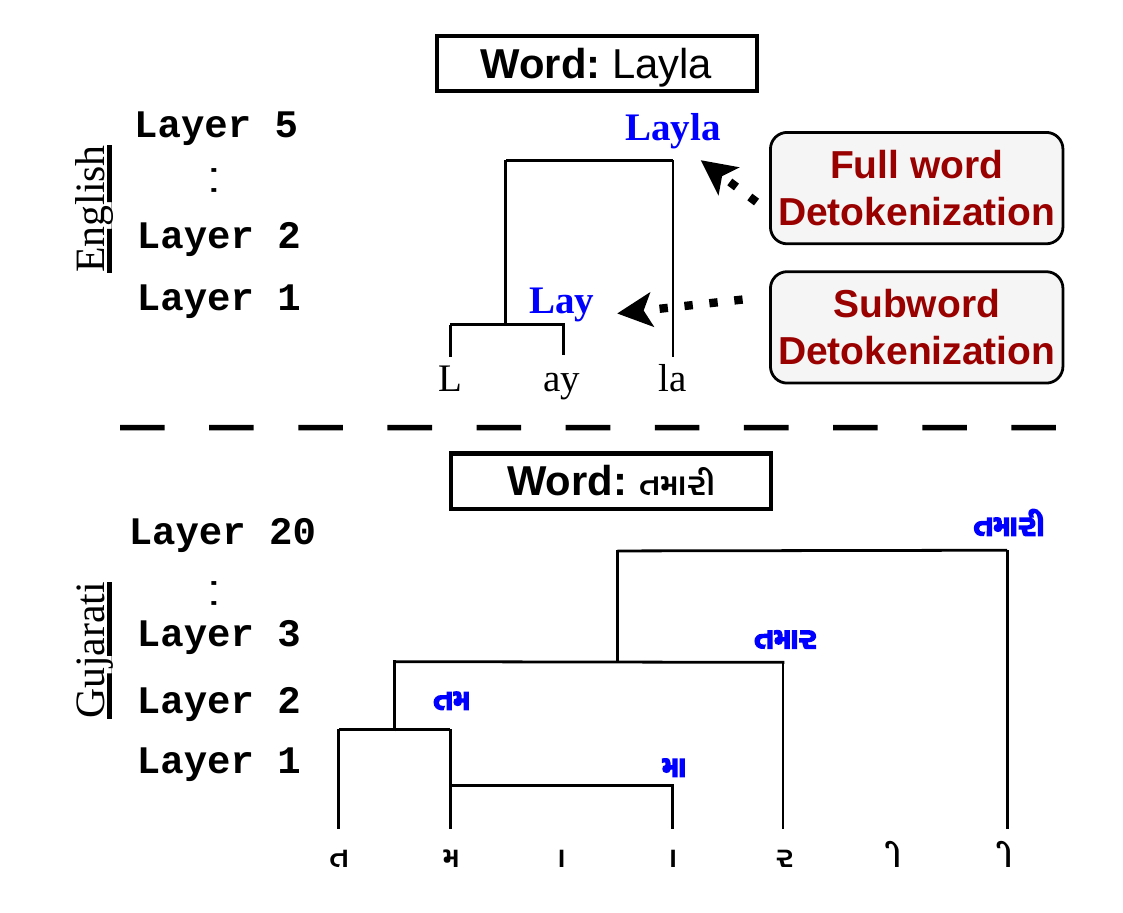}
        \caption{}
        %\phantomcaption{}
        \label{fig:sw_detokenization}
    \end{subfigure}%
    ~ 
    \begin{subfigure}{0.54\textwidth}
        \centering
        \includegraphics[width=\linewidth]{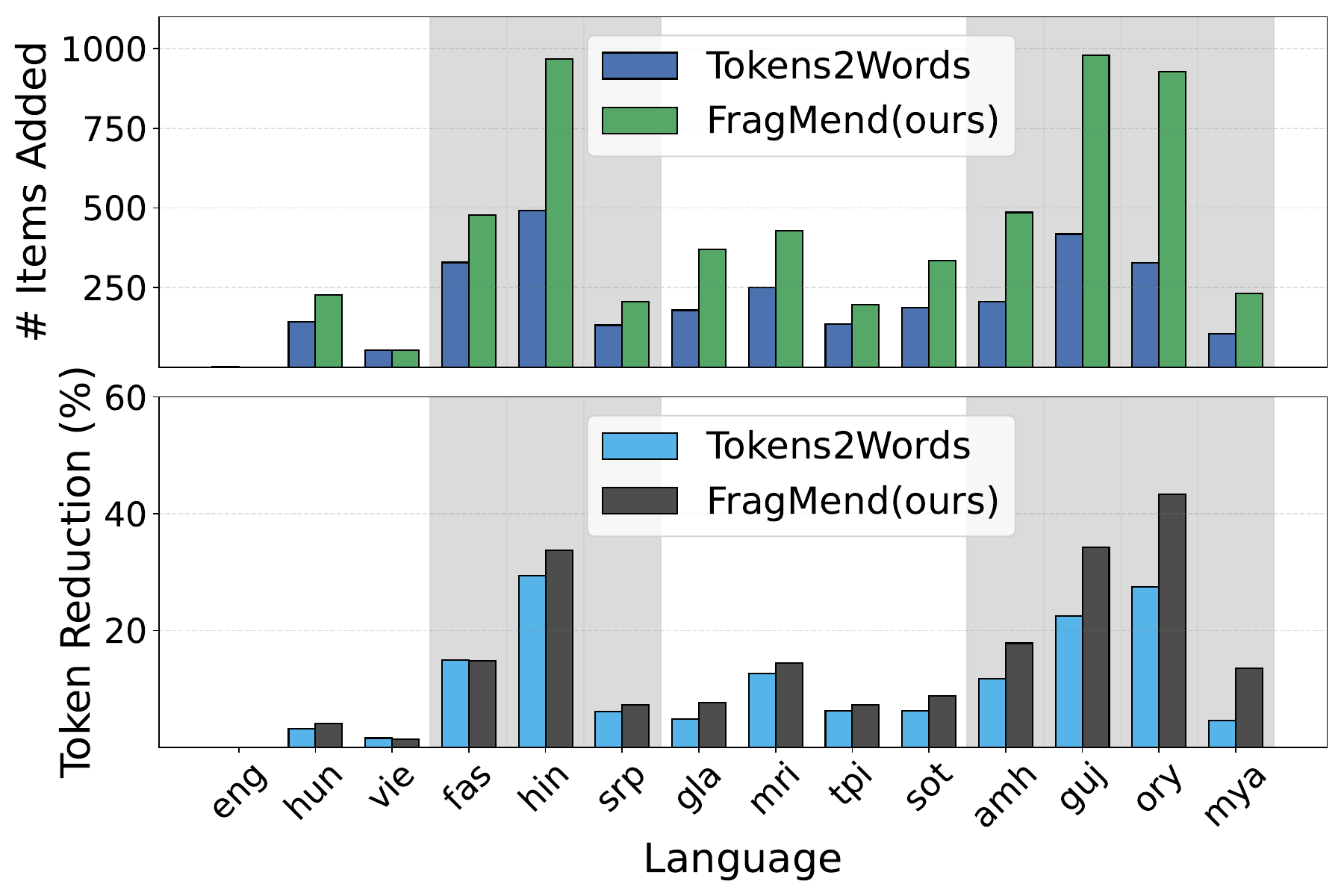}
        \caption{}
        %\phantomcaption{}
        \label{fig:sw_vspsc_efficiency_analysis}
    \end{subfigure}
    \caption{(\textit{a}) Subword detokenization typically occurs as progressive merges of prefix tokens across several languages. In the example, the word `Layla' is tokenized as `\texttt{L}'+`\texttt{ay}'+`\texttt{la}' by the original tokenizer. We observe that the subword `Lay' is detokenized in the first layer of the model, at the token corresponding to `\texttt{ay}' before the full word is detokenized at the fifth layer, on the last token. (\textit{b}) With a small 1000 sequence training corpus~($C_{train}$), we observe a large jump in added vocabulary items~(\textit{top}), and a much better efficiency ceiling as computed on the parallel SIB200 benchmark~\citep{adelani-etal-2024-sib}~(\textit{bottom}). For all plots in the paper, languages with a gray background are written in a non-Latin script.}
\end{figure*}

We find that LLMs tend to detokenize subword-level prefixes progressively before they detokenize the full word. As seen in \cref{fig:sw_detokenization}, we observe this phenomenon across several languages, ranging from high-resource languages like English to low-resource languages like Gujarati. This finding suggests that the model's internal lexicon is not binary (i.e., word or not word), but is compositional, with subwords serving as intermediate representational units on the path to full-word reconstruction. Beyond its use for vocabulary expansion in this work, our observations contribute to the literature on how LLMs organize lexical knowledge in a structured way~\citep{elhage2021mathematical,gurnee2023finding,marks2025sparse}.

Subwords detokenized successfully are used to form the expanded vocabulary, $\overline{V}$, with \embeddings initialized using the same learned mappings $T_{l,E}$ and $T_{l,U}$ as in \patchscopes. However, unlike full words, subword tokens are not necessarily unique to a single word, and a subword may occur as an affix in multiple words. For instance, the subword `\texttt{the}' may occur as a candidate subword in the words `\textit{they}' and `\textit{either}', yielding multiple distinct hidden activations, and consequently, multiple candidate initializations. We therefore define a `\textit{reduction}' strategy that reduces these candidate initializations into a single set of input and output \embeddings for a subword. We present experiments with different reduction strategies in \cref{sec:swpsc_design_choices}. The resulting \embeddings are then \calibrated via LAPT, as in \patchscopes.  

\section{Experimental Setup}
\label{sec:experimental_setup}

This section contains experimental details~(more in \cref{sec:exp_choices_design_decisions}). We divide this section into two parts, first explaining the baseline BPE-based vocabulary expansion method~(\texttt{Stage 1}) and different baseline \embedding initializations~(\texttt{Stage 2}), and then other experimental details.

\subsection{Baseline Methods for Vocabulary Selection and Embedding Initialization}
\textbf{BPE-based Item Selection.} BPE builds a tokenizer vocabulary by iteratively merging the most frequent consecutive token pairs, starting from characters/bytes, until the desired size is reached. Learned merges are ordered, with earlier merges taking encoding priority. In our experiments, we extend the pre-trained tokenizer's merges by continuing this process on the training data, initializing new \embeddings with one of the baseline methods below.

\textbf{Random Initialization.} \Embeddings for a newly added item $w^*$ are initialized by sampling from a normal distribution: $e_{w^{*}} \sim \mathcal{N}(\mu_E, \sigma_E^2)$ and $u_{w^{*}} \sim \mathcal{N}(\mu_U, \sigma_U^2)$, where $\mu_E$ and $\mu_U$ are means of the input and output embedding matrices, and $\sigma_E$ and $\sigma_U$ are their standard deviations.

\textbf{FVT Initialization~\citep{gee-etal-2022-fast}.} For a word $w^*$ tokenized into subwords $\{t_1,t_2,\ldots,t_n\}$, per the original tokenizer, \embeddings are initialized by averaging the \embeddings of the constituent subwords: $e_{w^{*}}= \Sigma_{i=1}^n e_{t_{i}}/n$  and $u_{w^{*}}= \Sigma_{i=1}^n u_{t_{i}}/ n$.

\textbf{FOCUS Initialization~\citep{dobler-de-melo-2023-focus}.} FOCUS learns a sparse weight vector~($\alpha$) over the original vocabulary items using a semantic auxiliary embedding space~(i.e., fastText~\citep{joulin-etal-2017-bag}) for every $w^*$: $e_{w^{*}} = \sum_{i=1}^{|V|} \alpha_{i,w^*} \, e_{t_i}$ and 
$u_{w^{*}} = \sum_{i=1}^{|V|} \alpha_{i,w^*} \, u_{t_i}$, 
where $\sum_{i=1}^{|V|} \alpha_{i,w^*} = 1$.

\subsection{Other Experimental Setup Components}

\textbf{Models.} We consider Qwen3-30B-A3B~\citep{qwen3technicalreport} as our primary model for analysis. We also consider small but recent TinyAya-Global~\citep{salamanca2026tiny} and Qwen3.5-4B~\citep{qwen3.5}. Qwen3.5-\{0.8B,2B,4B,9B\} are used for model-scale analysis~(\cref{sec:model_scale_analysis_sec}).

\textbf{Languages.} 
We experiment with fourteen languages, selected to span a balanced grid of script type (Latin/non-Latin) and resourcedness (high/low) according to \citet{joshi-etal-2020-state}'s taxonomy, with additional consideration of geographic and typological diversity~(details in \cref{sec:lang_choice}):
English (\texttt{eng}), Hungarian (\texttt{hun}), Vietnamese (\texttt{vie}), Persian (\texttt{pes/fas}), Hindi (\texttt{hin}), Serbian (\texttt{srp}), Scottish Gaelic (\texttt{gla}), M\={a}ori (\texttt{mri}), Tok Pisin (\texttt{tpi}), Southern Sotho (\texttt{sot}), Amharic (\texttt{amh}), Gujarati (\texttt{guj}), Odia (\texttt{ory}), and Burmese (\texttt{mya}). Vocabulary expansion is performed independently for every language. For each experiment, we aim to expand the tokenizer for a single target language.

\textbf{Corpora.} We source training and \calibration corpora for all fourteen languages from the Glot-500c corpus~\citep{imanigooghari-etal-2023-glot500}, considering two sizes of training data $|C_{train}|\in\{1000,10000\}$ sequences. The training data determines the candidate vocabulary items. For \calibration, we first sample an independent set $C_{\calibration}$ commensurate with the size of the training data of the experimental setting. The training and the \calibration~set in combination~($C_{train} \cup C_{\calibration}$) are used for LAPT, where we only \calibrate the \embeddings~(\cref{sec:other_exp_choices}). Additionally, we define a validation and test split for benchmarking containing 1000 sequences each.

\textbf{Evaluation Benchmark. }We evaluate our experiments using a topic classification benchmark---SIB200~\citep{adelani-etal-2024-sib}---using F1 score, and a reading comprehension task---Belebele~\citep{bandarkar-etal-2024-belebele}---using accuracy. Further, we report the bits-per-byte~\citep[BPB;][]{choe2019bridging, pagnoni-etal-2025-byte} performance on the Glot-500c test set. 

\section{Results and Analyses}
We summarize this section around the four-part cascading structure mentioned in \S\ref{sec:intro}. First, we challenge the use of frequency-based heuristics as the default selection strategy for choosing vocabulary items to add. We show that interpretability-based methods offer a better performance-efficiency~(\cref{sec:definitions}) tradeoff compared to BPE. Using this result, we present a comprehensive comparison of several \embedding initialization strategies for vocabulary items added using an interpretability-based approach. Our primary focus for these two sets of results will be on \patchscopes, to help establish the benefit of interpretability-based vocabulary expansion, an approach that has not been studied in multilingual literature. Having shown the promise of the interpretability approach, we demonstrate that our method, \swpatchscopes, breaks \patchscopes' efficiency ceiling with modest performance tradeoffs. Finally, we provide a detailed explanation of the design choices and analysis for \swpatchscopes.

\begin{figure*}[]
    \centering
    \begin{subfigure}{0.48\textwidth}
        \centering
        \includegraphics[width=\linewidth]{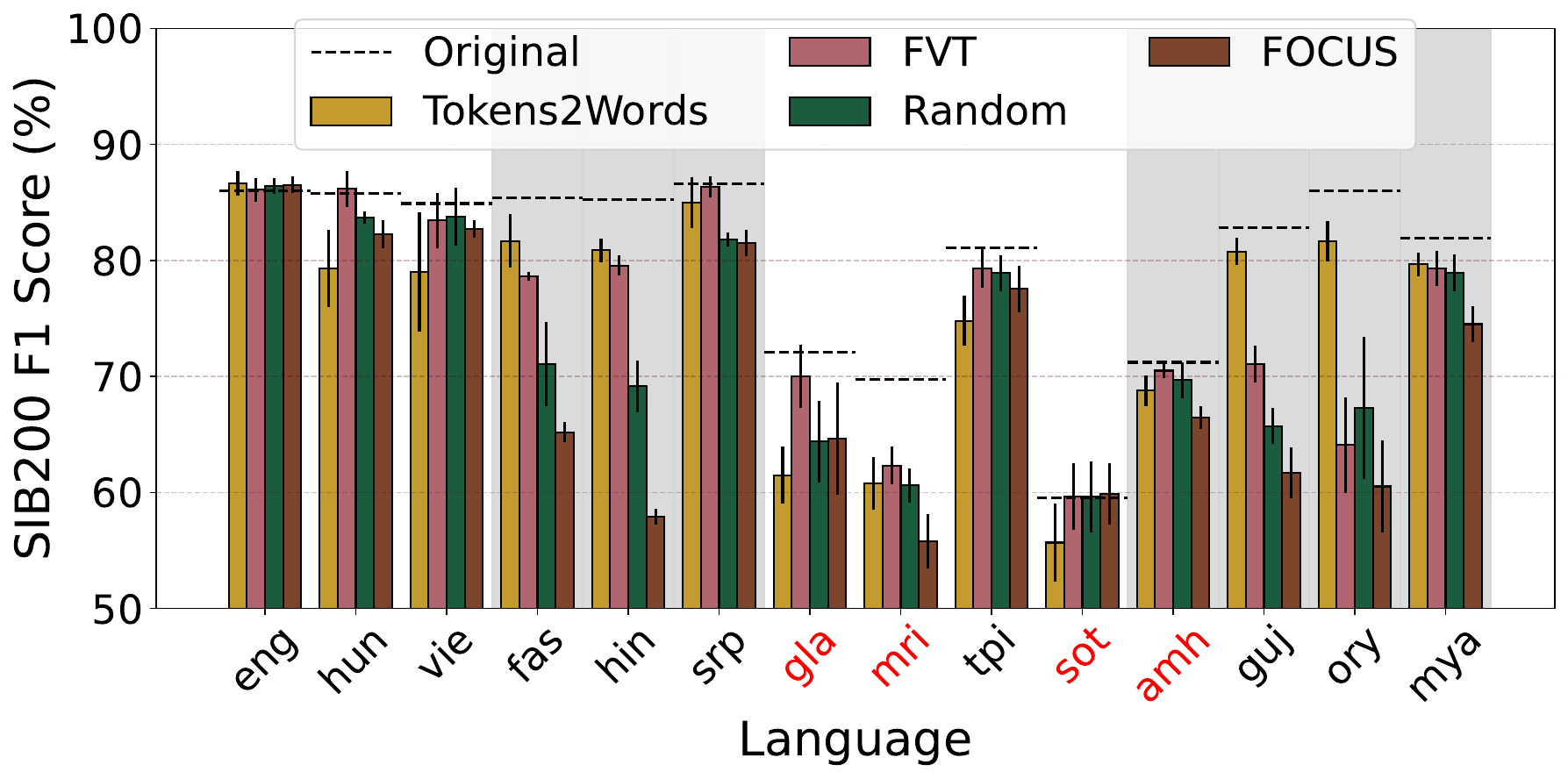}
        \caption{}
        %\phantomcaption{}
        \label{fig:sib200_main_result_init}
    \end{subfigure}%
    \begin{subfigure}{0.48\textwidth}
        % \vspace{0.5em}
        \centering
        \includegraphics[width=\linewidth]{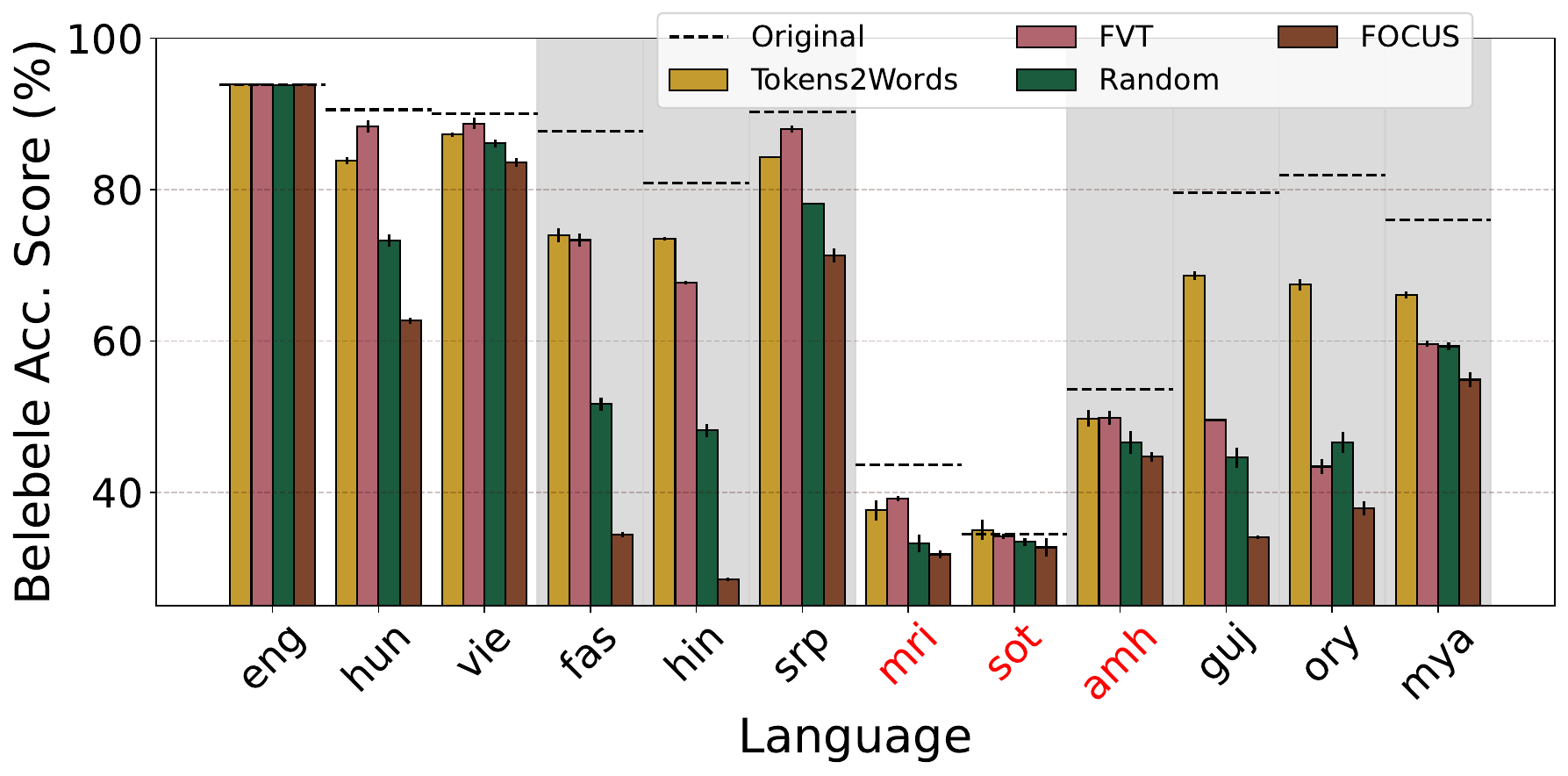}
        \caption{}
        %\phantomcaption{}
        \label{fig:belebele_main_result_init}
    \end{subfigure}
    \caption{Performance across initialization methods on (\textit{a}) SIB200 and (\textit{b}) Belebele benchmarks. \patchscopes-based initialization tends to outperform baseline methods in non-Latin-script languages ($|C_{train}|=$10k). This result also holds for BPB experiments~(\cref{fig:bpb_10k}). Model: Qwen3-30B-A3B. The model does not support languages marked in red.}
    \label{fig:main}
\end{figure*}

\subsection{Challenging Received Wisdom: A Case Against Frequency-based Heuristics}
\label{sec:challenging_bpe}
The use of frequency-based heuristics to decide on an expanded tokenizer~(\texttt{Stage 1}) has been treated as an unquestionable axiom~\citep[\textit{inter alia}]{cui2023efficient, yamaguchi2026howcan}. We challenge this convention by asking: \textit{Can we improve the performance-efficiency tradeoff offered by BPE-based expansion methods?} To that end, we construct a Pareto front using models expanded via BPE-selected items. In \cref{fig:pareto_qwen3_30b}, we show these Pareto fronts for five languages that are highly over-fragmented.  Each point on the front corresponds to a model trained with a certain expansion budget, with FVT \embedding initialization.\footnote{We cap the expansion budget at the number of unique words in $C_{train}$ successfully detokenized by \patchscopes. We analyze the performance and efficiency of models expanded using the BPE strategy for varying proportions, $\gamma$, of the maximum budget, where $\gamma\in\{1,0.5,0.25,0.125,0.0625,0.03125\}$. A more detailed explanation is provided in \S~\ref{sec:other_exp_choices}.} Overlaying the \patchscopes-expanded models on the same axes\footnote{\swpatchscopes-expanded models are also included for completeness.}, we show that interpretability-based expansion methods comfortably push the front towards a better tradeoff. Moreover, none of the \embedding initialization methods we consider approaches the performance-efficiency values achieved by the interpretability methods~(\cref{sec:additional_pareto_results}). In the supplementary Figure~\ref{fig:pareto_front_with_fvt_init} in Appendix~\ref{sec:augmenting_pareto_front_with_fvt}, we show that even when embeddings for items selected via \patchscopes are initialized using a heuristic like FVT, the result still beats the Pareto front, underscoring the true value of interpretability-based item selection. 
Even with a larger training corpus, BPE-expanded models gain some token efficiency over \patchscopes-expanded ones but suffer significant performance drops (up to 60+\% in some languages), showing the importance of vocabulary item selection at scale~(Figure~\ref{fig:bpe_vs_psc}).  Overall, our findings suggest that a poor choice of item selection strategy constitutes a fundamental bottleneck that hampers the effectiveness of \embedding initialization methods. 

\subsection{A Case for Interpretability-based Initialization}
\label{sec:case_for_interp_init}  
We turn our attention to \emph{how} to initialize \embeddings~(\texttt{Stage 2}). While random initialization seems strong, literature suggests that good initialization provides immense \calibration~efficiency~\citep[\emph{inter alia}]{dobler-de-melo-2023-focus, minixhofer2024zero,liu-etal-2024-ofa}. 

In \cref{fig:sib200_main_result_init,fig:belebele_main_result_init}, we compare \patchscopes initialization against other baseline initializations for the same expanded vocabulary determined using \patchscopes. Across the two benchmarks, we observe that \patchscopes outperforms all baselines for most languages written in non-Latin scripts, with pronounced gaps in the low-resource ones. For Latin alphabet-based languages, FVT performs best, suggesting that no single strategy dominates across all settings. Further, \patchscopes performs poorly for languages not officially supported by the model~(marked in red). Since these languages are not explicitly trained on, hidden activations are likely to be weaker, causing poor initializations. Lastly, FOCUS remains consistently poor, echoing \citet{yamaguchi2026howcan}'s findings.    

In \cref{fig:corpus_variance_tokreduction_performance}, we analyze token efficiency gains and their impact on performance when we move to a larger training and \calibration~corpus. We divide our analysis across two dimensions: script and language resourcedness. Each quadrant reveals a distinct pattern: 
\begin{inparaenum}[(i)]
    \item high-resource Latin-script languages~(\textit{top left}) show minimal efficiency gains with little performance drop, indicating smaller gaps in over-fragmentation;
    \item high-resource non-Latin-script languages~(\textit{top right}) show largely stable performance with smaller performance drops, the best case for vocabulary expansion;
    \item low-resource Latin-script languages~(\textit{bottom left}) often show a higher performance drop with minimal efficiency gains with a larger corpus; and
    \item low-resource non-Latin-script languages~(\textit{bottom right}) show high efficiency gains with moderate performance drops, indicating that this category stands to gain immensely from interpretability-based expansion. 
\end{inparaenum}

See~\cref{sec:intialization_comparison_results} and \cref{sec:more_perf_reduction_tradeoff} for more comparisons.

\subsection{\swpatchscopes: Pushing the Efficiency Ceiling with Minimal Performance Tradeoff}

\begin{figure*}
    \centering
    
    \begin{subfigure}{0.49\textwidth}
        \centering
        \includegraphics[width=\linewidth]{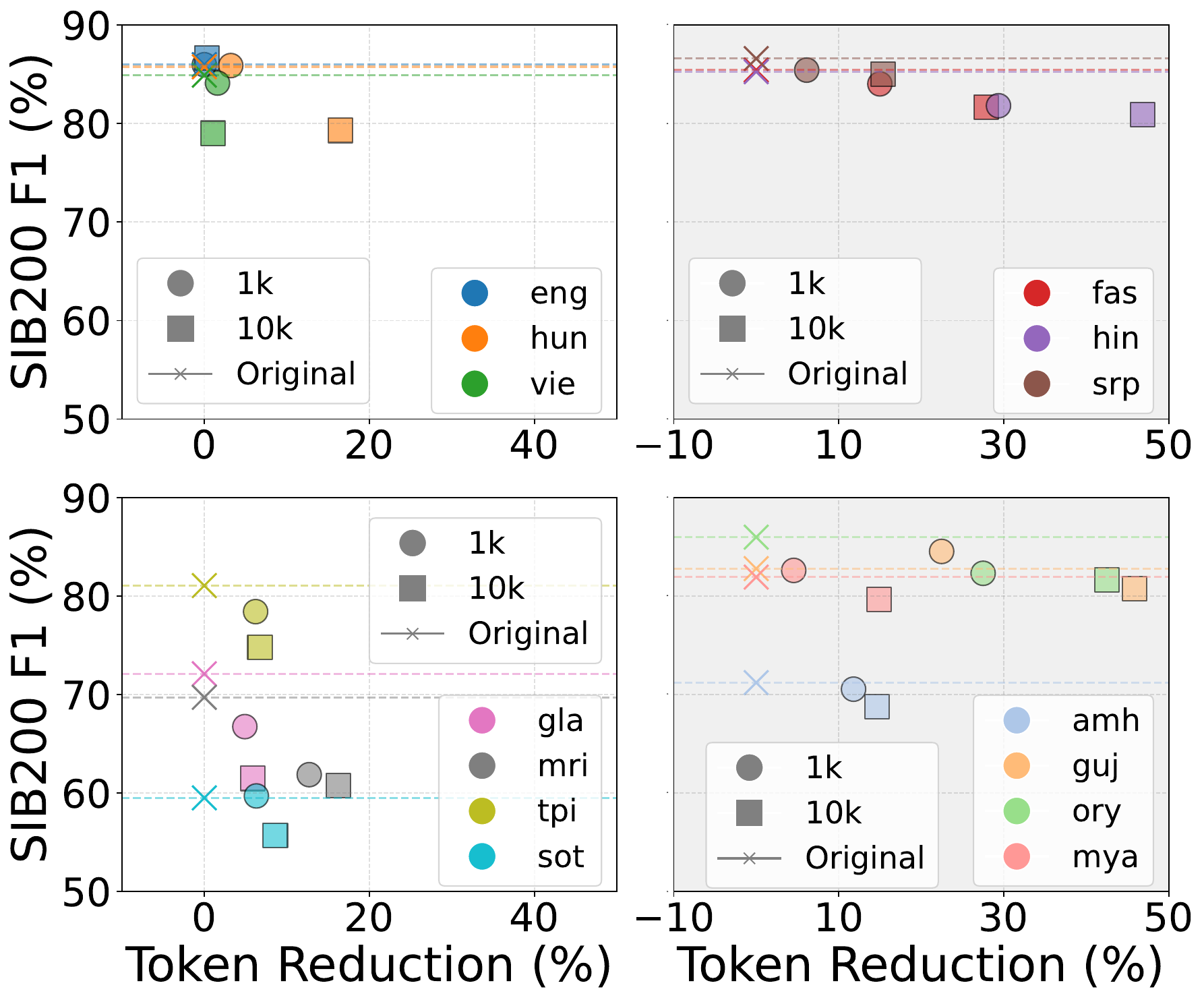}
        \caption{}
        \label{fig:corpus_variance_tokreduction_performance}
    \end{subfigure}%
    ~
    \begin{subfigure}{0.49\textwidth}
        \centering
        \includegraphics[width=\linewidth]{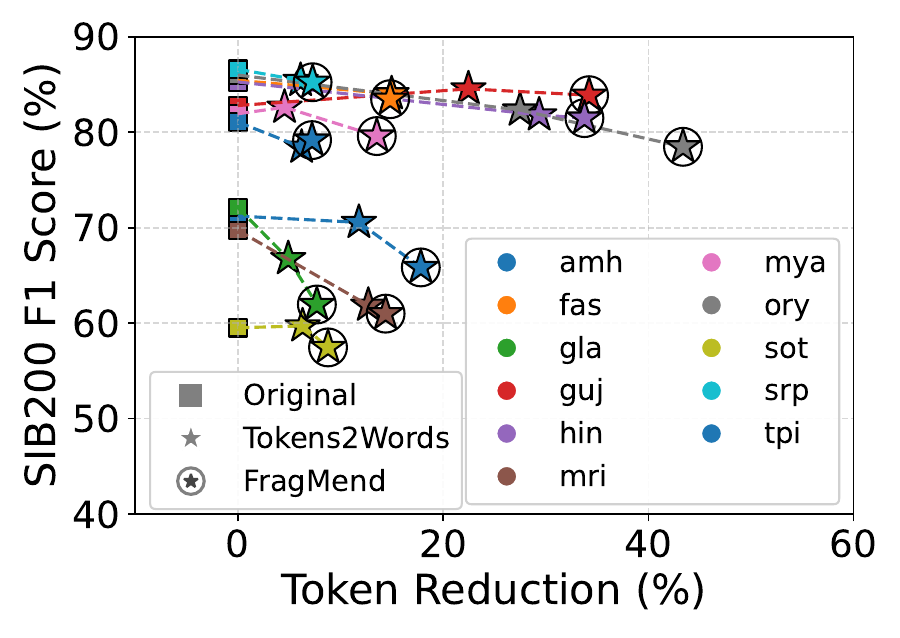}
        \caption{}
        %\phantomcaption{}
        \label{fig:swpsc_perf_vs_efficiency}
    \end{subfigure}%

    \caption{(a) An increase in $C_{train}$ size has different efficiency-performance effects for different language groups with \patchscopes initialization. (b) Performance-efficiency tradeoff between the original model~(lowest token efficiency), \patchscopes, and \swpatchscopes as computed with Qwen3-30B-A3B. \swpatchscopes provides unrealized efficiency gains, with minimal performance degradation over \patchscopes for non-Latin-script languages.}
    
\end{figure*}

As shown in \cref{fig:sw_vspsc_efficiency_analysis}, \patchscopes is bound by an efficiency ceiling imposed by its restriction to full words. This constraint affects low-resource languages the most, as they suffer from a higher rate of over-fragmentation due to the scarcity of corpora. By expanding the candidate set to include subwords, \swpatchscopes achieves 2-3$\times$ more vocabulary additions than \patchscopes with as few as 1000 training sequences, translating to token efficiency gains of up to 3$\times$ for low-resource languages such as Burmese~(\texttt{mya}). 
\cref{fig:swpsc_perf_vs_efficiency} shows that \swpatchscopes is more token-efficient than \patchscopes for every language, with minimal performance degradation. 
For \texttt{guj}, \texttt{hin}, and \texttt{srp}, \swpatchscopes gives us increased token efficiency for free (i.e., no performance drop relative to \patchscopes).  
Since both \swpatchscopes and \patchscopes depend on activation quality, we observe consistent drops in performance for out-of-distribution languages~(e.g., \texttt{amh}, \texttt{gla}, \texttt{sot}), echoing findings from \cref{sec:case_for_interp_init}. For \texttt{ory} and \texttt{mya}, efficiency gains are substantial with only minor drops in performance. 
This tradeoff is favorable, particularly given the downstream impact on API costs.

In \cref{sec:comparing_swpsc_vs_fvt_initialization}, we compare \swpatchscopes initialization against FVT for subword-detokenized vocabulary items. We select FVT for its consistent performance in our previous experiments~(\cref{sec:case_for_interp_init}). For Qwen3-30B-A3B, \swpatchscopes provides substantial gains for the non-Latin-script, low-resource languages of Gujarati and Odia. 
Gains are modest for smaller models~(Qwen3.5-4B and TinyAya-Global) due to lower-quality representations, owing to the `\textit{curse of multilinguality}'~\citep{conneau-etal-2020-unsupervised}, which directly affects interpretability-based detokenization (\cref{sec:model_scale_analysis_sec}). 
As models improve, we hypothesize that the efficacy gap between interpretability-based methods like \swpatchscopes and frequency-based methods will only grow.

\subsection{\swpatchscopes Design Choices}
\label{sec:swpsc_design_choices}

\begin{wraptable}{r}{0.39\textwidth}%[t]
\centering
\begin{tabular}{@{}lrrrr@{}}
\toprule
\small
\multirow{2}{*}{\textbf{Config.}}  & \multicolumn{4}{c}{\textbf{Languages}} \\ \cmidrule{2-5}
                    & \texttt{amh}  & \texttt{guj}  & \texttt{ory} & \texttt{mya} \\ \midrule
\textbf{R$_{m}$}    & \cellcolor{gray!30}44.8          & \cellcolor{gray!30}50.1          & \cellcolor{gray!30}44.7         & \cellcolor{gray!30}50.2 \\
\textbf{R$_{e}$}    & \cellcolor{gray!30}44.7          & \cellcolor{gray!30}55.0          & \cellcolor{gray!30}49.0         & \cellcolor{gray!30}54.4 \\
\textbf{R$_{m}$,T}  & \cellcolor{gray!30}47.3          & \cellcolor{gray!30}65.4          & \cellcolor{gray!30}58.9         & \cellcolor{gray!30}61.2 \\
\textbf{R$_{e}$,T}  &  \cellcolor{gray!30}47.1          & \cellcolor{gray!30}68.2          & \cellcolor{gray!30}62.5         & \cellcolor{gray!30}62.1 \\
\textbf{R$_{m}$,T,P}& \cellcolor{gray!30}\textbf{47.7} & \cellcolor{gray!30}67.0          & \cellcolor{gray!30}62.0         & \cellcolor{gray!30}\textbf{62.2} \\
\textbf{R$_{e}$,T,P}& \cellcolor{gray!30}47.6          & \cellcolor{gray!30}\textbf{68.9} & \cellcolor{gray!30}\textbf{63.0}& \cellcolor{gray!30}61.6 \\\bottomrule
\end{tabular}
\caption{\swpatchscopes~results on the Belebele benchmark. Full set of results in \cref{tab:analysis_sw_full}.}
\label{tab:analysis_sw_short}
\end{wraptable}

\textbf{Reduction Strategy.} 
As a given subword may be successfully detokenized across multiple words, we need a reduction strategy to select/aggregate among multiple candidate initializations. We explore two strategies: (a) averaging the candidate initializations~(R$_m$), and (b) selecting the initialization from the instance where detokenization succeeded at the earliest layer in the transformer~(R$_e$). The latter is inspired by previous work showing that compositionality often emerges in early layers~\citep{lad2024the,kaplan2025from}.

\textbf{Minimum Subword Length (T).} Vocabulary items selected via interpretability-based methods are added without intermediate merges on the original vocabulary. This causes `\textit{vocabulary expansion perversity}'~(\cref{sec:definitions}), which leads to worsened token efficiency. To mitigate this phenomenon, we introduce an untuned length filter that removes items shorter than 3 UTF-8 characters, as shorter items are more likely to increase perversity.

\textbf{Full-word Preference (P). } Since \swpatchscopes detects both subword and full-word detokenization and allows for the addition of complete words, \patchscopes is a strict subset of \swpatchscopes. When a full word also occurs as an affix in a different word~(e.g., `\textit{cat}' in `\textit{caterpillar}'), this design choice prioritizes choosing initializations corresponding to full-words over those of subwords. Intuitively, activations for a full-word~(say, `\textit{cat}' as a standalone word) better capture its semantic content than those for subwords~(say, `\textit{cat}' in `\textit{caterpillar}'). 

\cref{tab:analysis_sw_short} reports \swpatchscopes performance on the Belebele benchmark across configurations of these parameters. Minimum length thresholding~(T) is perhaps the most important configuration choice, with performance jumps of up to 15 points in certain cases. Although thresholding reduces the number of added items, its impact on token efficiency is negligible~($\sim$1\% increase). Earliest-layer reduction~(\textbf{R$_e$}) and full-word preference~(\textbf{P}) each contribute modest gains. We therefore adopt~(\textbf{R$_{e}$,T,P}) as our preferred configuration.

\subsection{Model Scale Analysis}
\label{sec:model_scale_analysis_sec}
\begin{figure*}[]
    \centering
    \begin{subfigure}{0.46\textwidth}
        \centering
        \includegraphics[width=\linewidth]{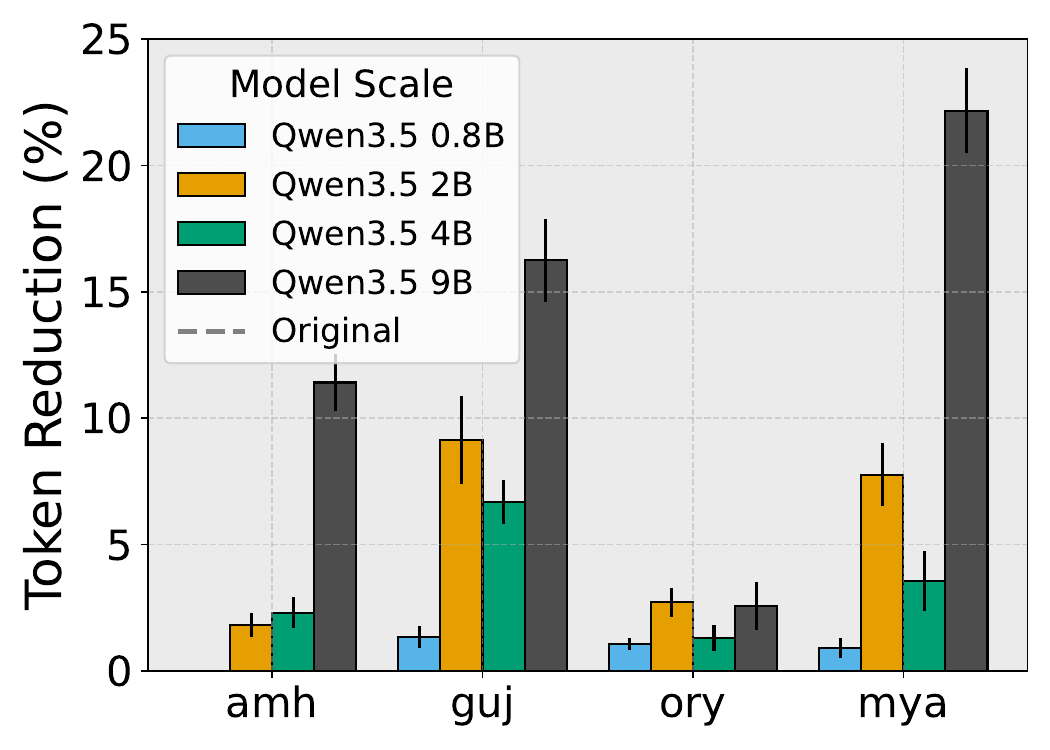}
        \caption{}
        %\phantomcaption{}
        \label{fig:model_scale_analysis_swpsc}
    \end{subfigure}%
    ~ 
    \begin{subfigure}{0.46\textwidth}
        \centering
        \includegraphics[width=\linewidth]{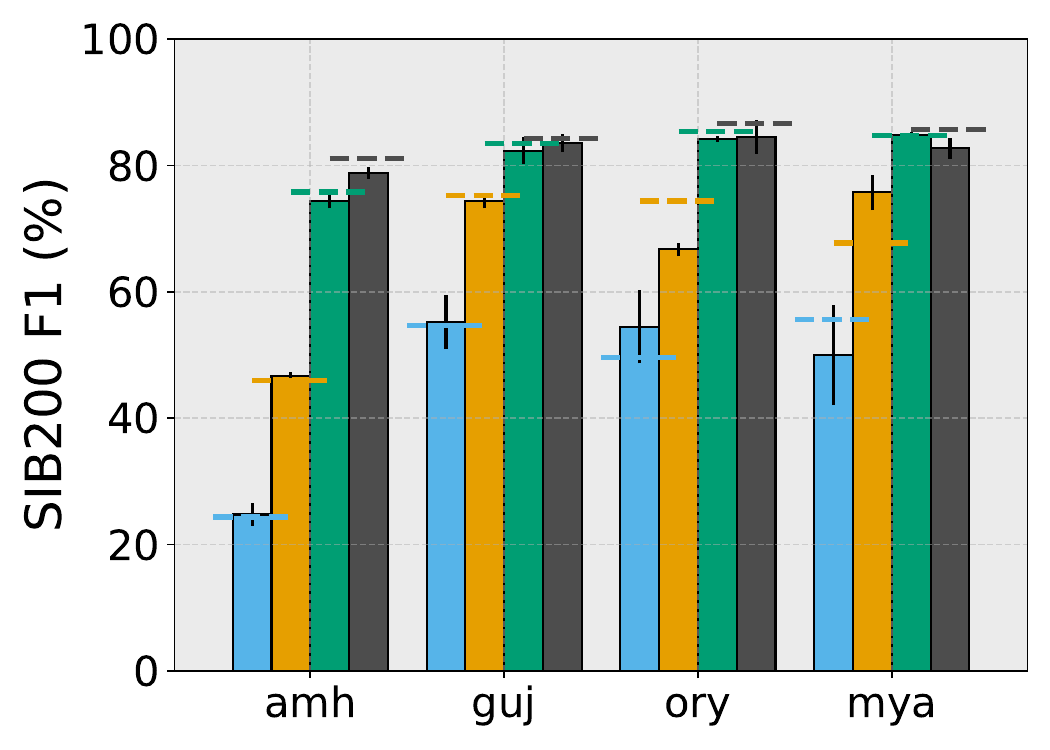}
        \caption{}
        %\phantomcaption{}
        \label{fig:model_scale_analysis_swpsc_perf}
    \end{subfigure}
    \caption{(a) Token reduction and (b) SIB200 performance across different model sizes with \swpatchscopes-based item selection and initialization~($|C_{train}|=$1k). }
\end{figure*}

The success of interpretability-based methods depends heavily on activation quality. Smaller LMs are disadvantaged at detokenization for two reasons: 
\begin{inparaenum}[(a)]
    \item multiple languages compete for a smaller set of parameters~(`\textit{curse of multilinguality}'), and
    \item fewer layers reduce the search space for detokenization.
\end{inparaenum}
We hypothesize that detokenization improves with model scale. Figures~\ref{fig:model_scale_analysis_swpsc} and \ref{fig:model_scale_analysis_swpsc_perf} examine this hypothesis across four Qwen3.5 models for the four low-resource non-Latin-script languages. Token reduction rates~(Fig.~\ref{fig:model_scale_analysis_swpsc}) increase from 0.8B to 9B parameters, with the exception of the 4B model. We hypothesize that this exception is due to the model's propensity to generate thinking tokens, which inhibit detokenization. Results (Fig.\ref{fig:model_scale_analysis_swpsc_perf}) show comparable performances relative to the original model despite substantial efficiency gains. These results suggest that larger models benefit more from interpretability-based methods.

\section{Related Work}
We study vocabulary expansion, where the original vocabulary is a subset of the expanded one, distinguishing it from general vocabulary replacement strategies~\citep{dobler-de-melo-2023-focus,nakash2025adaptivocab} that often discard some original items. While we choose the expansion setting, our method trivially extends to vocabulary replacement settings.
Embedding initialization for newly added tokens has been approached through three broad strategies. \textit{Heuristics-based} methods range from simple randomized sampling~\citep{hewitt2021initializing}, or averaging of constituent subword \embeddings~\citep{gee-etal-2022-fast}, to similarity-weighted combinations, often using auxiliary resources such as embedding spaces~\citep{minixhofer-etal-2022-wechsel,dobler-de-melo-2023-focus,mundra-etal-2024-empirical,liu-etal-2024-ofa,remy2024transtokenization,li-etal-2025-tokalign} and/or bilingual dictionaries~\citep{minixhofer-etal-2022-wechsel, sakajo-etal-2025-dictionaries}. \citet{yamaguchi2026howcan} provides a comprehensive comparison of these approaches. \textit{Hypernetwork-based} methods instead train a network to directly predict new \embeddings~\citep{pinter-etal-2017-mimicking, minixhofer2024zero, ozeren-etal-2025-hyperofa}; while natural baselines, we benchmark against heuristics-based methods, as they represent the dominant approach in the vocabulary expansion literature. Most recently, \textit{interpretability-based} methods~\citep{kaplan2025from, dobler2026token} use model activations to inform both item selection and \embedding initialization: a direction our work explores. In the interest of space, we include a detailed discussion in \cref{sec:ext_related_work}.

\section{Conclusion}
We present an extensive study of interpretability-based vocabulary expansion methods for reducing over-fragmentation, offering a much better performance-efficiency tradeoff as compared to frequency-based methods for vocabulary expansion.
The primary interpretability-based method we study, \patchscopes, serves as a better initialization strategy than other popular baseline methods, yet possesses a low efficiency ceiling. 
We push this ceiling by introducing \swpatchscopes. \swpatchscopes leverages the phenomenon of "\textit{subword detokenization}," where LLMs progressively reconstruct intermediate subword tokens before recovering the full word. 
\swpatchscopes improves token efficiency while preserving downstream performance relative to \patchscopes, particularly for low-resource languages that use non-Latin scripts. 

\section*{Acknowledgments}
We are grateful to the members of the Utah NLP lab for their feedback, with particular thanks to Ana Marasovi\'{c}, Nate Stringham, Fateme Hashemi Chaleshtori, Lucas Pearce, and Md Farhan Ishmam for their insightful discussions. We thank Kenneth Marino for inspiring the opening line in the abstract. We also thank Guy Kaplan and Yuval Reif for their patient support with the \patchscopes codebase. We thank the anonymous reviewers for their valuable feedback and engagement throughout the review process; their comments have substantially improved this paper. The support and resources from the Center for High Performance Computing at the University of Utah are gratefully acknowledged. This material is based in part upon work supported by the National Science Foundation under Grants \#2411319 and \#2217154. Any opinions, findings, and conclusions or recommendations expressed in this material are those of the authors and do not necessarily reflect the views of the National Science Foundation.

\bibliography{colm2026_conference}
\bibliographystyle{colm2026_conference}

\appendix
\section{Definitions}
\label{sec:definitions}

In this section, we briefly introduce the metrics and concepts we use in the paper. 

\paragraph{Tokenizer/Token Efficiency. }We define tokenizer efficiency, or token efficiency, as the number of tokens required to encode a given amount of information. Since our premise centers on token over-fragmentation in the original model's tokenizer across various languages, we consider the original model the least token-efficient. Subsequently, every vocabulary-expanded model can be measured relative to the original, in terms of the number of tokens reduced post-expansion on some dataset $D$. Consequently, we define ``\textit{Token Reduction~(\%)}'' as a metric to measure the tokenizer efficiency of an expanded tokenizer relative to the original tokenizer.
\begin{equation}
    Token Reduction_D~(T_{exp}) =  (1-T_{exp}(D)/T_{orig}(D))
    \label{eq:reduction}
\end{equation}
where $T_{exp}$ is the expanded tokenizer, and $T_{orig}$ is the original model's tokenizer. When mentioned in the context of a downstream task, $Token Reduction_D$ is always computed on the respective task data. When mentioned generally, $Token Reduction_D$ is computed on the selected Glot-500c test set~(\S\ref{sec:experimental_setup}).

\paragraph{Performance Conservation.} We use this metric in our Pareto analysis. It simply computes the ratio of the performance of the expanded model relative to that of the original model. However, in most of our experiments, we report the raw performance metrics for each task.

\paragraph{Token Ratio.} This metric is used to measure the severity of over-fragmentation in tokenizers. Specifically, \textit{Token Ratio} for a language $l$ computes the ratio of the number of tokens required by the tokenizer to encode a certain parallel dataset~($D$) in $l$, over the number of tokens required to encode the same dataset in English. This is equivalent to the \textit{tokenization premium} as defined by \citet{petrov2023language}. 
\begin{equation}
    Token Ratio_l~(T) =  T(D_l)/T(D_{eng})
\end{equation}

\paragraph{Character Ratio. }Similar to its token counterpart, \textit{Character Ratio} for a language $l$ is the number of UTF-8 characters  used to encode a certain parallel dataset~($D$) in $l$ over the number of UTF-8 characters used to encode the same dataset in English.

\paragraph{Vocabulary Expansion Perversity. }In popular tokenizer implementations~(in our case, HF tokenizers), newly added vocabulary items possess a special encoding preference. These items are encoded between the pre-tokenization and tokenization steps and hence take priority over the original vocabulary tokens, which are then tokenized using the base tokenizer's merges over this pre-encoded sequence. This preference often leads to competition with the original tokenization process. As a result, in certain cases, adding certain new items leads to increased over-fragmentation. As an example, the word `\textit{development}' might be tokenized as: \texttt{deve + lop + ment} per the original tokenizer. Say the item `\textit{elop}' is added, a tokenizer with the expanded vocabulary could then tokenize it as: \texttt{de+v+elop+ment}, as the item `\textit{elop}' has higher priority, causing over-fragmentation to worsen. We term this phenomenon ``\textit{vocabulary expansion perversity}.`` 
\section{Discussion and Expanded Related Work}
\label{sec:ext_related_work}

\paragraph{Comparisons to Vocabulary Replacement. }Vocabulary replacement, or vocabulary transfer, is the general idea of switching a pre-trained language model's vocabulary with a new one better suited for the target language~\citep[e.g.,][]{dobler-de-melo-2023-focus} or domain~\citep[e.g.,][]{nakash2025adaptivocab}, where the new one may or may not overlap with the original vocabulary. Vocabulary expansion is the specific scenario in which the original vocabulary is a perfect subset of the new one. The choice between expansion and replacement is often dictated by the intended application and model-size constraints. Replacement is preferred when the use case is restricted within the target language/domain~\citep{mundra-etal-2024-empirical}, and cross-language/domain performance is secondary. In contrast, \citet{yamaguchi2026howcan} show that replacement underperforms expansion in low-resource scenarios. While expansion may introduce a significant number of parameters for older encoder models~\citep{devlin-etal-2019-bert, conneau-etal-2020-unsupervised}, the overhead is negligible for much larger, contemporary LLMs~(e.g., expanding with 1000 items introduces $<$5M parameters). Regardless, our method easily transfers to a replacement setting.

\paragraph{Embedding Initialization Strategies. }Literature on embedding initialization for new vocabulary items is rich. For the ease of the reader, we organize the approaches into three categories: 
\begin{inparaenum}[(a)]
    \item \textit{Heuristics-based},
    \item \textit{Hypernetwork-based}, and
    \item \textit{Interpretability-based} initialization.
\end{inparaenum}

\textit{Heuristics-based} initialization relies on informed, sometimes model-based, strategies to initialize embeddings. The simplest approach samples from a multivariate normal distribution centered at the average input and output embeddings~\citep{hewitt2021initializing}. \citet{gee-etal-2022-fast} initialize \embeddings of a new item by averaging those of its constituent subwords, per the original tokenizer. Several prior studies instead train and/or use auxiliary embeddings to compute similarities between new and original vocabulary items. W{\small ECHSEL}~\citep{minixhofer-etal-2022-wechsel} initializes new item \embeddings as a weighted average of the base model's \embeddings, where the weights are computed using fastText n-gram vectors aligned across a source and target language via a bilingual dictionary. FOCUS~\citep{dobler-de-melo-2023-focus} eliminates the need for dictionary-based alignment by using overlapping items between the original and expanded tokens as anchors; \embeddings for a new item are a weighted sum over the anchor \embeddings, with weights derived using fastText similarity. Several methods learn variants of alignment weight matrices to transform original \embeddings into initializations for the newly added items~\citep{mundra-etal-2024-empirical,liu-etal-2024-ofa,remy2024transtokenization, li-etal-2025-tokalign}.  \citet{sakajo-etal-2025-dictionaries} reintroduce bilingual dictionaries in an approach that iteratively removes target vocabulary items from the tokenizer once a one-to-many mapping to the source language's definition is established. As a separate direction, \citep{kim2024efficienteffectivevocabularyexpansion} study a multi-stage \calibration strategy for expanded vocabularies in Korean. \cite{yamaguchi2026howcan} present a detailed comparison and analysis of heuristics-based initialization and \calibration methods. 

\textit{Hypernetwork-based} initialization trains networks to directly predict the \embeddings of a new vocabulary item, instead of relying on similarity heuristics. Such a network that generates weights for a different network is called a hypernetwork~\citep{ha2016hypernetworks}. Among the earliest works, \citet{pinter-etal-2017-mimicking} introduced MIMICK, which learns a BiLSTM-based hypernetwork that maps character sequences to embeddings. MIMICK trains on sequences of in-vocabulary items and is applied to out-of-vocabulary items. More recently, ZeTT~\citep{minixhofer2024zero} trains a language model-based hypernetwork using the causal language modeling objective. The hypernetwork replaces the original tokenizer with a target one and learns the mapping between an out-of-vocabulary item's decomposition, per the original tokenizer and its corresponding \embeddings. Similarly, \citet{ozeren-etal-2025-hyperofa} introduce H{\small YPER}O{\small FA}, which replaces the much simpler convex combination over original \embeddings used by its predecessor O{\small FA} with a more expressive hypernetwork. \citet{han2025adapters} use vocabulary adapters to learn new vocabulary items embeddings as a combination of the original ones. We defer a direct comparison with these methods to future work and instead benchmark against heuristics-based methods, which represent the dominant approach in the vocabulary expansion literature.

\textit{Interpretability-based} initialization is a new direction that uses model internals to find the best initialization. \citet{kaplan2025from}~(as described in detail in \cref{sec:background}) use the concept of `\textit{detokenization}', which argues for an internal lexicon in LLMs larger than the model vocabulary. Their method, \patchscopes, uses the activation where a certain out-of-vocabulary item is detokenized, and transforms it to create (un-)embedding initializations. Recently, \citet{dobler2026token} introduce TokenDistillation, where the input embedding initializations for new tokens are trained such that the contextual representations of the originally fragmented item and those of the item post-addition remain identical. Note that token distillation is only applicable to input embeddings, and hence is an inappropriate comparison for token efficiency. Both \citet{kaplan2025from} and \citet{dobler2026token} present analyses of higher-resource languages~(Arabic and French, etc.). Our work is the first to extensively study the promise of interpretability-based initialization methods for low-resource languages, especially those written in a non-Latin script.

\paragraph{Connection to the Broader Interpretability Literature.}
Our vocabulary expansion work falls under the umbrella of actionable interpretability where we develop and use interpretability techniques for a clear downstream purpose~\citep{orgad2026position}.
Actionable interpretability also includes work on model internal confidence estimation~\citep{beigi-etal-2024-internalinspector, subramani-etal-2025-mice, subramani2026the}, inference-time intervention \& control including concept or attribute steering~\citep[\textit{inter alia}]{subramani2019can, subramani-etal-2022-extracting, li2023inferencetime, todd2024function}, and harmful behavior detection, safety, \& remediation~\citep[\textit{inter alia}]{sun2025redeep, dunefsky2025oneshot, kramar2026building}.

We use interpretability for vocabulary expansion, which requires understanding of how language models encode different linguistic properties across languages.
There have been a long line of work in this direction, sometimes referred to as BERTology research~\citep[\textit{inter alia}]{hewitt-manning-2019-structural, jawahar-etal-2019-bert,tenney-etal-2019-bert,niu-etal-2022-bert, he-etal-2024-decoding, brinkmann-etal-2025-large, acevedo2026differential, bayazit-etal-2026-crosscoding}.
More recently,~\citet{li-subramani-2026-model} expanded this analysis, focusing on inflectional morphology and lexical information to modern language models across a variety of languages.

\section{Over-fragmentation Across Various LLMs}
\begin{figure*}[h]
    \centering
    \begin{subfigure}{0.49\textwidth}
        \centering
        \includegraphics[height=1.5in]{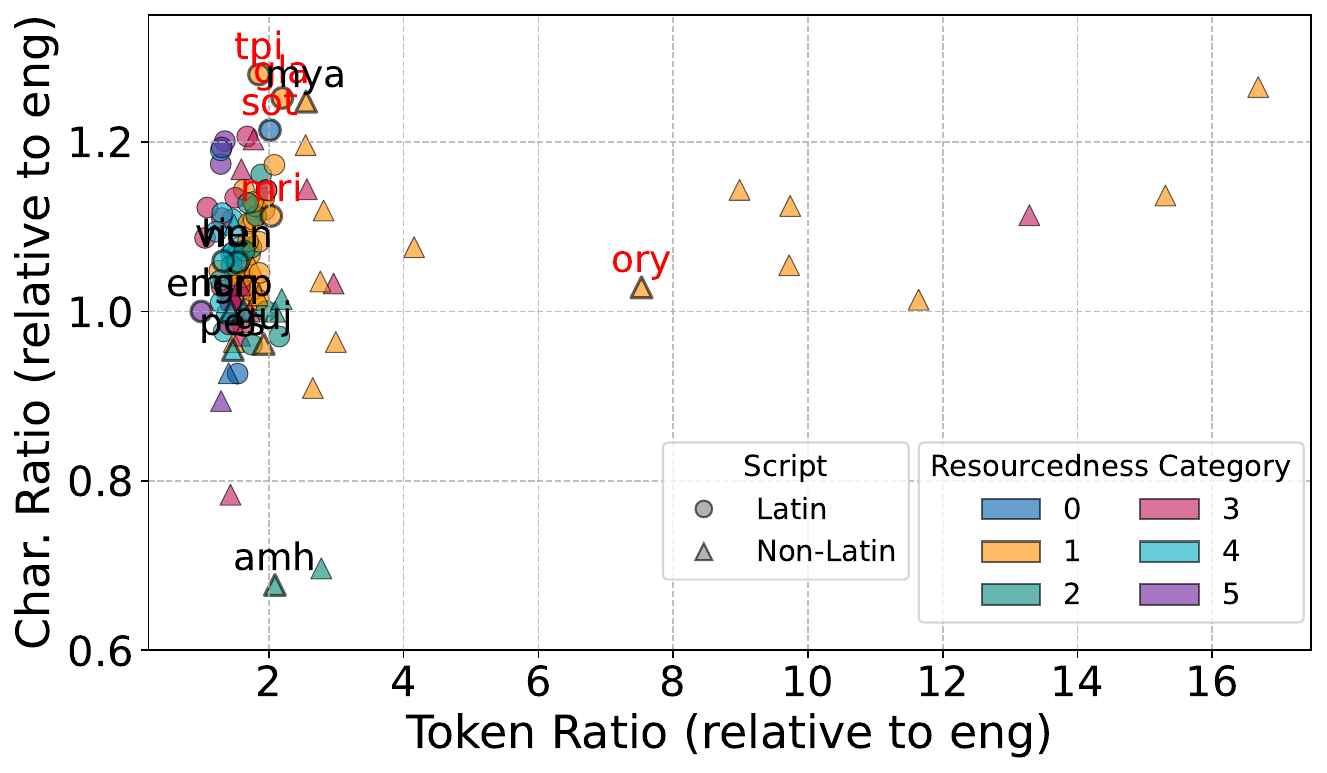}
        \caption{}
        \label{fig:token_ratio_tinyaya_3b}
    \end{subfigure}%
    ~ 
    \begin{subfigure}{0.49\textwidth}
        \centering
        \includegraphics[height=1.5in]{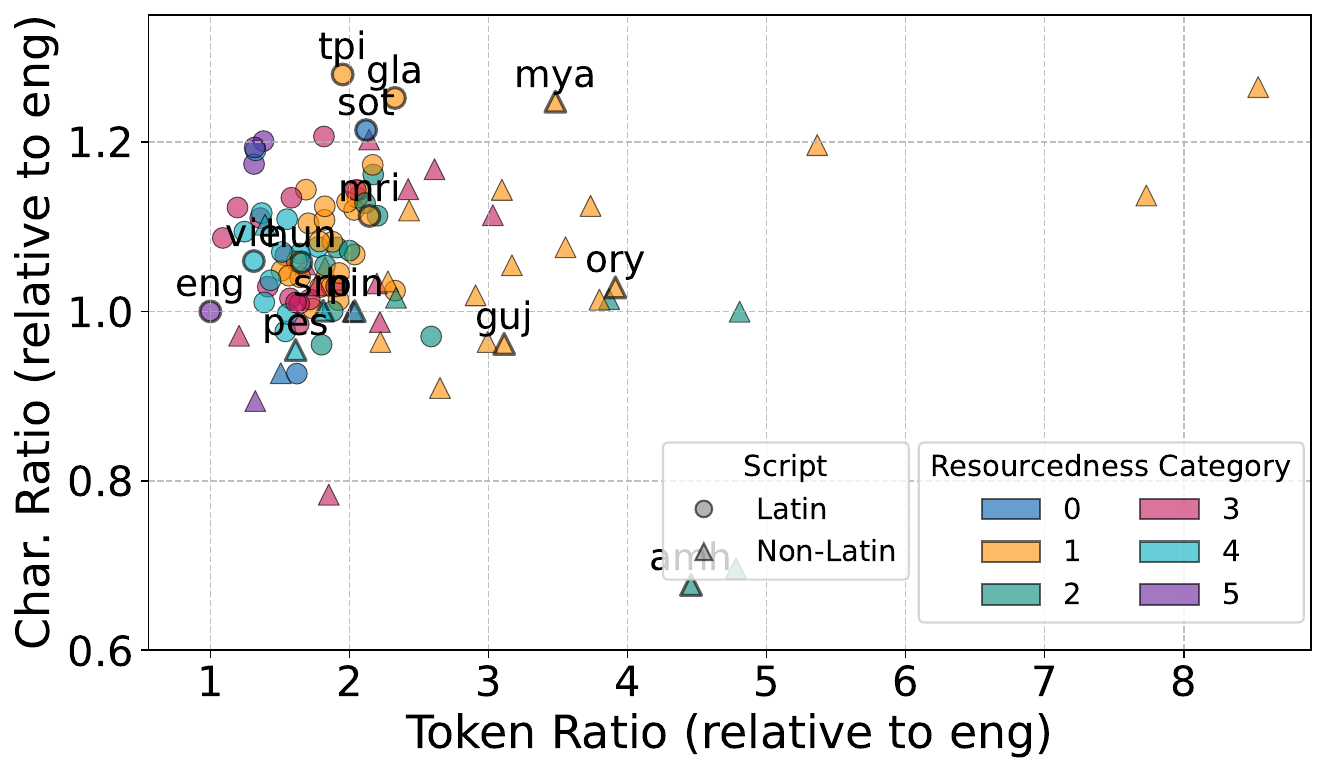}
        \caption{}
        \label{fig:token_ratio_qwen35_4b}
    \end{subfigure}
    \caption{Token disparity across different languages for (a) TinyAya-Global and (b) Qwen3.5-4B compared using the ratio of tokens vs. characters for the same information~(SIB200~\citep{adelani-etal-2024-sib} training split) as compared to English (similar to \cref{fig:token_ratio_qwen3_30b}). Languages studied in the paper are annotated; languages in red are not officially supported by the respective models.}
    
\end{figure*}

Additional token disparity analysis is shown for Tiny-Aya Global and Qwen3.5-4B~(Figures~\ref{fig:token_ratio_tinyaya_3b} and  \ref{fig:token_ratio_qwen35_4b}, respectively). Disparities against languages written in a non-Latin script persist. However, improvements can be seen across several languages when compared to the older Qwen3-30B-A3B model.

\paragraph{Character Length as a Comparative Metric for Token Ratio.} There are two tokenizer-independent atomic units at which text can be represented: bytes and characters (in our case, UTF-8). For readers, characters are the most atomic elements of text, and hence we focus on characters as a tokenizer-invariant metric. As seen in Figure~\ref{fig:token_ratio_qwen3_30b}, almost all of the languages lie within the range [0.9,1.3], implying that the information density across all the languages considered in the figure is roughly similar~(CJK languages are notable exceptions and are not represented in the figure). Here, the variance may simply be a factor of translational artifacts. Character ratios account for such artifacts and are therefore a useful metric to compare with token ratio. Consequently, we can expect tokens to scale proportionally to the character lengths~(hence, on or near the diagonal). 

\section{Experimental Choices and Design Decisions}
\label{sec:exp_choices_design_decisions}
In this section, we detail our experimental choices and design decisions.

\subsection{Choice of Languages}
\label{sec:lang_choice}
Our languages of study are carefully selected keeping in mind aspects such as script type, language-resourcedness, presence of evaluation resources, diversity and tokenizer disparity. We define a budget of 14 languages split into two even groups: those written in the Latin script and those using a non-Latin script. Each group was further split into subgroups of three and four languages, high-resource and low-resource respectively. According to \citet{joshi-etal-2020-state}'s taxonomy, we define languages belonging to categories 0,1,2 as low-resource, and those in categories 4 and 5 as high-resource.  

We start with 150 languages and apply a filtering step, retaining languages present in the Glot-500c~\citep{imanigooghari-etal-2023-glot500}, and in at least one of our downstream tasks, Belebele~\citep{bandarkar-etal-2024-belebele} and SIB-200~\citep{adelani-etal-2024-sib}, yielding 111 languages. We then perform the token ratio analysis on these languages~(as shown by the x-axis in Figures~\ref{fig:token_ratio_qwen3_30b}, \ref{fig:token_ratio_tinyaya_3b} and \ref{fig:token_ratio_qwen35_4b}) on these languages. As Qwen3-30-A3B serves as our primary model for analysis, we select languages that show a poor token ratio relative to English and optimize further for script and geographic diversity. This process yields the fourteen languages for our study as shown in \cref{tab:selection_languages}.

\begin{table}[]
    \centering
    \begin{tabular}{ccc} \toprule
         & \textbf{Latin}& \textbf{Non-Latin} \\\midrule
     \textbf{High-Resource}    & \texttt{eng$_{1.00}$}, \texttt{hun$_{2.13}$}, \texttt{vie$_{1.43}$} & \cellcolor{gray!30}\texttt{pes$_{2.62}$}, \texttt{hin$_{4.46}$}, \texttt{srp$_{2.36}$}\\
     \textbf{Low-Resource}     & \texttt{gla$_{2.41}$}, \texttt{mri$_{2.33}$}, \texttt{tpi$_{2.02}$}, \texttt{sot$_{2.20}$}& \cellcolor{gray!30}\texttt{amh$_{4.12}$}, \texttt{guj$_{6.80}$}, \texttt{ory$_{9.79}$}, \texttt{mya$_{9.02}$}\\\bottomrule 
    \end{tabular}
    \caption{Languages in our study divided by script type and resourcedness. The subscript indicates the token ratio with respect to English for the Qwen-3-30B A3B tokenizer on the SIB200 training set.}
    \label{tab:selection_languages}
\end{table}

\subsection{Experimental and Hyperparameter Details}
\label{sec:hyperparameter_details}
All experiments are reported on three random seeds: $\{42,20,1984\}$. The only exception are the BPE methods on the Pareto analysis which are reported on a single seed. However, as show in Figure~\ref{fig:bpe_vs_psc}(b-d), BPE methods are seed stable even with the maximum vocabulary budget.  

For the two training corpus sizes~($|C_{train}|=\{1k,10k\}$ sequences), we consider all word-boundary delimited tokens, where hyphenated and contracted forms are considered a single token. To tackle heavy tail tokens, we filter out tokens occurring less than five times. The remaining tokens form the candidate set for the corresponding corpus size~(Table \ref{tab:candidate_set_stats}). For the BPE experiments, the maximum vocabulary addition budget is set equal to the size of this candidate set for fair comparison. In Figures \ref{fig:pareto_qwen3_30b} and \ref{fig:pareto_other_inits}, we obtain different points for the BPE method on the Pareto front by varying the proportions of this maximum budget.

\begin{table}[]
    \centering
    \begin{tabular}{crrrrrrr} \toprule
    \multirow{2}{*}{$\mathbf{|C_{train}|}$}     & \multicolumn{7}{c}{\textbf{Languages}} \\\cmidrule{2-8}
         & \texttt{eng} & \texttt{hun} & \texttt{vie} & \texttt{pes} & \texttt{hin} & \texttt{srp} & \texttt{gla} \\ \midrule
    1k  &   3 &  190  &   57  &  \cellcolor{gray!30}427  &  \cellcolor{gray!30}584 &  \cellcolor{gray!30}206  &  629  \\  
    10k &  97 & 2904  & 1256  & \cellcolor{gray!30}3772  & \cellcolor{gray!30}4342 & \cellcolor{gray!30}3441  & 5039  \\ \midrule
        & \texttt{mri} & \texttt{tpi} & \texttt{sot} & \texttt{amh} & \texttt{guj} & \texttt{ory} & \texttt{mya} \\ \midrule
    1k  & 554  &  357  &  695  &  \cellcolor{gray!30}982  &  \cellcolor{gray!30}533  &  \cellcolor{gray!30}522  &  \cellcolor{gray!30}193 \\
    10k & 4751  & 1272  & 5629  & \cellcolor{gray!30}9895  & \cellcolor{gray!30}5101  & \cellcolor{gray!30}4253   & \cellcolor{gray!30}3135 \\
    \bottomrule
     \end{tabular}
    \caption{Number of candidate words occurring at least five times in the two corpora for each language. Languages written in a non-Latin script have a gray background.}
    \label{tab:candidate_set_stats}
\end{table}

The total training~($C_{train}$) and \calibration~($C_{tuning}$) tokens are provided in Table~\ref{tab:tokens_stats} for the $|C_{train}|=10k$ corpus. Remember that $C_{train}$ is used for determining the candidate set, and both $C_{train}$ and $C_{tuning}$ sets are used for tuning. 

\begin{table}[]
    \centering
    \begin{tabular}{crrrrrrr} \toprule
    \multirow{2}{*}{Dataset}     & \multicolumn{7}{c}{\textbf{Languages}} \\\cmidrule{2-8}
         & \texttt{eng} & \texttt{hun} & \texttt{vie} & \texttt{pes} & \texttt{hin} & \texttt{srp} & \texttt{gla} \\ \midrule
    $C_{train}$  &   2.09 &  4.07  &  2.98  &  \cellcolor{gray!30}4.94  &  \cellcolor{gray!30}10.04 &  \cellcolor{gray!30}4.62  &  8.60  \\  
    $C_{tuning}$ &  2.03 & 3.97  & 2.88  & \cellcolor{gray!30}5.08  & \cellcolor{gray!30}10.58 & \cellcolor{gray!30}4.70  & 8.54  \\ \midrule
        & \texttt{mri} & \texttt{tpi} & \texttt{sot} & \texttt{amh} & \texttt{guj} & \texttt{ory} & \texttt{mya} \\ \midrule
    $C_{train}$ & 7.65  &  6.24  &  8.22  &  \cellcolor{gray!30}22.86  &  \cellcolor{gray!30}17.02  &  \cellcolor{gray!30}19.73  &  \cellcolor{gray!30}40.11 \\
    $C_{tuning}$ & 7.68  & 6.30  & 8.41  & \cellcolor{gray!30}23.12  & \cellcolor{gray!30}17.16  & \cellcolor{gray!30}19.70   & \cellcolor{gray!30}41.71 \\
    \bottomrule
     \end{tabular}
    \caption{Number of tokens~(in millions) in $C_{train}$ and $C_{tuning}$ for the 10k corpus size.}
    \label{tab:tokens_stats}
\end{table}

We mainly use the hyperparameter values defined in \citet{kaplan2025from} for Patchscopes, mapper training, and \calibration~(\cref{tab:hyperparameter_values}). 
\begin{table}[]
    \centering
    \begin{tabular}{lr} \toprule
    \textbf{Hyperparameter}     & \textbf{Value} \\ \toprule
    BPE Expansion Budgets    &  2$^i\times$ max. expansion budget; i=[-5,0] \\
    Model Quantization        &     fp16 \\
    Patchscopes Batch Size    &     16~(\patchscopes) , 2~(\swpatchscopes)\\
    Mapper Layer Batch Size   &     8 \\
    Tuning (LAPT) Batch Size  &     4 \\
    Tuning LR                 &     $10^{-4}$ \\
    \# Tuning Epochs          &     3   \\
    Tuning Schedule           &     Linear \\
    \# Warm-up Steps          &     3\% of tuning data size \\
    Optimizer                 &     AdamW   \\\bottomrule

    \end{tabular}
    \caption{Hyperparameter Values.}
    \label{tab:hyperparameter_values}
\end{table}

During \calibration, we only \calibrate the (un-)embeddings corresponding to the newly added tokens. Model weights remian unchanged. Every experiment was performed using either a single H200, H200~(3g.71gb MIG), or H200~(2g.35gb MIG).

\subsection{Prompt Templates}

We provide the generation~(second inference pass) template in C.1. For every language, we translate the phrase ``\textit{In language$_l$}'' in to corresponding languages~(e.g., ``In English:'' for English). SIB200 and Belebele evaluation prompts are given in C.2 and C.3.

\begin{promptbox}{Patchscopes Generation Prompt}

\medskip
\texttt{$<$In language \textit{l}$>_l$: X, X, X, X,}
\label{psc_gen_prompt}
\end{promptbox}

\begin{promptbox}{SIB200 Evaluation Prompt Template}
\label{prompt:sib200_prompt}
\medskip
\texttt{Topic Classification: science/technology, travel, politics, sports, health, entertainment, geography. Only respond with one of these categories.}\\
\texttt{\#\# Examples}\\

\texttt{Sentence: "$<$ex1$>$"}\\
\texttt{Topic: science/technology}\\

\texttt{Sentence: "$<$ex2$>$"}\\
\texttt{Topic: travel}\\

\texttt{Sentence: "$<$ex3$>$"}\\
\texttt{Topic: politics}\\

\texttt{Sentence: "$<$ex4$>$"}\\
\texttt{Topic: sports}\\

\texttt{Sentence: "$<$ex5$>$"}\\
\texttt{Topic: health}\\

\texttt{Sentence: "$<$ex6$>$"}\\
\texttt{Topic: entertainment}\\

\texttt{Sentence: "$<$ex7$>$"}\\
\texttt{Topic: geography}\\

\texttt{The topic of the news "$<$test\_ex$>$" is:}
    
\end{promptbox}

\begin{promptbox}{Belebele Evaluation Prompt Template}
\label{prompt:belebele_prompt}
\medskip
\texttt{Given the following passage, query, and answer choices, output only the letter corresponding to the correct answer. Do not add any explanation.}\\
\texttt{\#\#\#}\\ 
\texttt{Passage:}\\
\texttt{$<$passage$>$}\\
\texttt{\#\#\#}\\ 
\texttt{Query:}\\
\texttt{$<$query$>$}\\
\texttt{\#\#\#}\\
\texttt{Choices:}\\
\texttt{$($A$)$ $<$choice\_a$>$}\\
\texttt{$($B$)$ $<$choice\_b$>$}\\
\texttt{$($C$)$ $<$choice\_c$>$}\\
\texttt{$($D$)$ $<$choice\_d$>$}\\
\texttt{\#\#\#}\\
\texttt{Answer:}
\end{promptbox}

\subsection{Other Experimental Choices}
\label{sec:other_exp_choices}
\paragraph{Why both the training set and a separate \calibration~set for \calibration?} Using the training set for LAPT provides sufficient guarantees that some of the newly added vocabulary items are \calibrated. A separate \calibration dataset improves generalizability and increases the chance of training subwords that are subsumed by larger (sub-)words in the training set.

\paragraph{Which parameters to tune?} The input embeddings, output embeddings, and the tokenizer are the only vocabulary-dependent components in the model. As a result, one can minimally adapt the vocabulary-expanded model by \calibration just the \embedding corresponding to the newly added items. We follow this strategy and keep all other model parameters frozen during \calibration, and keep it fixed across all experiments. However, more sophisticated strategies have been studied  which may further improve the models considered~\citep{kim2024efficienteffectivevocabularyexpansion,yamaguchi2026howcan}.

\paragraph{How do we construct the Pareto front?} Here, we explain the details behind the Pareto fronts seen in Figures~\ref{fig:pareto_qwen3_30b} and \ref{fig:pareto_other_inits}. We start by defining the maximum expansion budget. For our experiments, we define the maximum expansion budget as the number of items successfully detokenized and added to the expanded vocabulary under the \patchscopes approach. Note that this number differs across languages, as the number of items added varies by language. A fair comparison for the BPE-based item selection approach would also allow the same number of added items as we had under \patchscopes. This is the $\gamma$=1 setting, where the sizes of the expanded vocabularies for both \patchscopes and BPE experiments are equivalent. This setting typically corresponds to the point with maximum efficiency on the front (the \textit{right-most} point). However, we need a Pareto front for the BPE experiments to show that even with reduced token efficiency (i.e., when the number of added items is less than the maximum budget), models trained with BPE-selected items perform worse on the performance-efficiency tradeoff. Hence, the $\gamma$ parameter allows us to restrict the size of the expanded vocabulary by taking different proportions of the maximum expansion budget. 
\section{Additional Results}
In this section, we present additional results to supplement our main findings.

\subsection{Augmenting Pareto Front Analysis with FVT Initialization}
\label{sec:augmenting_pareto_front_with_fvt}
\begin{figure}
    \centering
    \includegraphics[width=0.5\linewidth]{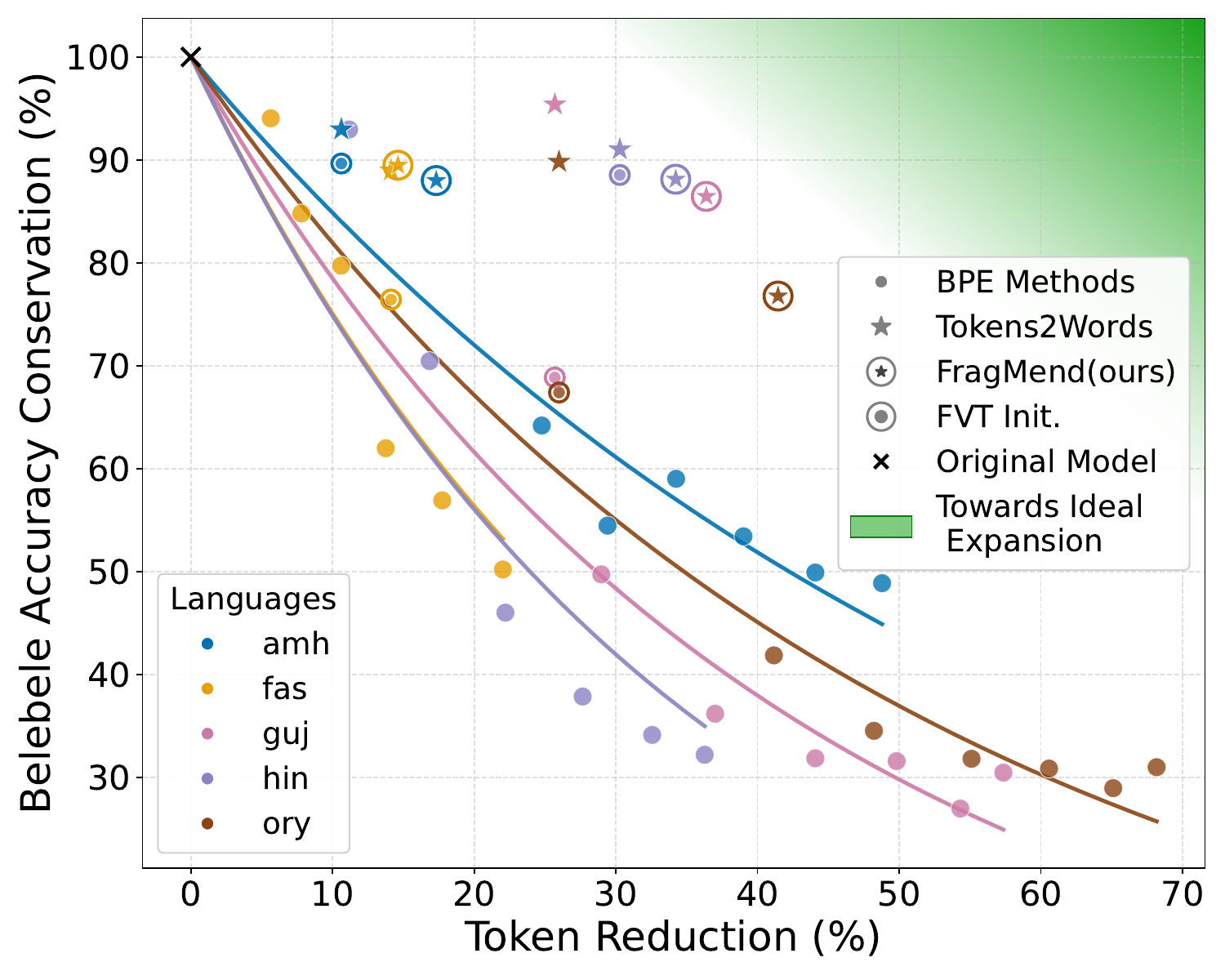}
    \caption{Pareto Analysis with additional points corresponding to models expanded using \patchscopes-based item selection and FVT-based initialization. These points are marked by the `o-dot' symbol~($\odot$). }
    \label{fig:pareto_front_with_fvt_init}
\end{figure}

Figure~\ref{fig:pareto_front_with_fvt_init} presents Figure~\ref{fig:pareto_qwen3_30b} with added points corresponding to expanded models where \patchscopes is used item selection, and FVT is used for embedding initialization. We observe that even with the FVT initialization, the models expanded using interpretability-based item selection methods outperform the BPE-defined Pareto constructed using the same initialization. This result consolidates interpretability-based item selection as the superior item selection approach. 

\subsection{Constructing Pareto Front With Other Initializations}
\label{sec:additional_pareto_results}
\begin{figure*}[h]
    \centering
    \begin{subfigure}{0.49\textwidth}
        \centering
        \includegraphics[width=\textwidth]{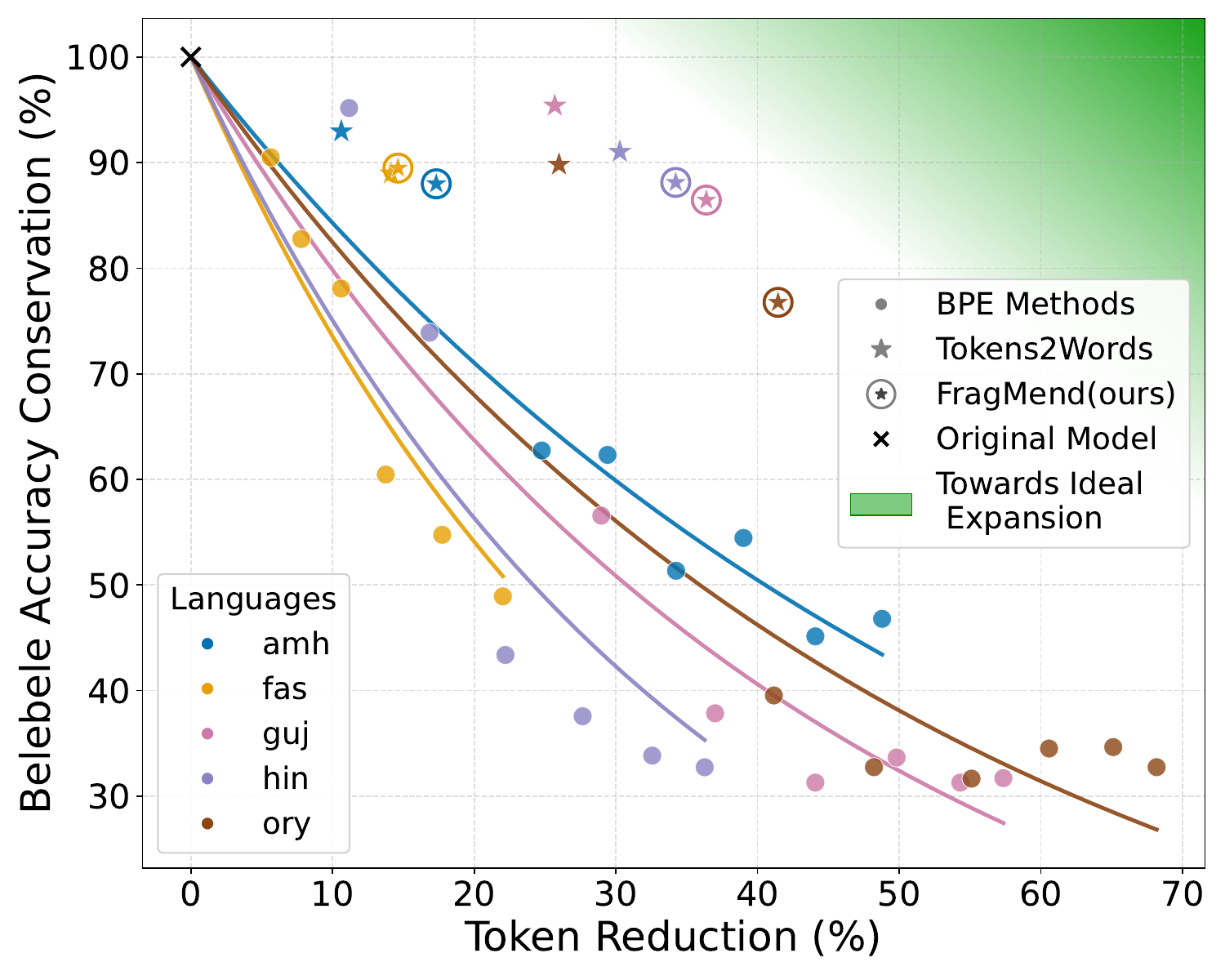}
        \caption{}
        \label{fig:pareto_qwen3_30b_random}
    \end{subfigure}%
    ~
    \begin{subfigure}{0.49\textwidth}
        \centering
        \includegraphics[width=\textwidth]{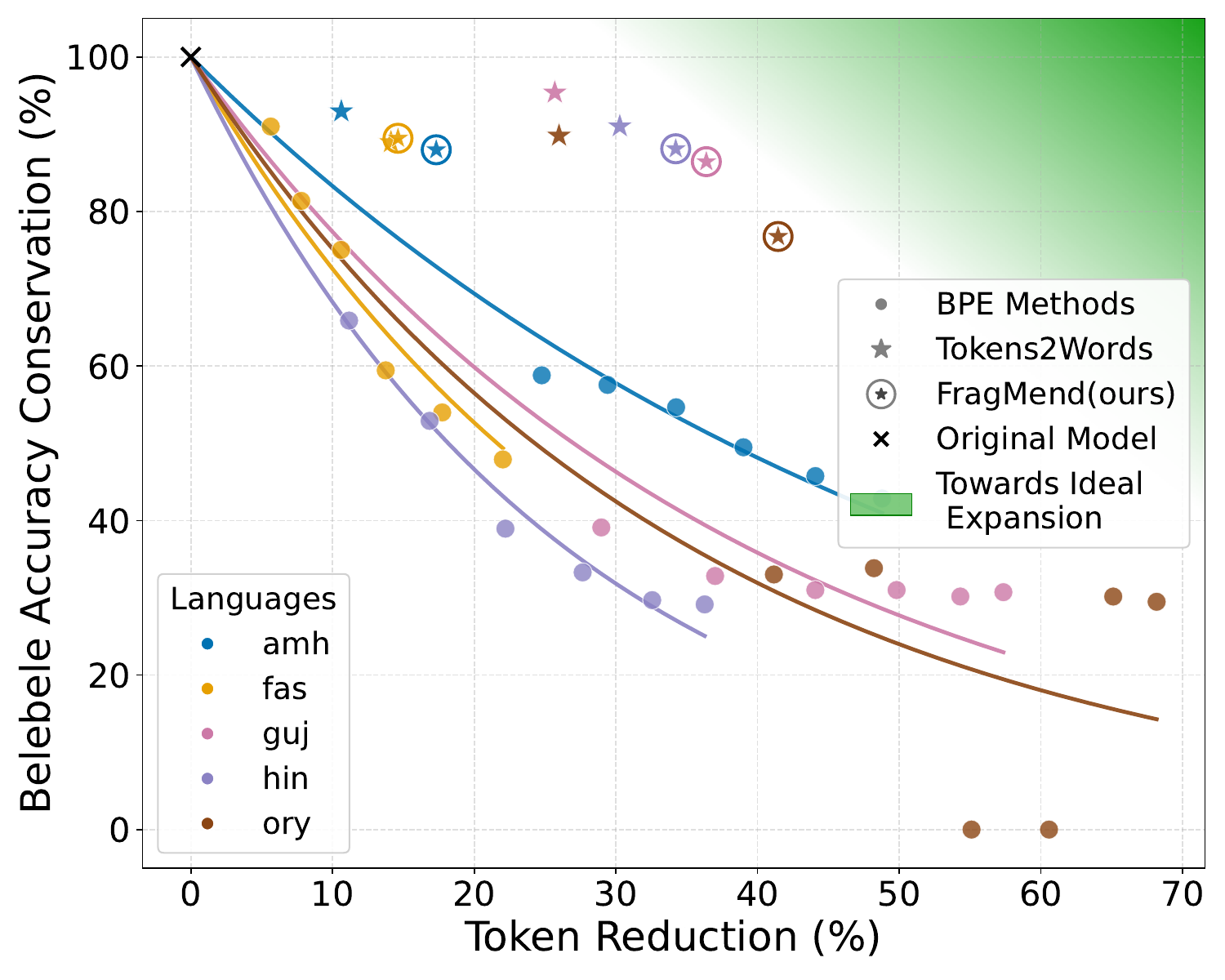}
        \caption{}
        \label{fig:pareto_qwen3_30b_focus}
    \end{subfigure}
    \caption{Pareto analysis of BPE methods as compared to the interpretability-based method with (a) Random and (b) FOCUS initialization for the BPE-added items. All Pareto analyses are on $|C_{train}|=$1k sequences. }
    \label{fig:pareto_other_inits}
    
\end{figure*}

In \cref{fig:pareto_other_inits}, we show the pareto analysis with different \embedding initializations for the items added via BPE. Note that unlike Figure~\ref{fig:pareto_front_with_fvt_init}, here the initialization used to construct the Pareto front is different. The plot largely follows the trends observed in \cref{fig:pareto_qwen3_30b}, with interpretability-based methods consistently advancing the BPE Pareto front. This result reinforces our finding as initialization-independent, at least among the set of methods considered.

\subsection{BPE vs \patchscopes Detailed Comparison}
\begin{figure*}[h]
    \centering
    \begin{subfigure}[b]{0.49\textwidth}
        %\centering
        \includegraphics[width=\textwidth]{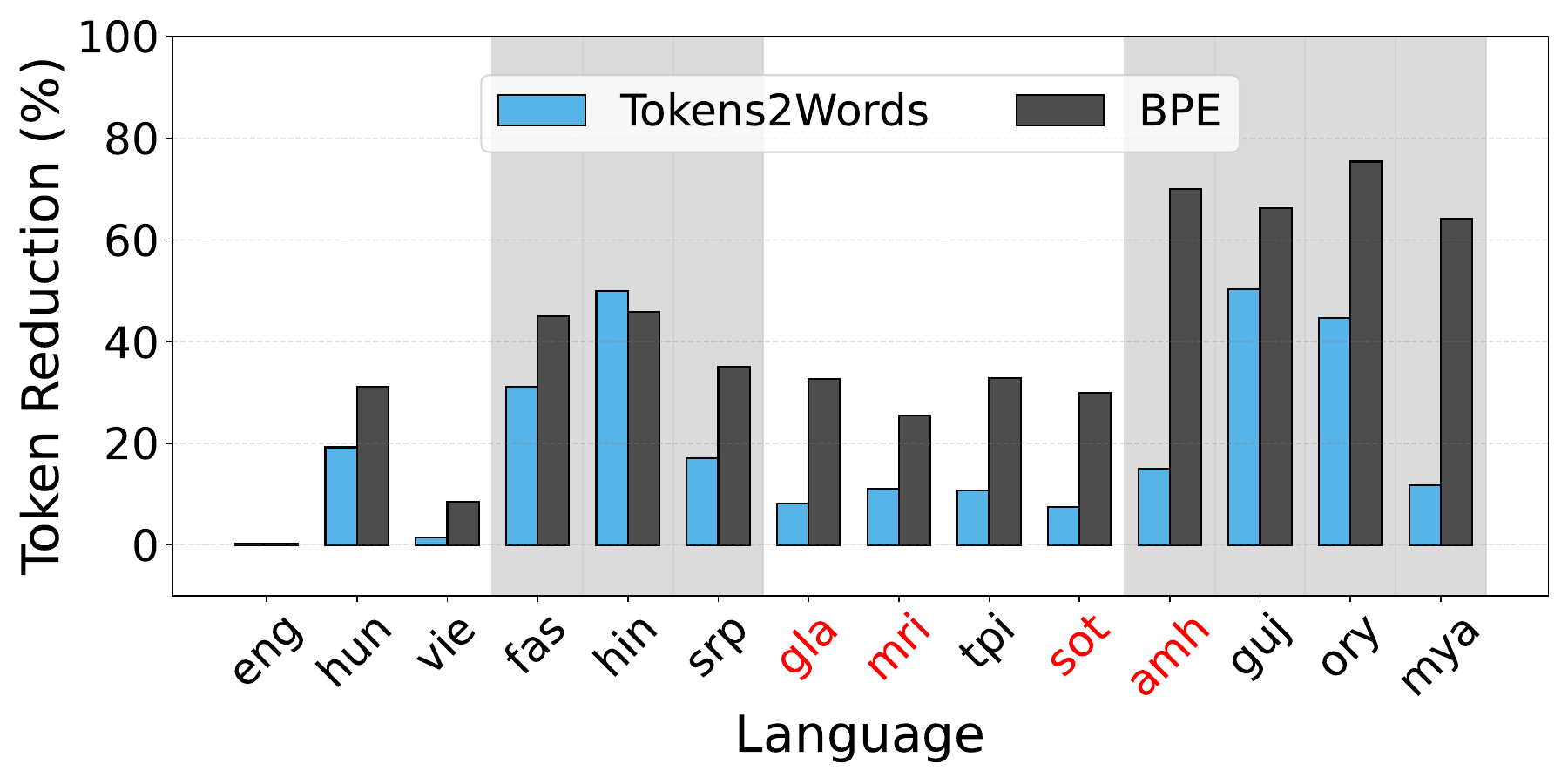}
        \caption{}
        \label{fig:success_rate_bpevspsc}
    \end{subfigure}
    \hfill
    \begin{subfigure}[b]{0.49\textwidth}
        %\centering
        \includegraphics[width=\textwidth]{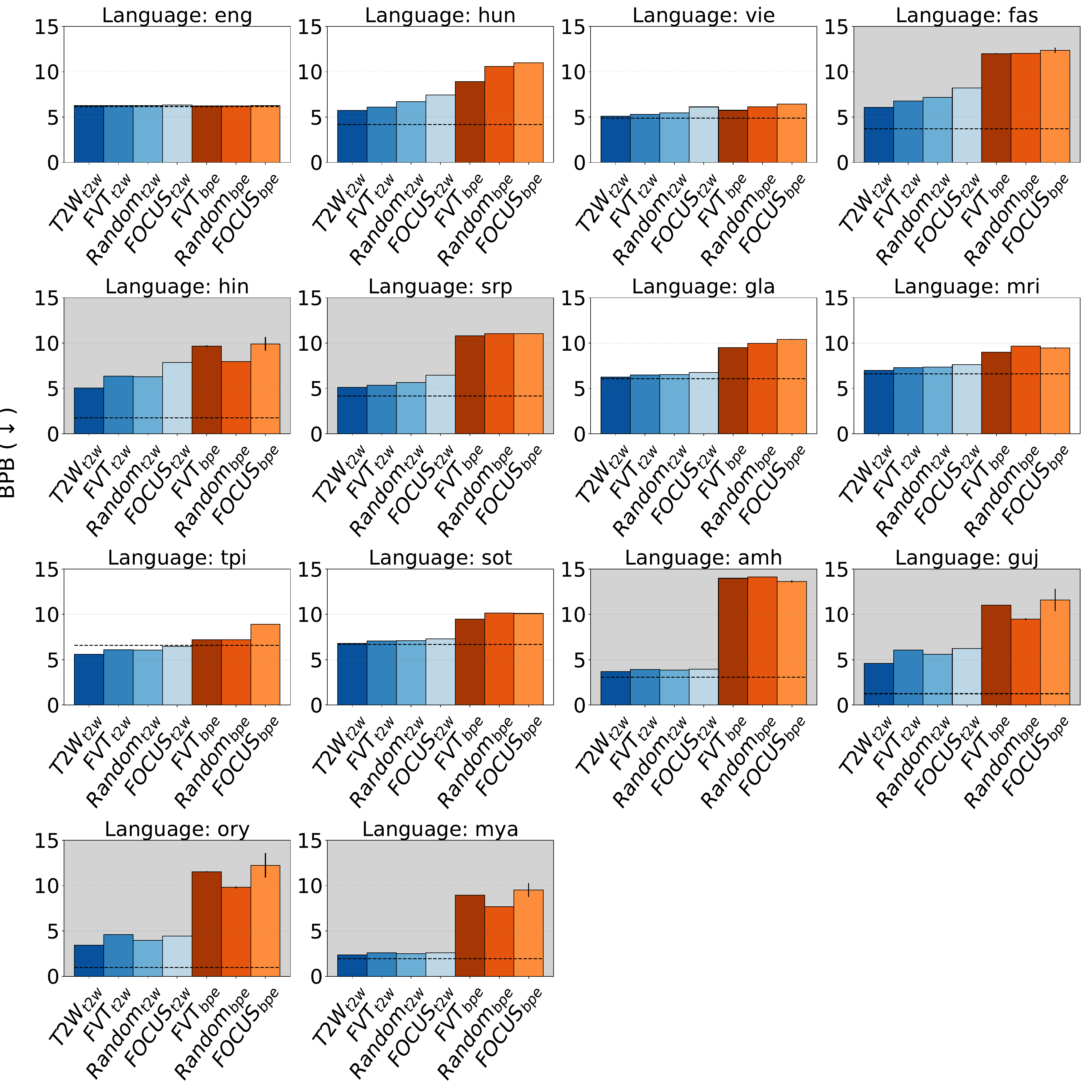}
        \caption{}
        \label{fig:bpevspsc_bpb}
    \end{subfigure}
  
    \vspace{1cm}
    \begin{subfigure}[b]{0.49\textwidth}
        %\centering
        \includegraphics[width=\textwidth]{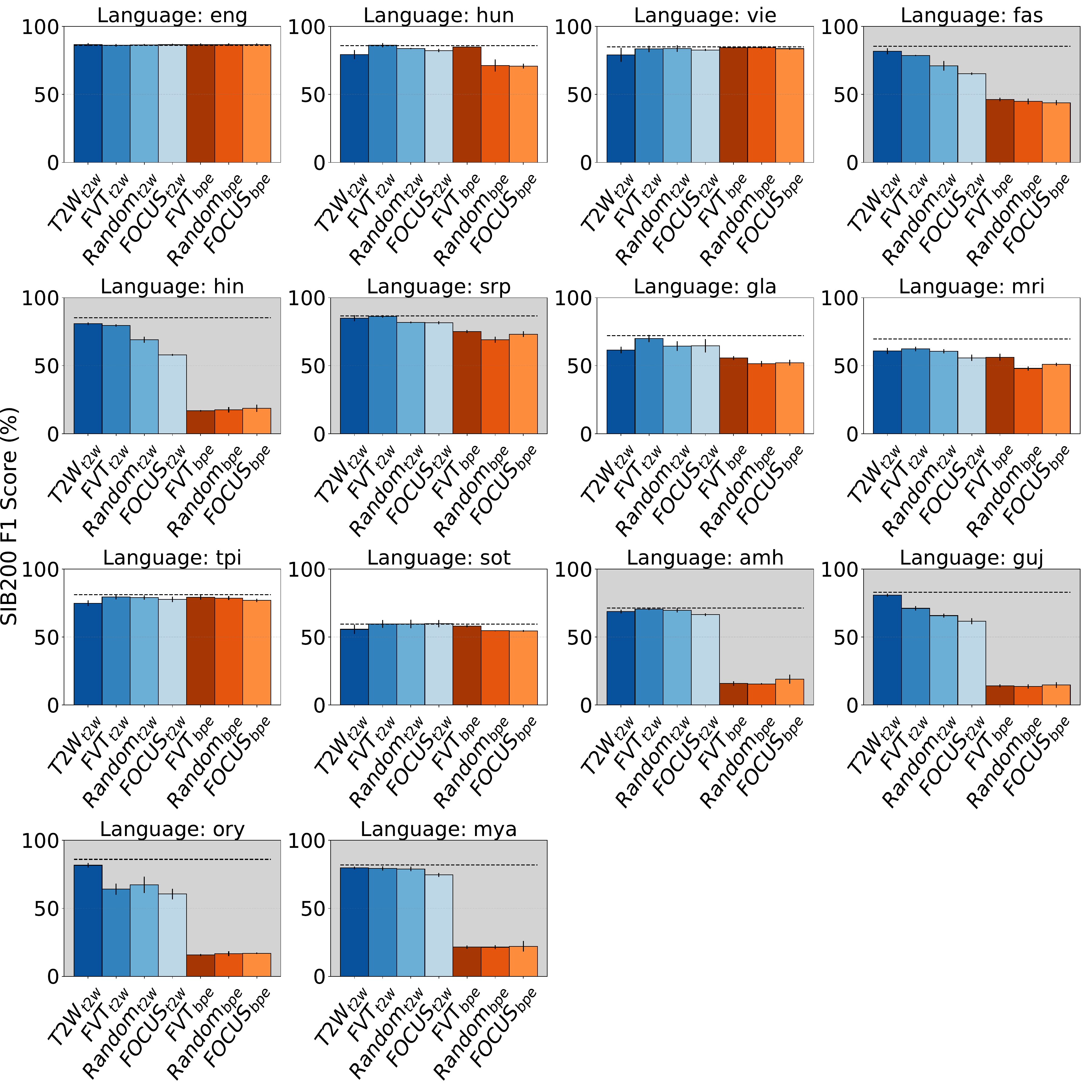}
        \caption{}
        \label{fig:bpevspsc_sib200}
    \end{subfigure}
    \hfill
    \begin{subfigure}[b]{0.49\textwidth}
        %\centering
        \includegraphics[width=\textwidth]{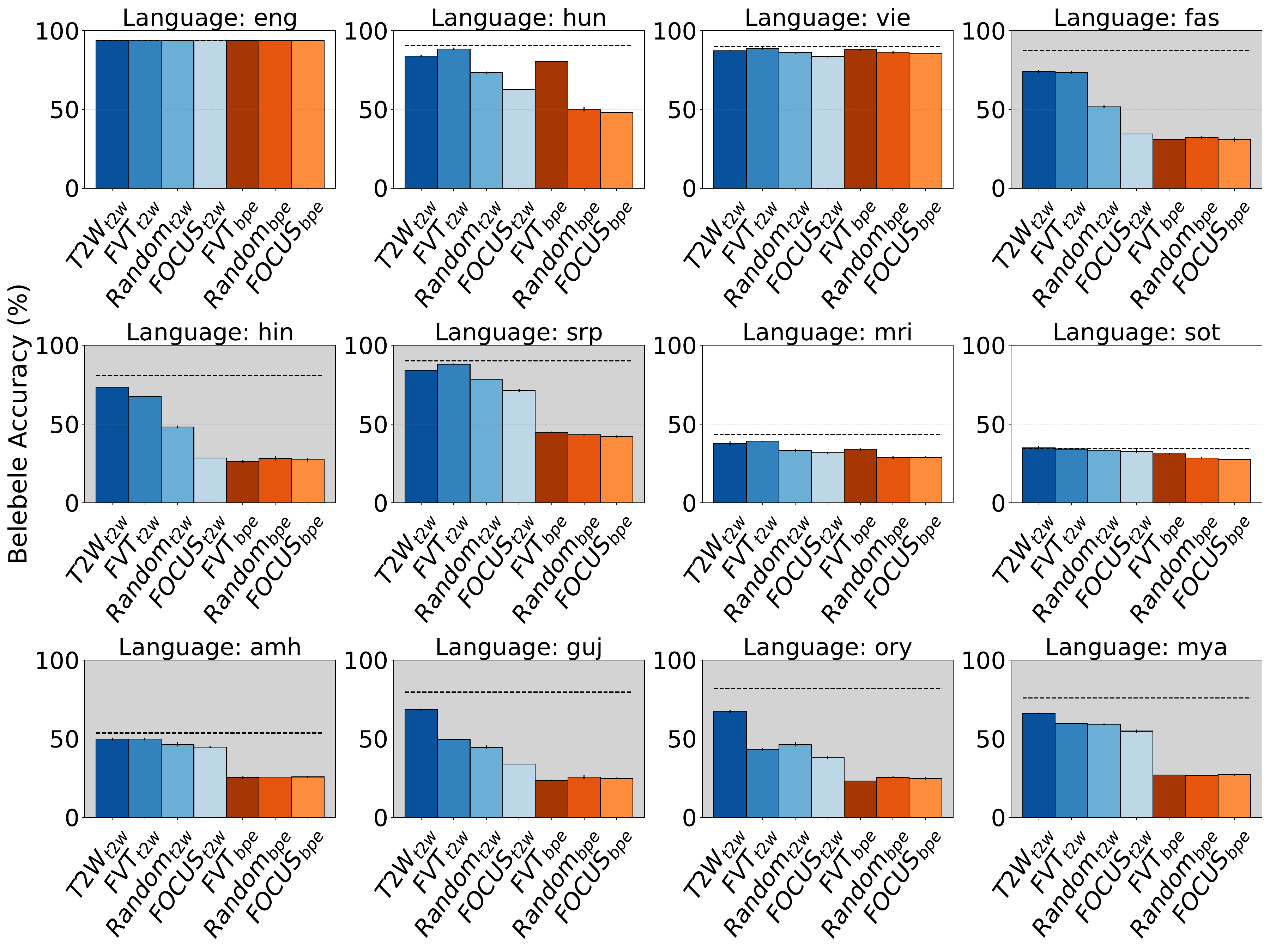}
        \caption{}
        \label{fig:bpevspsc_belebele}
    \end{subfigure}
    \caption{BPE-based~(subscript '\textit{bpe}') vs. \patchscopes-based~(subscript '\textit{t2w}') vocabulary expansion: (a) token-reduction rate, and (b-d) performance comparisons with different initializations. Names indicate initialization methods, subscripts indicate the method to choose expansion items. Dotted line indicates original model performance. $|C_{train}|=$10k sequences. While BPE provides high efficiency gains, the performance drops are significant~(almost rendering models corresponding to low-resource non-Latin-script languages useless)}
    \label{fig:bpe_vs_psc}
    
\end{figure*}

\cref{fig:bpe_vs_psc} shows the comparison between interpretability-based vocabulary expansion~(\patchscopes) and BPE-driven vocabulary expansion given the same expansion budget. While BPE can provide higher efficiency gains, it can lead to significant performance drops (as high as 60 points; e.g., see SIB200 for \texttt{ory}). For Hindi~(\texttt{hin}), we observe that \patchscopes outperforms BPE on both the efficiency and performance dimensions. These results support our thesis from the Pareto analysis that interpretability-based methods offer a superior efficiency-performance tradeoff.

\subsection{\patchscopes: Detokenization Success Rate and Token Reduction}
\label{sec:detok_successrate_psc}
\begin{figure*}[h]
    \centering
    \begin{subfigure}{0.75\textwidth}
        \centering
        \includegraphics[width=\textwidth]{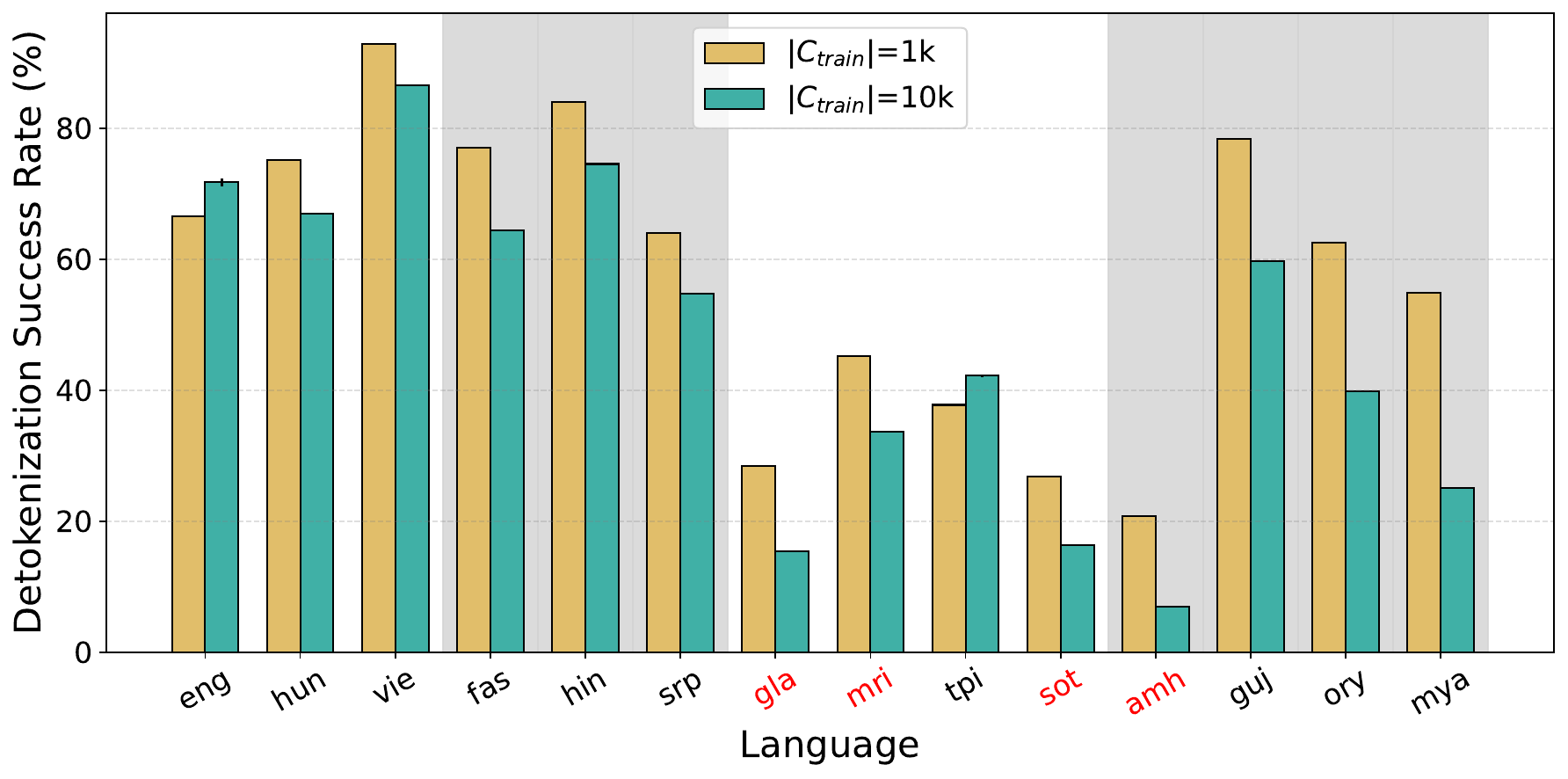}
        \caption{}
        \label{fig:detok_success_rate_across_corpus_qwen3_30b}
    \end{subfigure}
    
    \begin{subfigure}{0.75\textwidth}
        \centering
        \includegraphics[width=\textwidth]{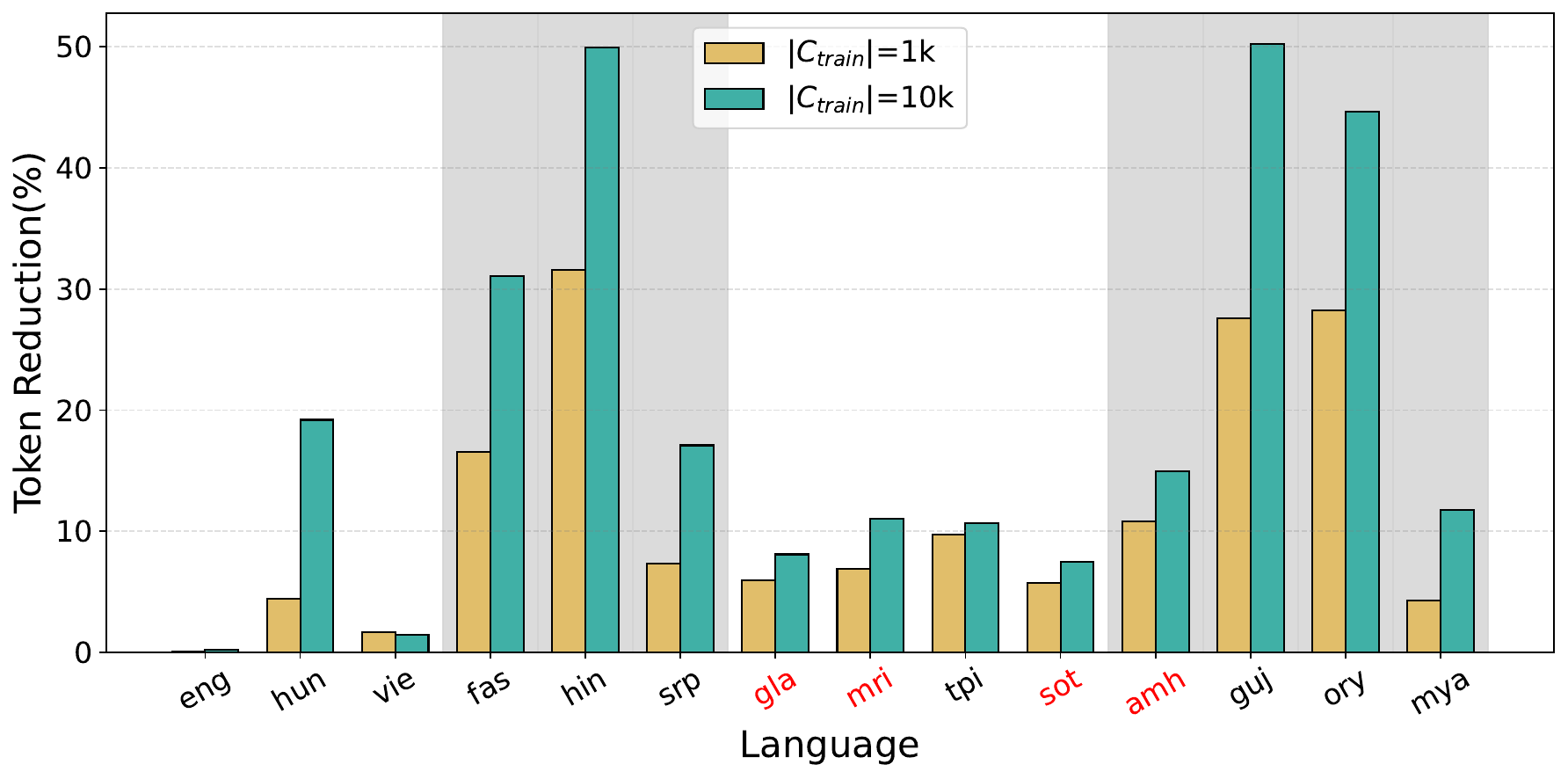}
        \caption{}
        \label{fig:tok_reduction_across_corpus_qwen3_30b}
    \end{subfigure}
    \caption{Qwen3-30B-A3B (a) detokenization success rate~ and (b) token reduction when expanding the model vocabulary with successfully detokenized words. Languages in `red' are not officially supported by the model. Languages written in a non-Latin script are in a gray background.}
    \label{fig:detok_tok_red_qwen3_30b}
\end{figure*}

\begin{figure*}[h]
    \centering
    \begin{subfigure}{0.75\textwidth}
        \centering
        \includegraphics[width=\textwidth]{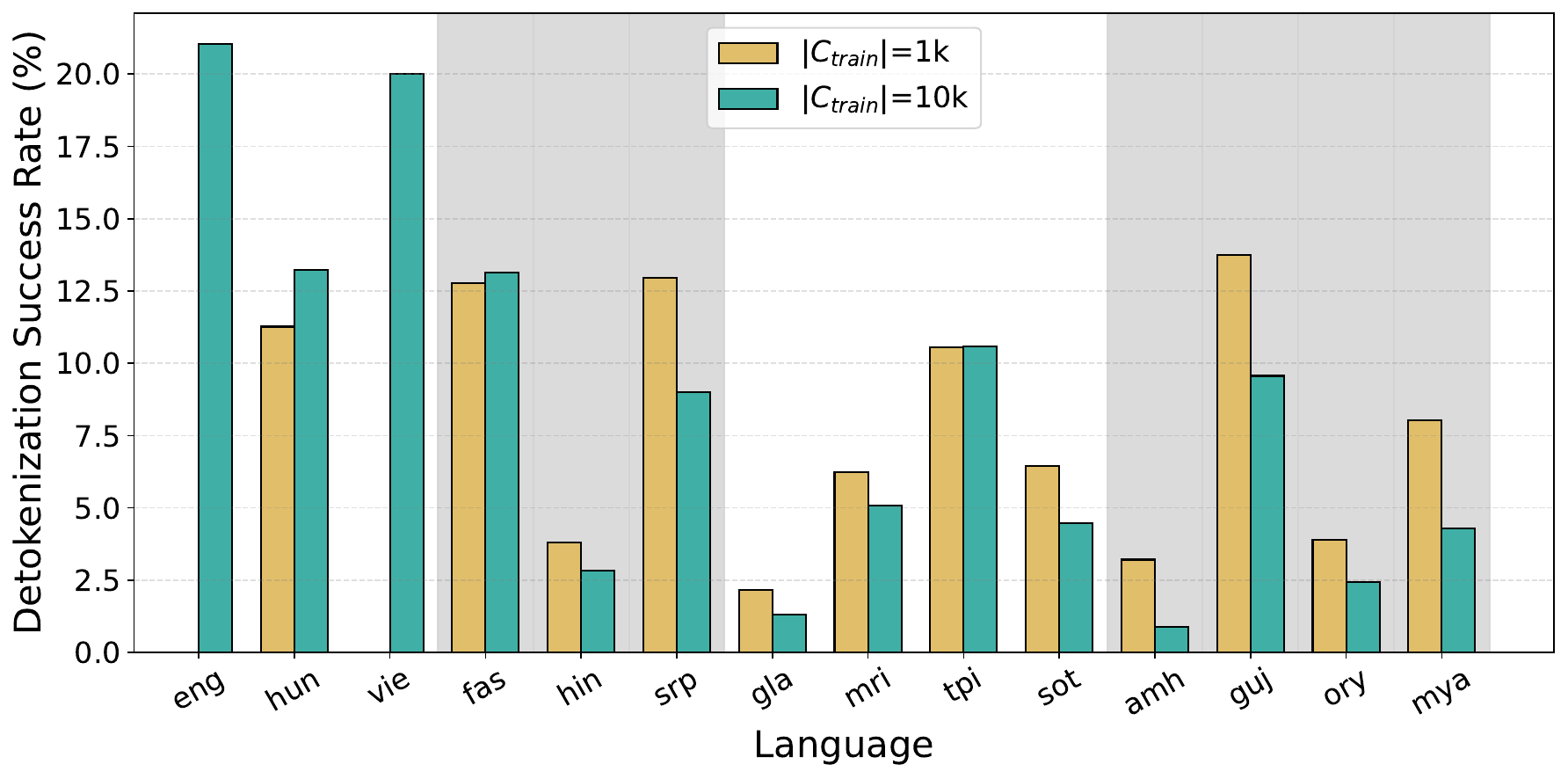}
        \caption{}
        \label{fig:detok_success_rate_across_corpus_qwen35_4b}
    \end{subfigure}
    
    \begin{subfigure}{0.75\textwidth}
        \centering
        \includegraphics[width=\textwidth]{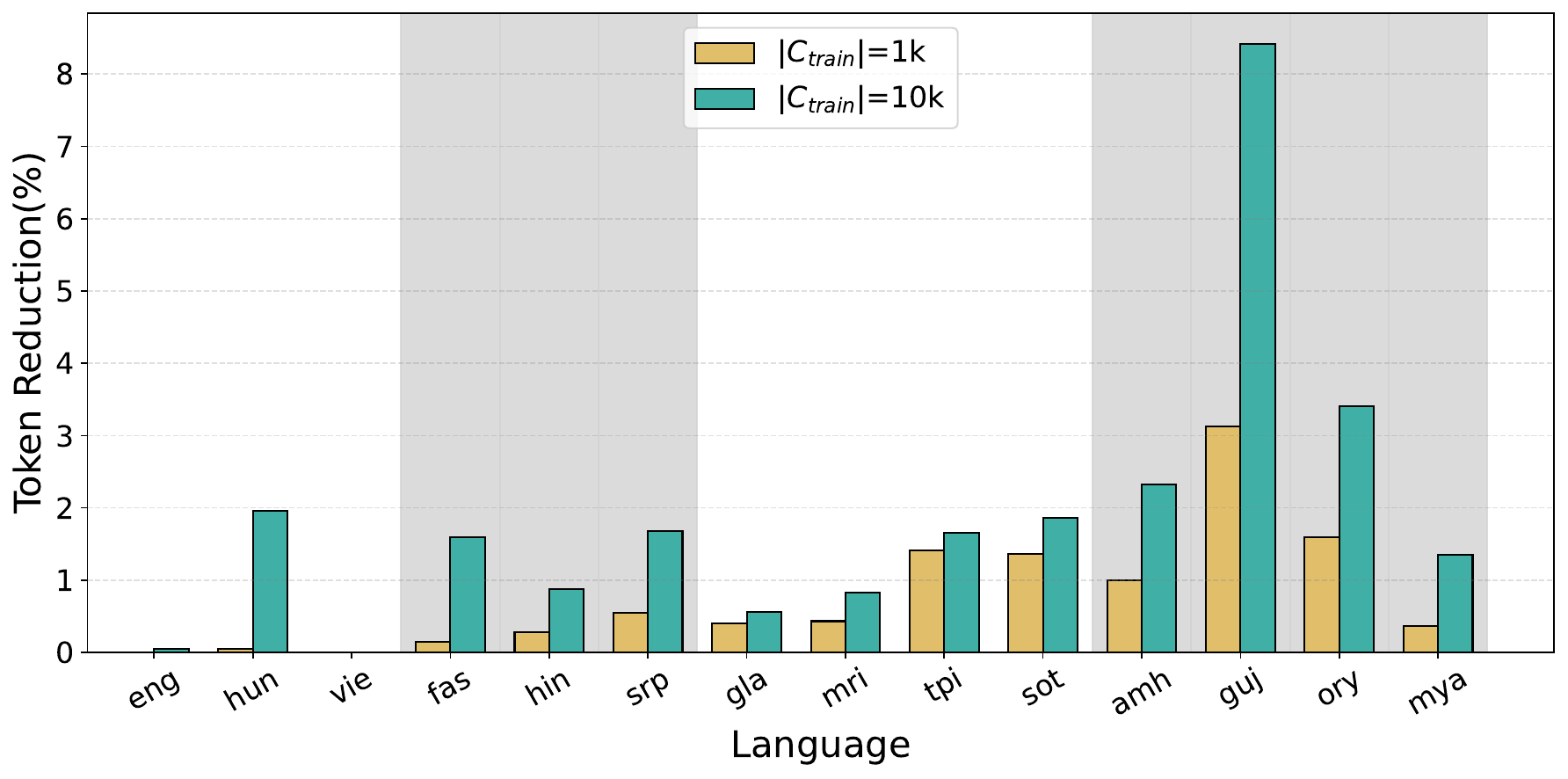}
        \caption{}
        \label{fig:tok_reduction_across_corpus_qwen35_4b}
    \end{subfigure}
    \caption{Qwen3.5-4B (a) detokenization success rate~ and (b) token reduction when expanding the model vocabulary with successfully detokenized words. Languages written in a non-Latin script are in a gray background.}
    \label{fig:detok_tok_red_qwen35_4b}
\end{figure*}

\begin{figure*}[h]
    \centering
    \begin{subfigure}{0.75\textwidth}
        \centering
        \includegraphics[width=\textwidth]{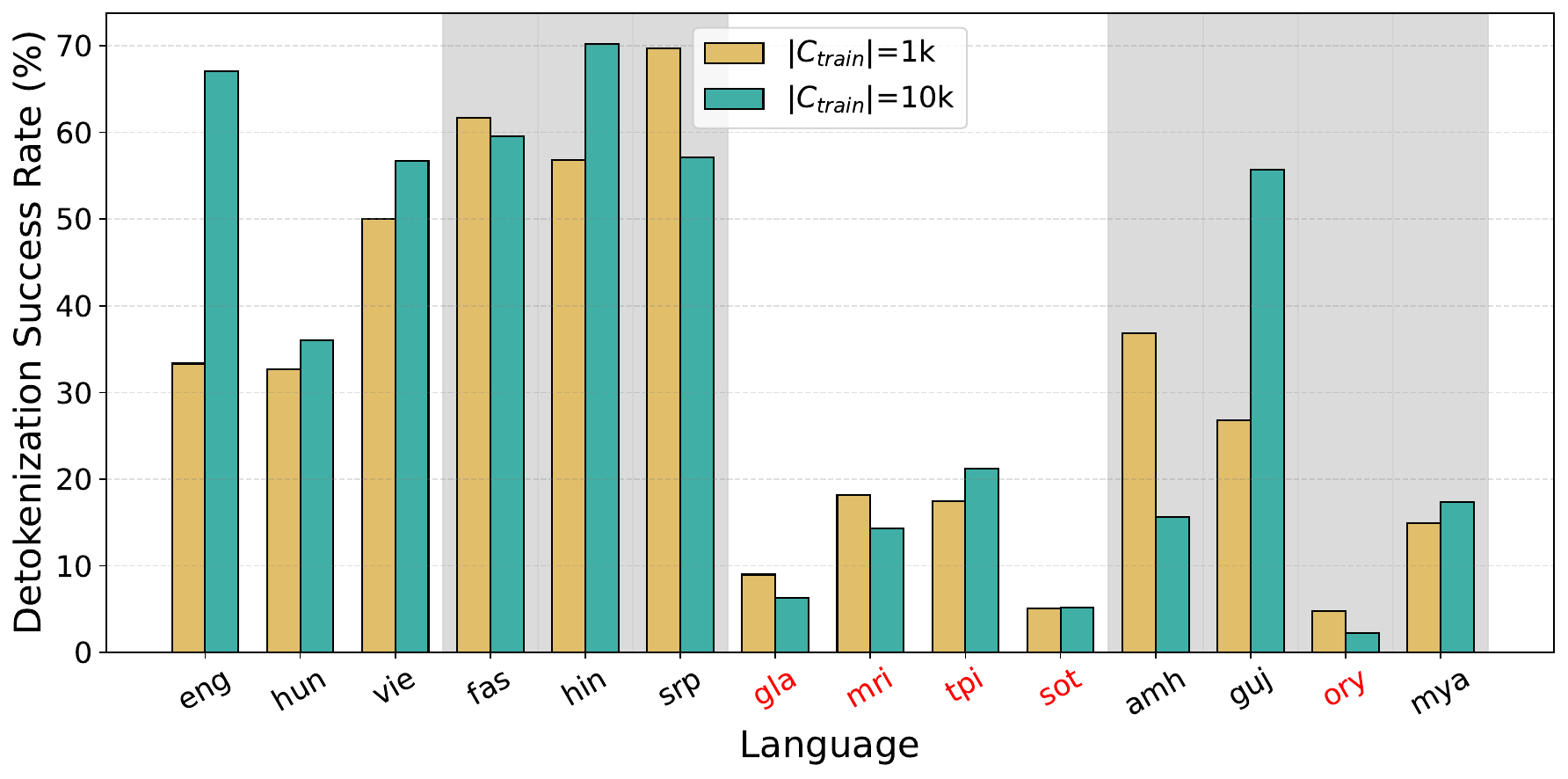}
        \caption{}
        \label{fig:detok_success_rate_across_corpus_tinyaya}
    \end{subfigure}
    
    \begin{subfigure}{0.75\textwidth}
        \centering
        \includegraphics[width=\textwidth]{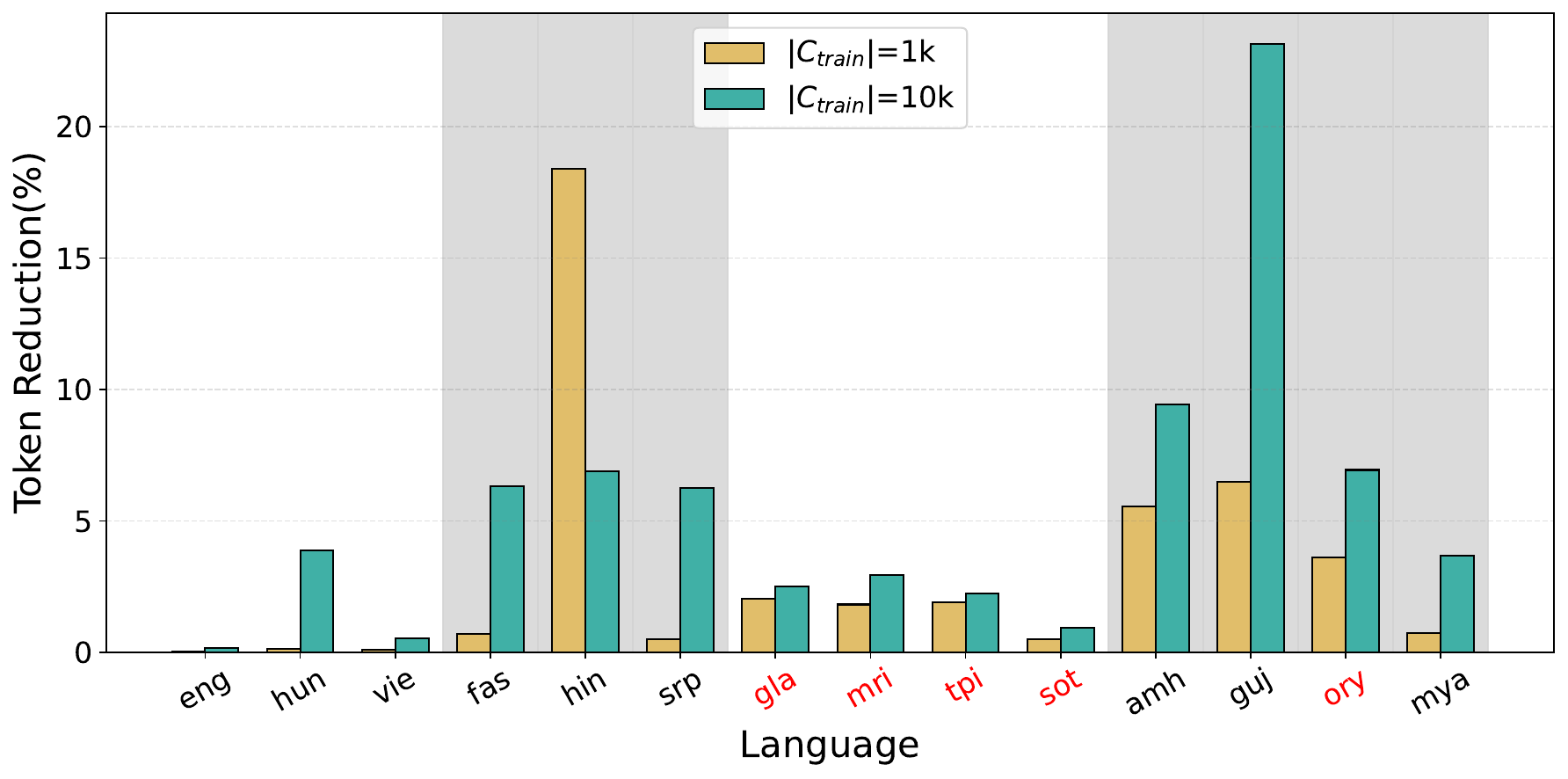}
        \caption{}
        \label{fig:tok_reduction_across_corpus_tinyaya}
    \end{subfigure}
    \caption{TinyAya-Global (a) detokenization success rate~ and (b) token reduction when expanding the model vocabulary with successfully detokenized words. Languages in `red' are not officially supported by the model. Languages written in a non-Latin script are in a gray background.}
    \label{fig:detok_tok_red_tinyaya}
\end{figure*}

Figures \ref{fig:detok_tok_red_qwen3_30b}, \ref{fig:detok_tok_red_qwen35_4b}, and \ref{fig:detok_tok_red_tinyaya} show detokenization success rates for the \patchscopes method and token reduction across the two corpus sizes for Qwen3-30B-A3B, Qwen3.5-4B, and TinyAya-Global, respectively. Detokenization success rate is the proportion of candidate words successfully detokenized, per the condition described in \S\ref{sec:background_addition}. The general trend shows a higher detokenization rate for the smaller corpus, as a larger corpus implies more heavy-tailed entities that are harder to detokenize. Token reduction is generally higher with larger corpus as more number of vocabulary items are added with a larger corpus. 

\subsection{\patchscopes: Initialization Comparison Additional Results}
\label{sec:intialization_comparison_results}
This section presents additional results for comparing the \patchscopes initialization method as compared to the baseline initialization.
\subsubsection{BPB Results: Qwen3-30B-A3B}
\begin{figure*}[h]
    \centering
    \begin{subfigure}{0.49\textwidth}
        \centering
        \includegraphics[width=\textwidth]{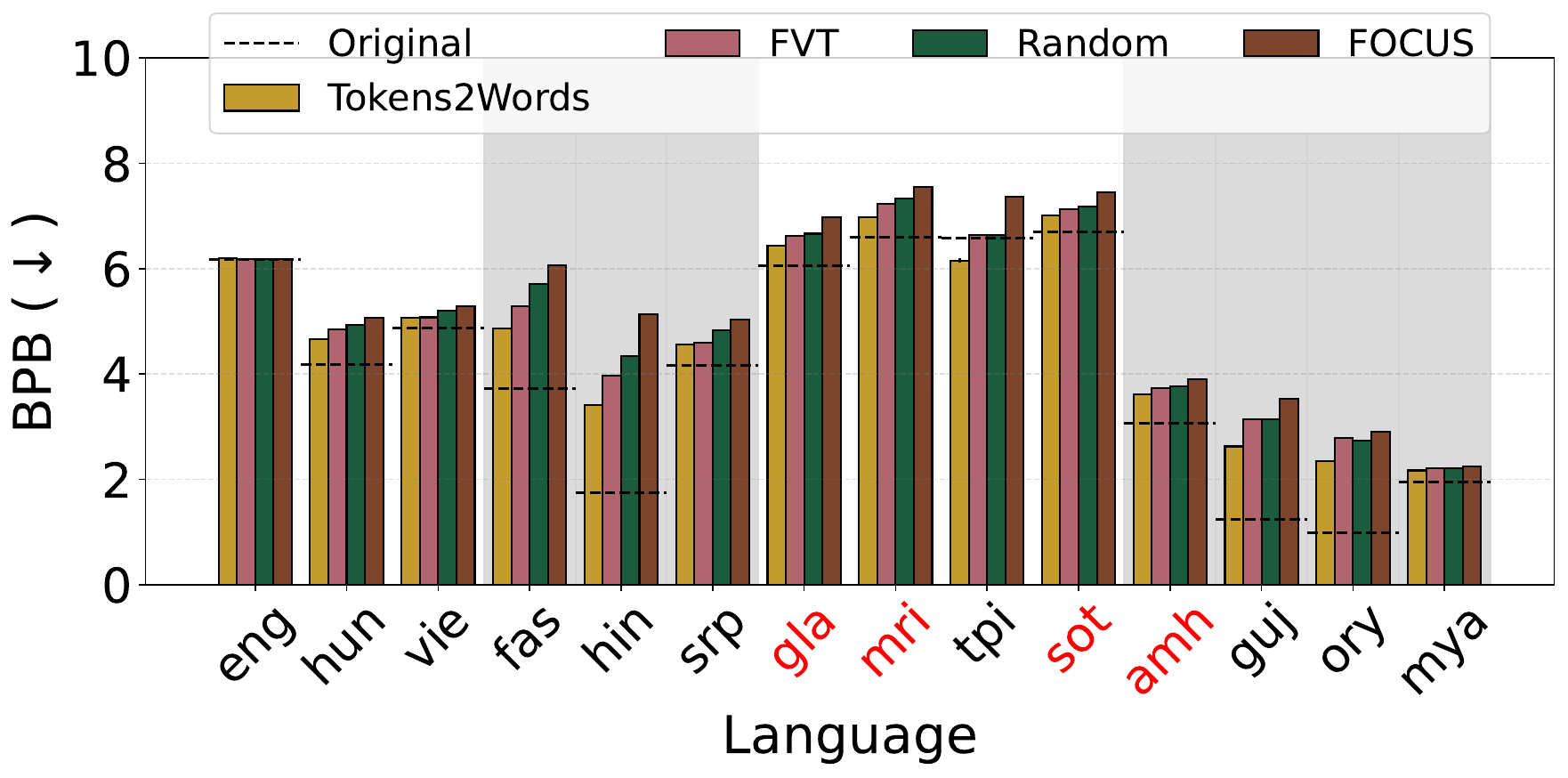}
        \caption{}
        \label{fig:bpb_1k}
    \end{subfigure}%
    ~
    \begin{subfigure}{0.49\textwidth}
        \centering
        \includegraphics[width=\textwidth]{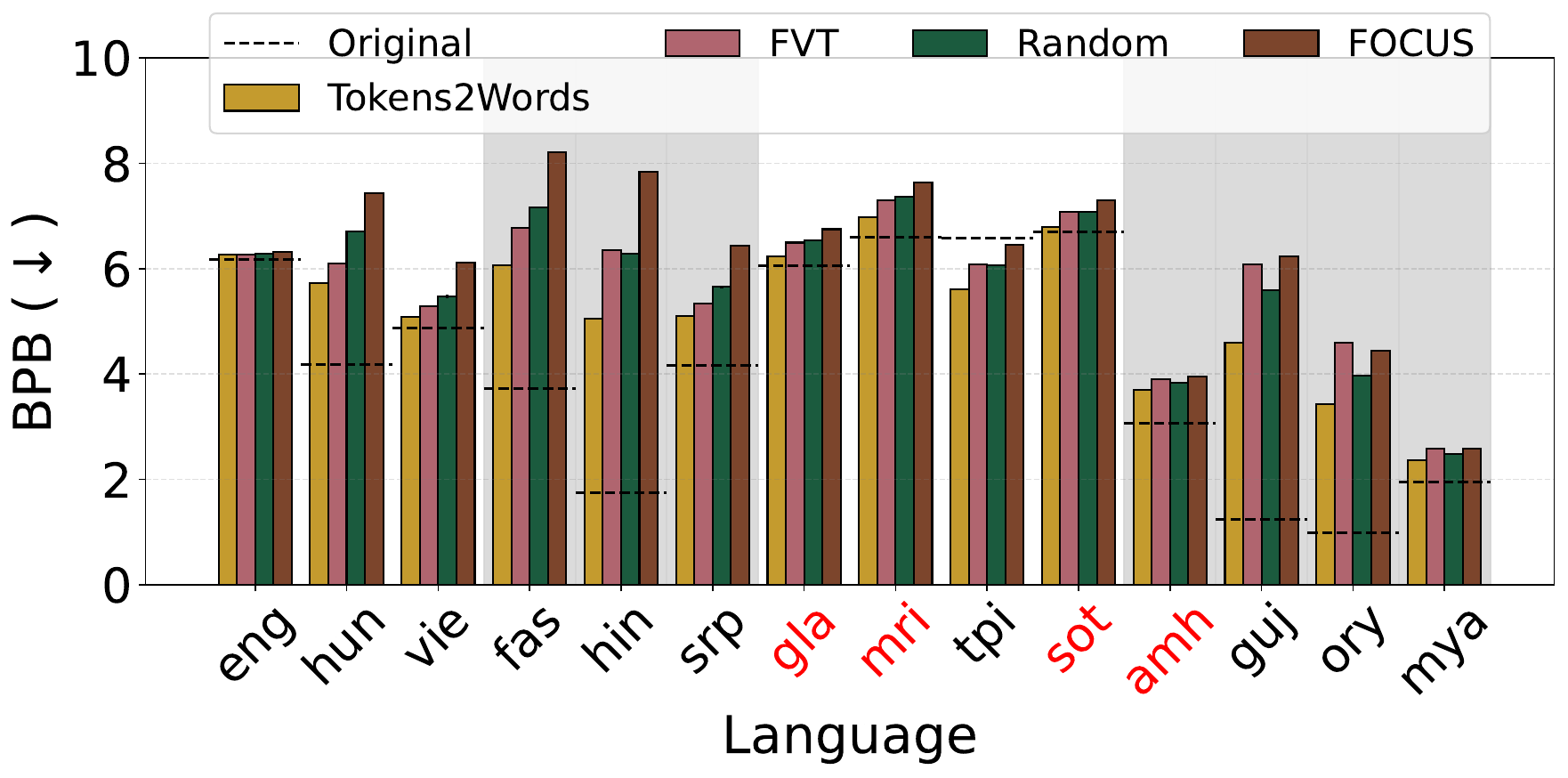}
        \caption{}
        \label{fig:bpb_10k}
    \end{subfigure}
    \caption{BPB performance across different initializations for Qwen3-30B results with corpus sizes: (a) $|C_{train}|=$1k, and (b) $|C_{train}|=$10k sequences. }
    \label{fig:qwen3_30b_bpb_init_comparison}
\end{figure*}

\cref{fig:qwen3_30b_bpb_init_comparison} shows the BPB performance on the Glot500c test set for different initializations across the two corpus sizes. For almost all the languages across the two sizes, \patchscopes shows the best performance as compared to all other baseline initializations. We also observe that the BPB performance degrades when we move to a larger corpus, following the efficiency-performance tradeoff.

\subsubsection{SIB200 and Belebele Results}
\begin{figure*}[h]
    \centering
    \begin{subfigure}{0.49\textwidth}
        \centering
        \includegraphics[width=\textwidth]{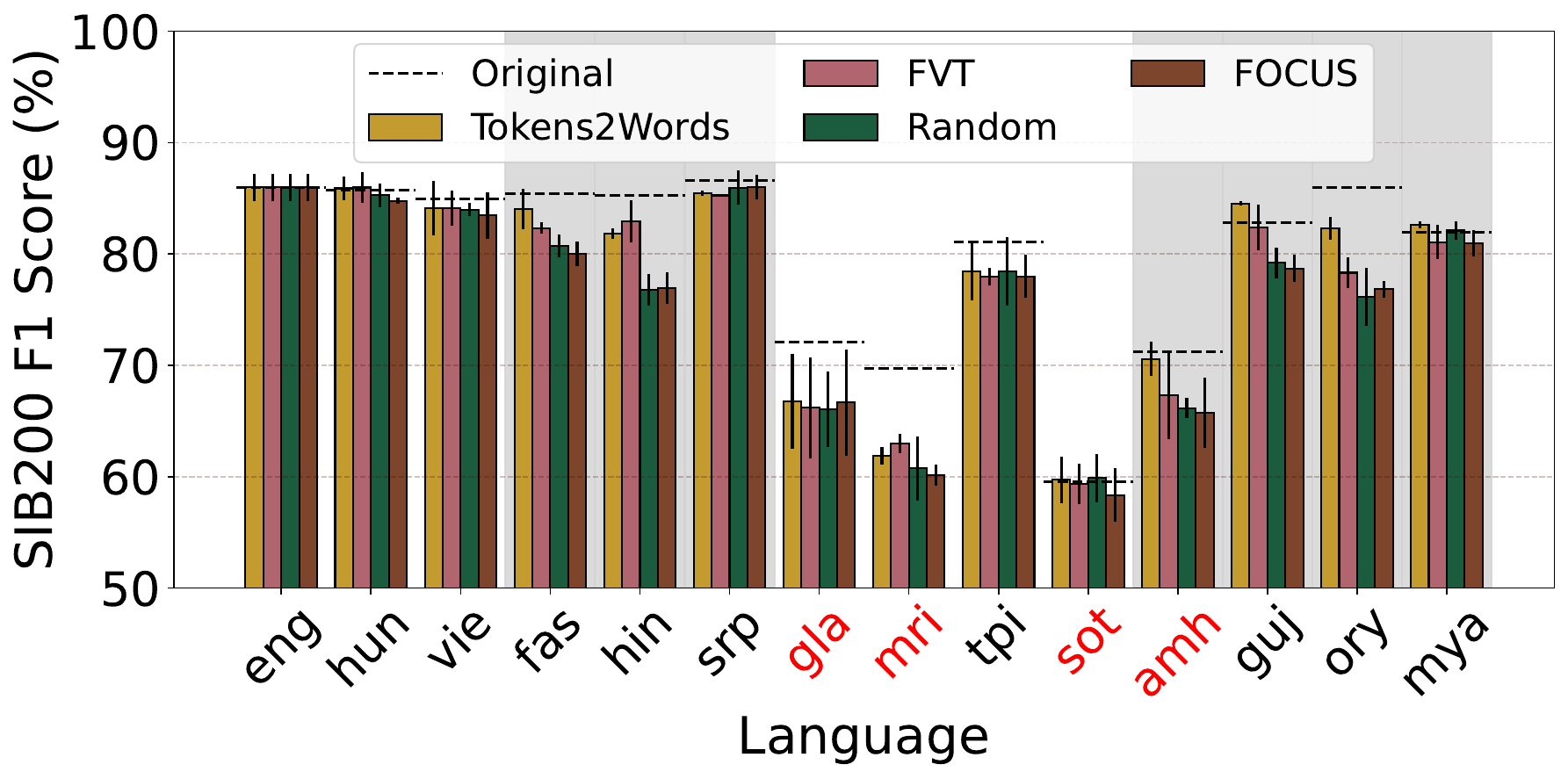}
        \caption{}
        \label{fig:sib200_1k_qwen3_30b}
    \end{subfigure}%
    ~
    \begin{subfigure}{0.49\textwidth}
        \centering
        \includegraphics[width=\textwidth]{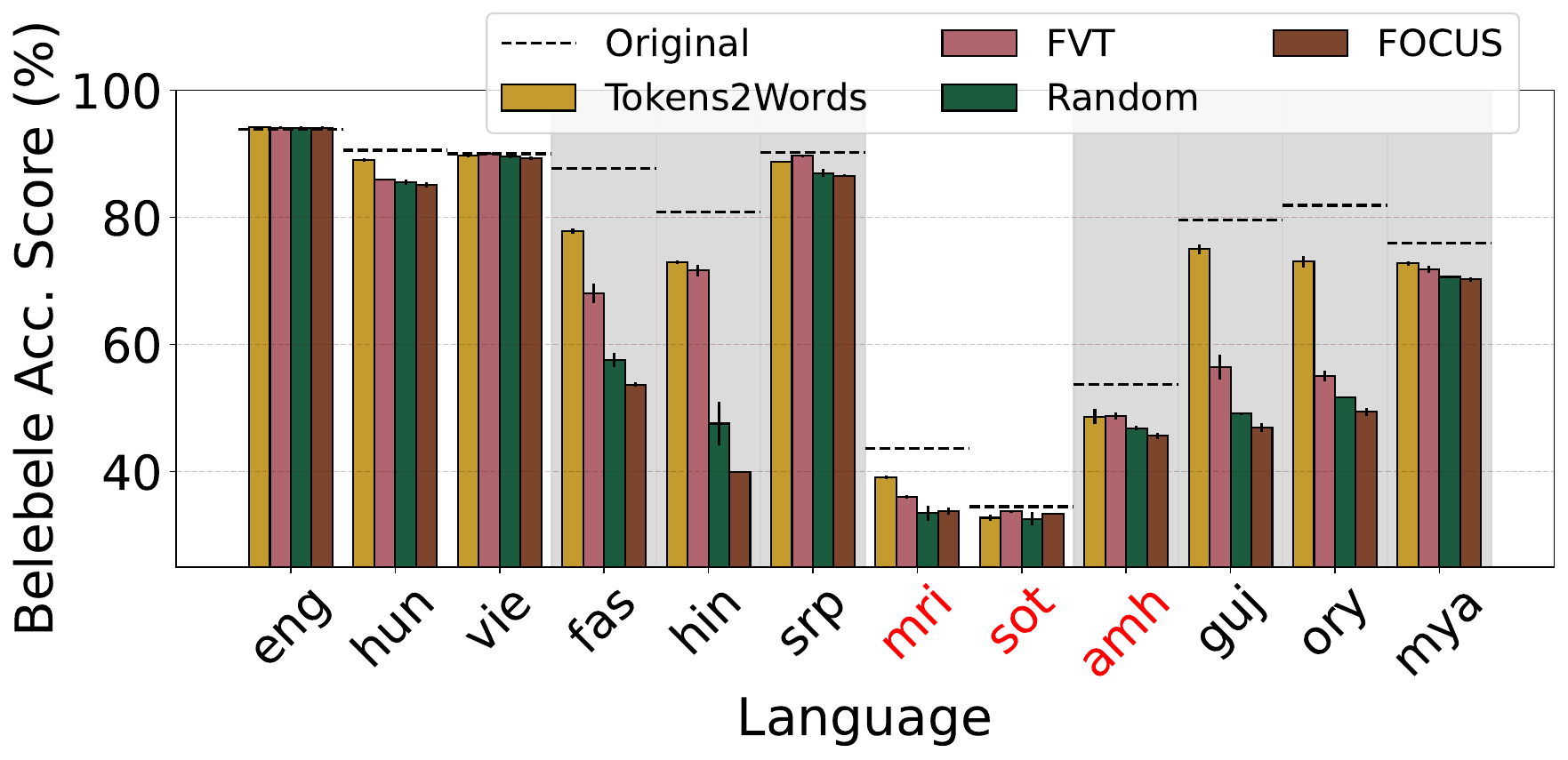}
        \caption{}
        \label{fig:belebele_1k_qwen3_30b}
    \end{subfigure}
    \caption{Performance across initialization methods on (\textit{a}) SIB200 and (\textit{b}) Belebele benchmarks. Model: Qwen3-30B results, $|C_{train}|=$1k.}
    \label{fig:qwen3_30b_1k_disc_init_comparison}
\end{figure*}

\begin{figure*}[h]
    \centering
    \begin{subfigure}{0.49\textwidth}
        \centering
        \includegraphics[width=\textwidth]{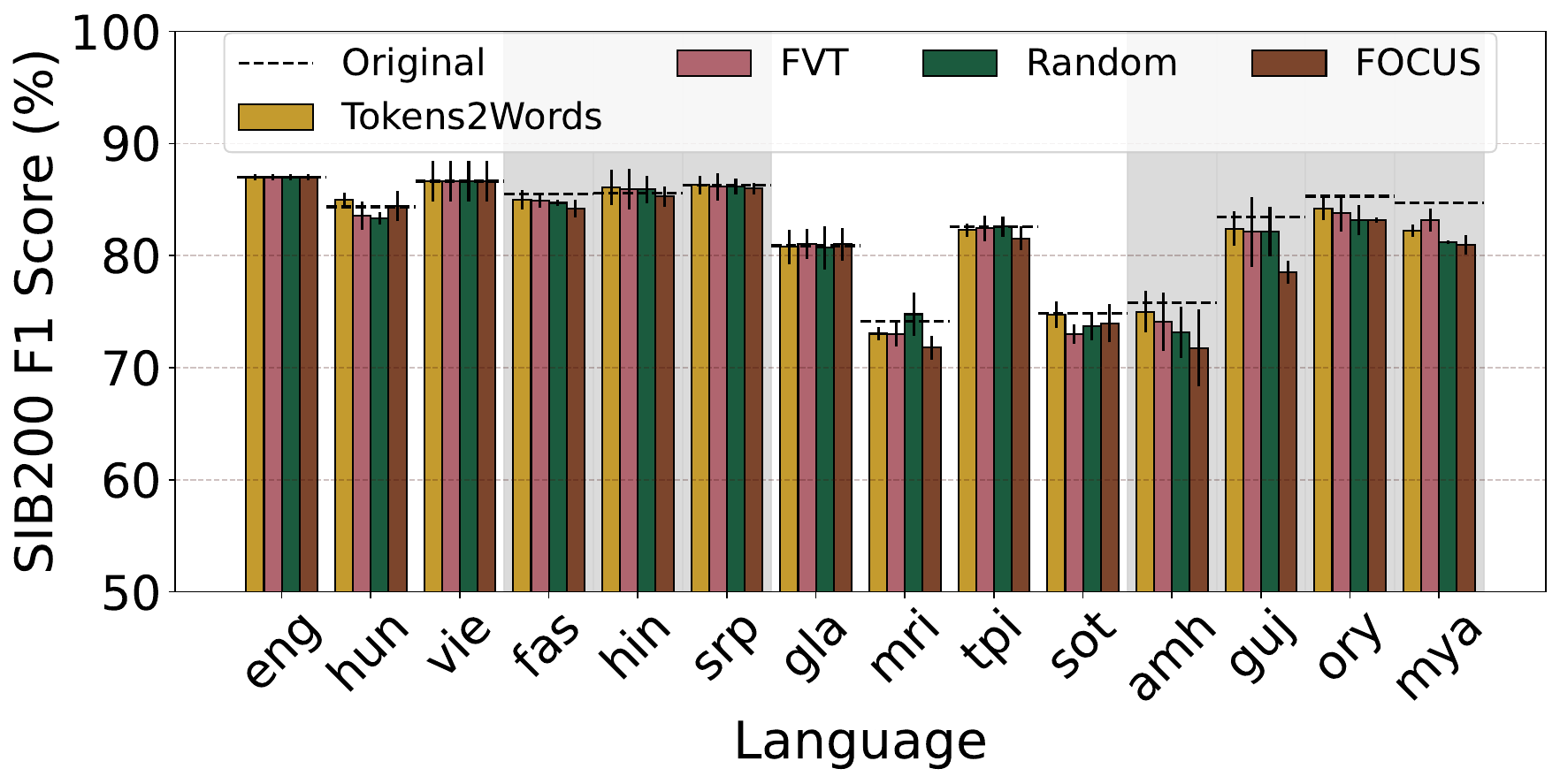}
        \caption{}
        \label{fig:sib200_10k_qwen35_4b}
    \end{subfigure}%
    ~
    \begin{subfigure}{0.49\textwidth}
        \centering
        \includegraphics[width=\textwidth]{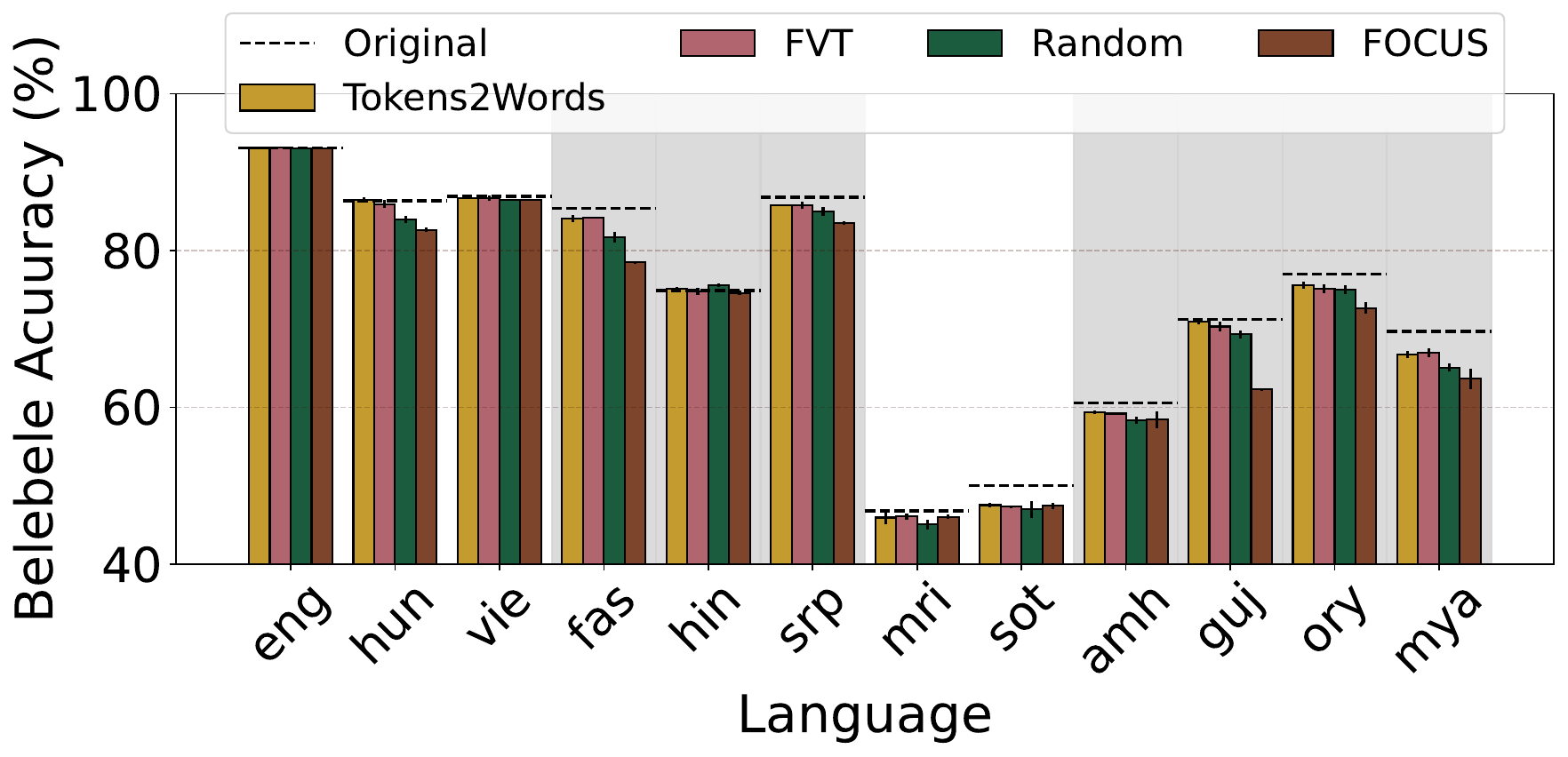}
        \caption{}
        \label{fig:belebele_10k_qwen35_4b}
    \end{subfigure}
    \caption{Performance across initialization methods on (\textit{a}) SIB200 and (\textit{b}) Belebele benchmarks. Model: Qwen3.5-4B results, $|C_{train}|=$10k.}
    \label{fig:qwen35_4b_10k_disc_init_comparison}
    
\end{figure*}

\begin{figure*}[h]
    \centering
    \begin{subfigure}{0.49\textwidth}
        \centering
        \includegraphics[width=\textwidth]{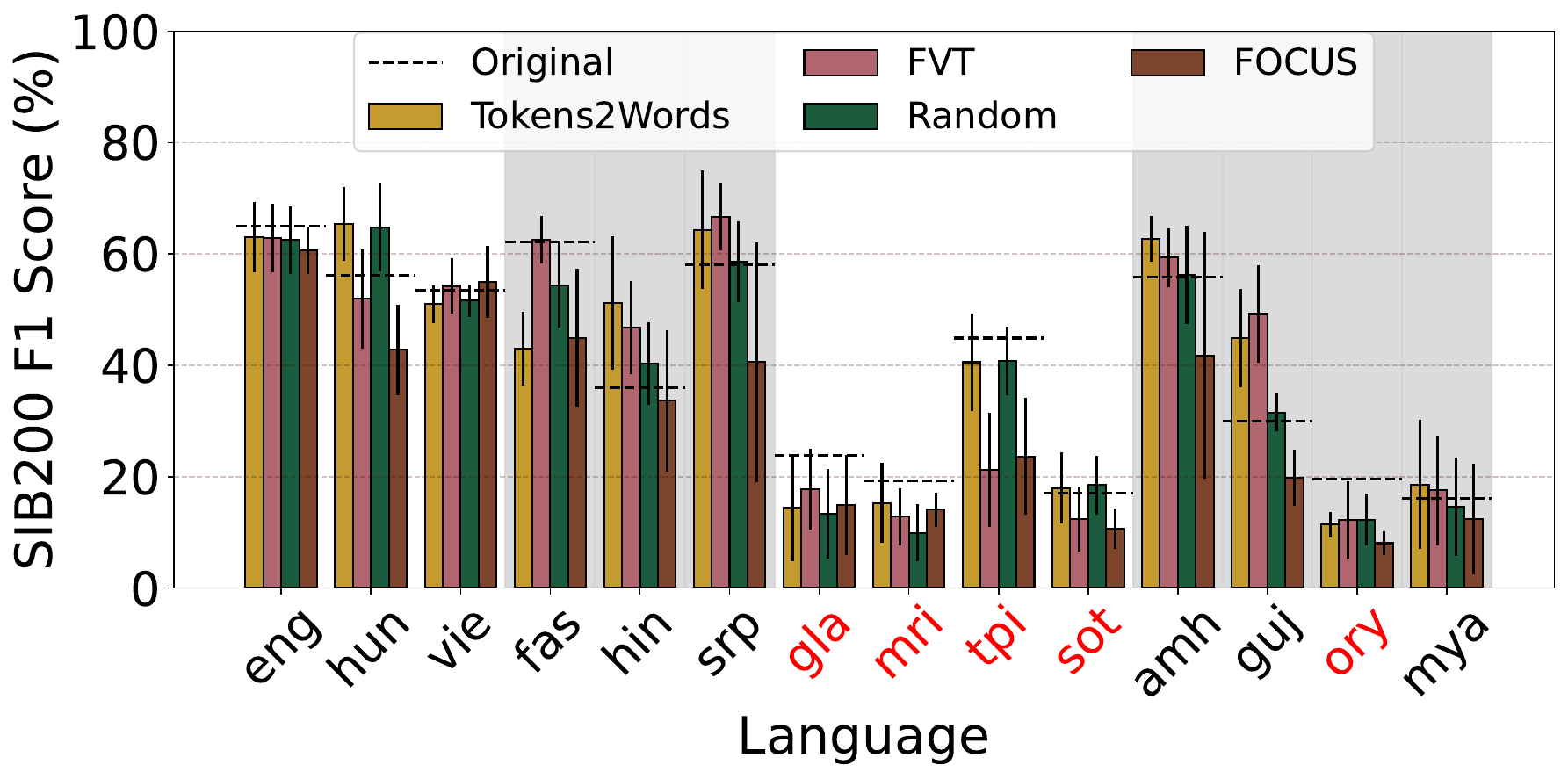}
        \caption{}
        \label{fig:sib200_10k_tinyaya}
    \end{subfigure}%
    ~
    \begin{subfigure}{0.49\textwidth}
        \centering
        \includegraphics[width=\textwidth]{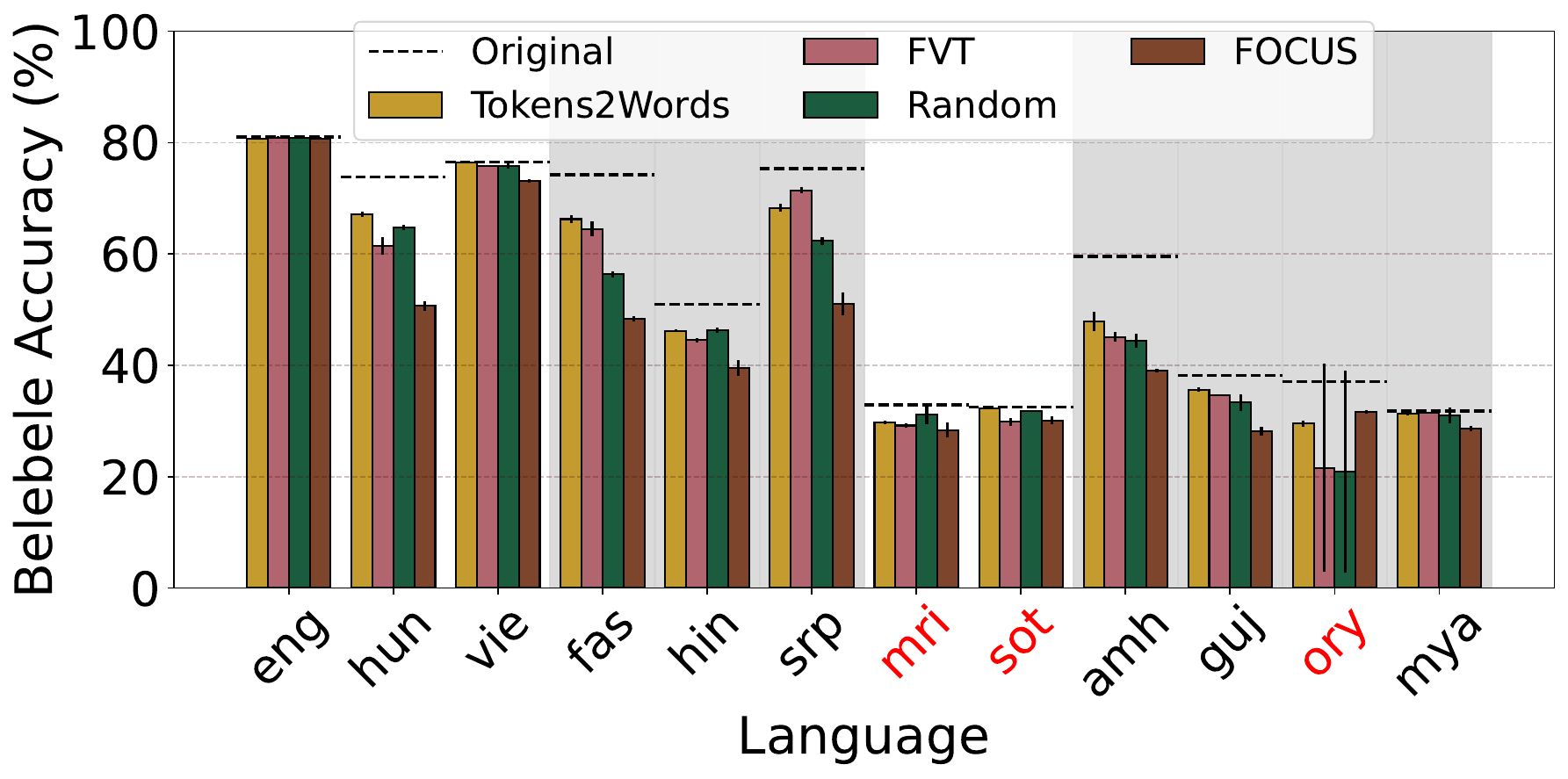}
        \caption{}
        \label{fig:belebele_10k_tinyaya}
    \end{subfigure}
    \caption{Performance across initialization methods on (\textit{a}) SIB200 and (\textit{b}) Belebele benchmarks. Model:TinyAya-Global results, $|C_{train}|=$1k.}
    \label{fig:tinyaya_10k_disc_init_comparison}
    
\end{figure*}

\cref{fig:qwen3_30b_1k_disc_init_comparison} shows SIB200 and Belebele performance on the 1k sequence training corpus on Qwen3-30B. The results follow our findings in \cref{sec:case_for_interp_init} for the 10k corpus where \patchscopes out-performs other baseline initializations for these tasks. Figures~\ref{fig:qwen35_4b_10k_disc_init_comparison} and \ref{fig:tinyaya_10k_disc_init_comparison} show the results on the two benchmarks on Qwen3.5-4B and TinyAya-Global. For both models, we observe that the results follow the trend observed in Qwen3-30B-A3B, however, to smaller degree. We speculate this result to be due to the smaller amount of token reduction, and due to poorer activations owing to smaller model size.

\subsection{\patchscopes: Performance vs Token Reduction Tradeoff}
\label{sec:more_perf_reduction_tradeoff}
\begin{figure*}[h]
    \centering
    \begin{subfigure}[b]{0.48\textwidth}
        %\centering
        \includegraphics[width=\textwidth]{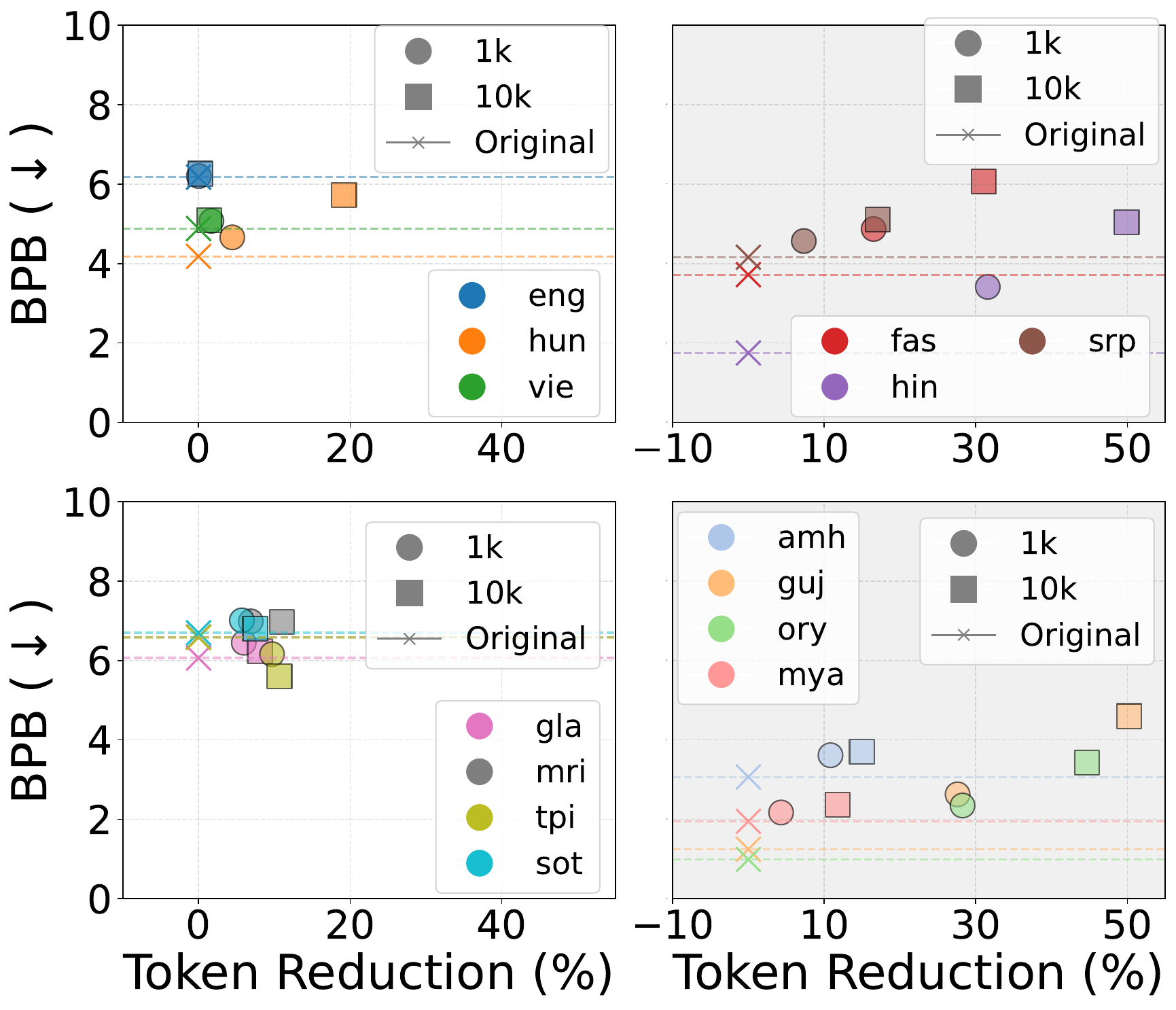}
        \caption{}
        \label{fig:tokreduction_vs_bpb_perf_psc}
    \end{subfigure}
    \hfill
    \begin{subfigure}[b]{0.48\textwidth}
        %\centering
        \includegraphics[width=\textwidth]{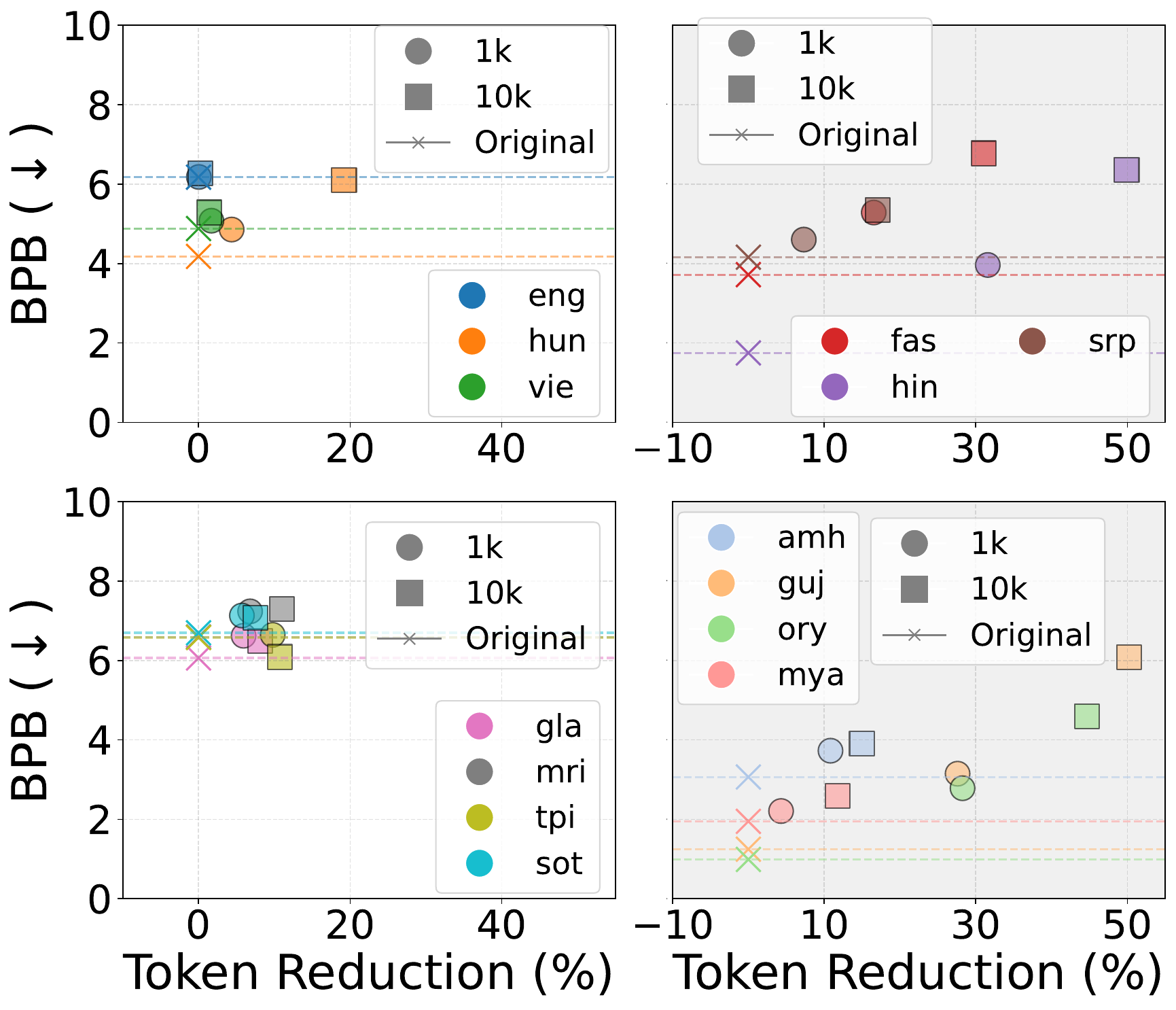}
        \caption{}
        \label{fig:tokreduction_vs_bpb_perf_fvt}
    \end{subfigure}
  
    \vspace{1cm}
    \begin{subfigure}[b]{0.48\textwidth}
        %\centering
        \includegraphics[width=\textwidth]{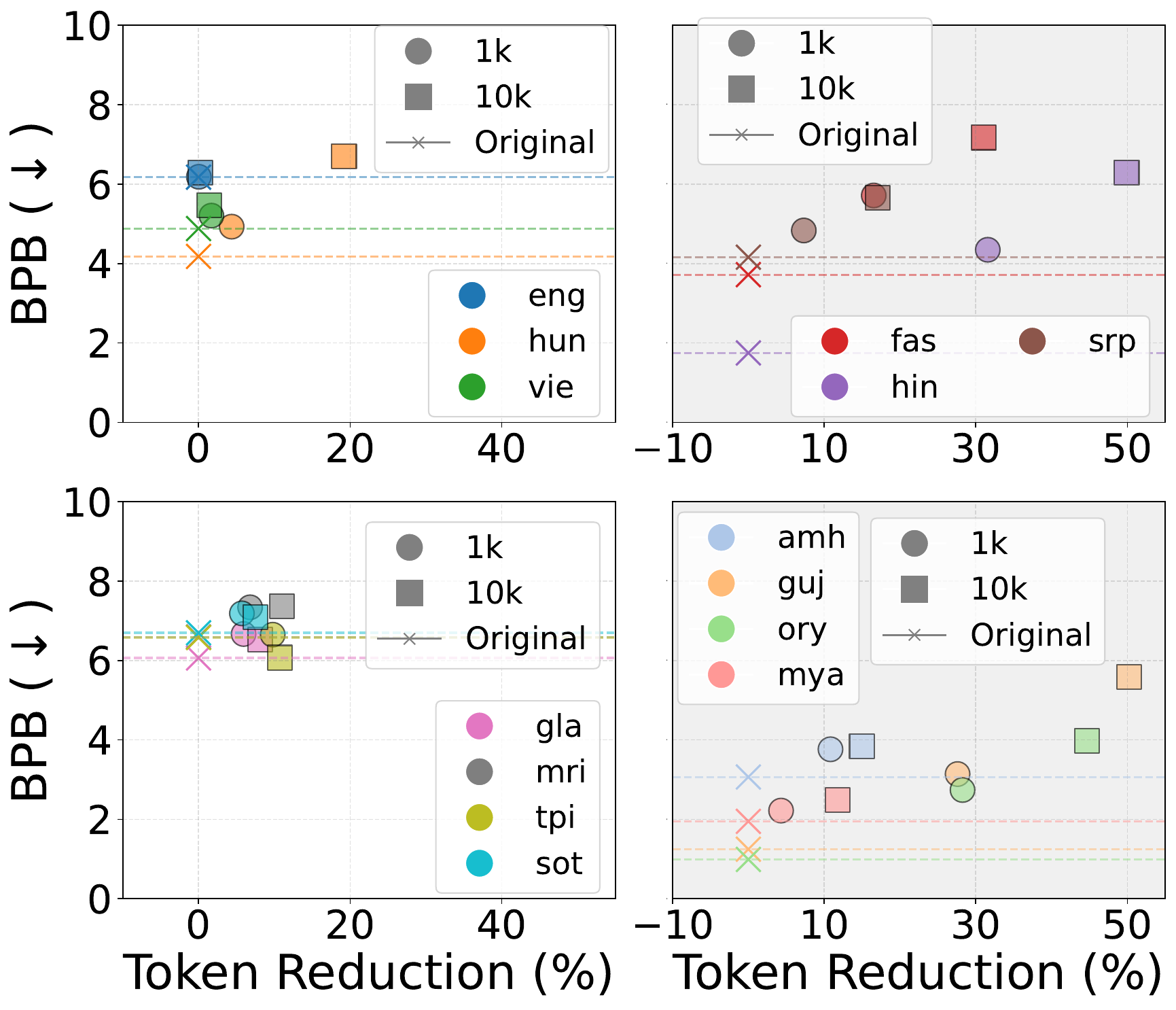}
        \caption{}
        \label{fig:tokreduction_vs_bpb_perf_randomn}
    \end{subfigure}
    \hfill
    \begin{subfigure}[b]{0.48\textwidth}
        %\centering
        \includegraphics[width=\textwidth]{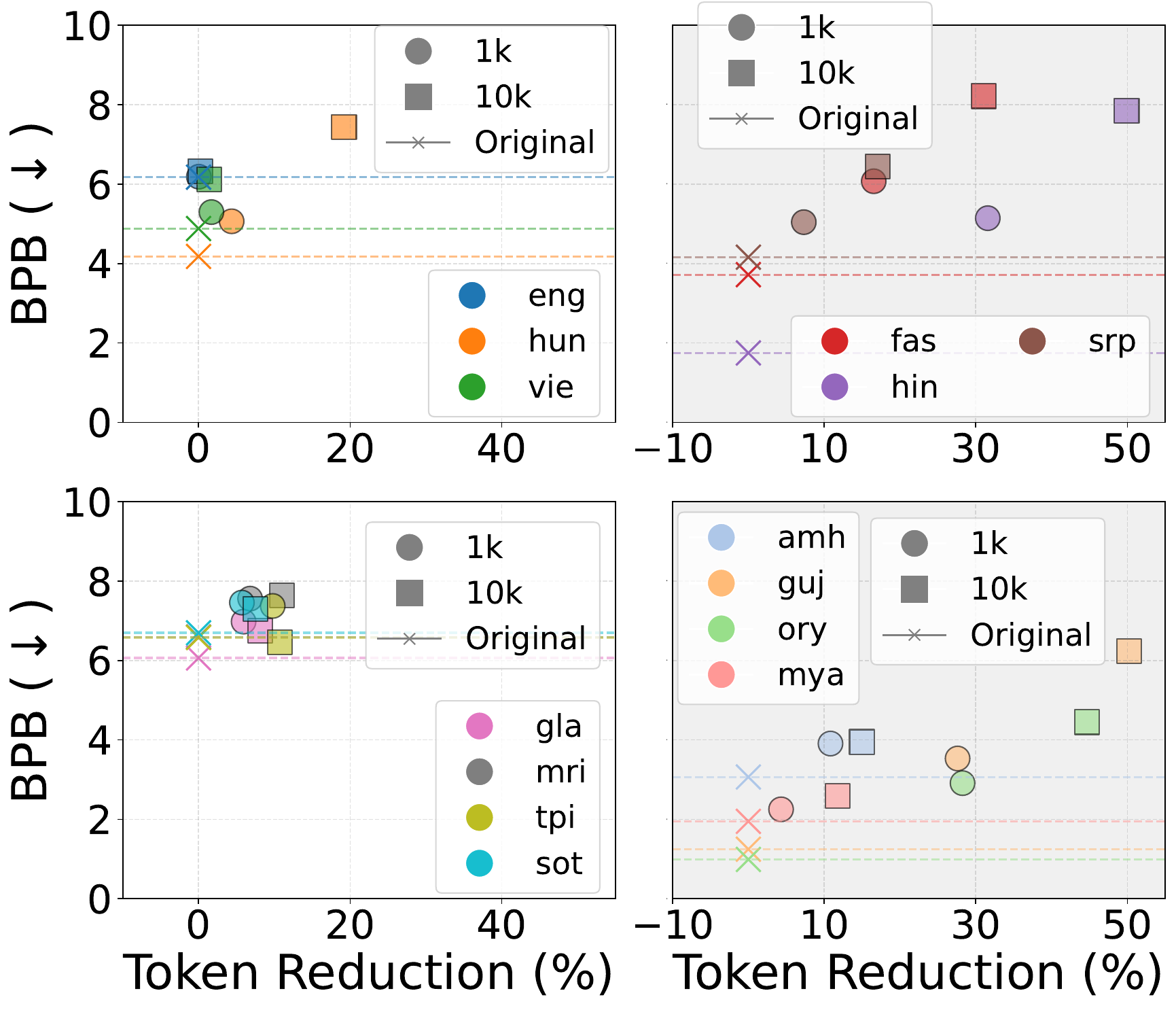}
        \caption{}
        \label{fig:tokreduction_vs_bpb_perf_focus}
    \end{subfigure}
    \caption{Token Reduction vs BPB Performance over different corpus sizes. Each subfigure contains results corresponding to different initialization: (a)\patchscopes, (b) FVT, (c) Random, and (d) FOCUS. Each subfigure contains four plots, each showing languages by resourcedness~(\texttt{high} vs. \texttt{low}) and script category~(\texttt{Latin} vs. \texttt{non-Latin}). Note that in the case of BPB, lower is better.}
    \label{fig:tokreduction_vs_bpb_perf}
    
\end{figure*}

\begin{figure*}[h]
    \centering
    \begin{subfigure}[b]{0.48\textwidth}
        %\centering
        \includegraphics[width=\textwidth]{figures/scatter_sib200_f1_vs_token_reduction_by_language_groups_qwen3_30b_Patchscopes.pdf}
        \caption{}
        \label{fig:tokreduction_vs_sib200_perf_psc}
    \end{subfigure}
    \hfill
    \begin{subfigure}[b]{0.48\textwidth}
        %\centering
        \includegraphics[width=\textwidth]{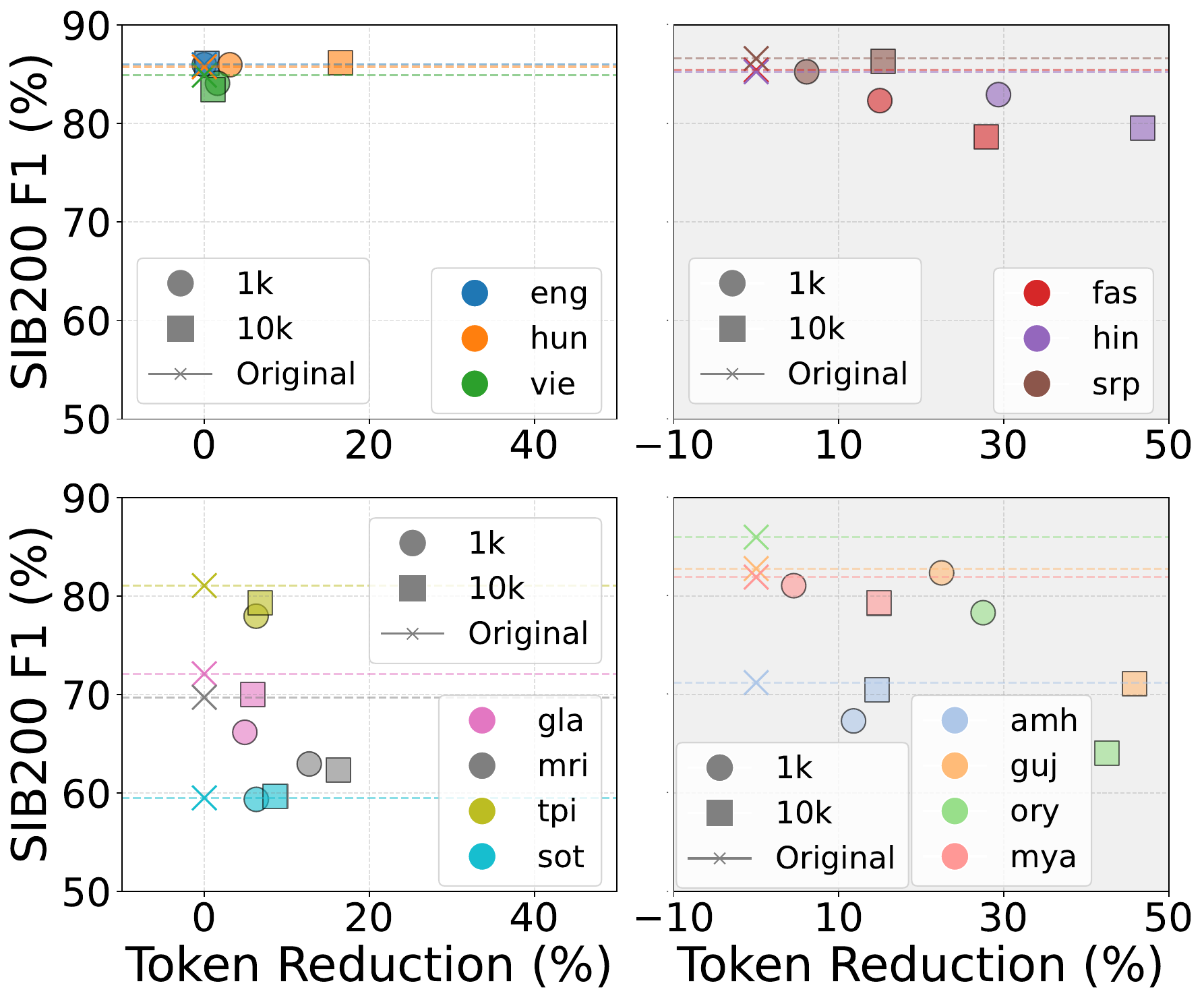}
        \caption{}
        \label{fig:tokreduction_vs_sib200_perf_fvt}
    \end{subfigure}
  
    \vspace{1cm}
    \begin{subfigure}[b]{0.48\textwidth}
        %\centering
        \includegraphics[width=\textwidth]{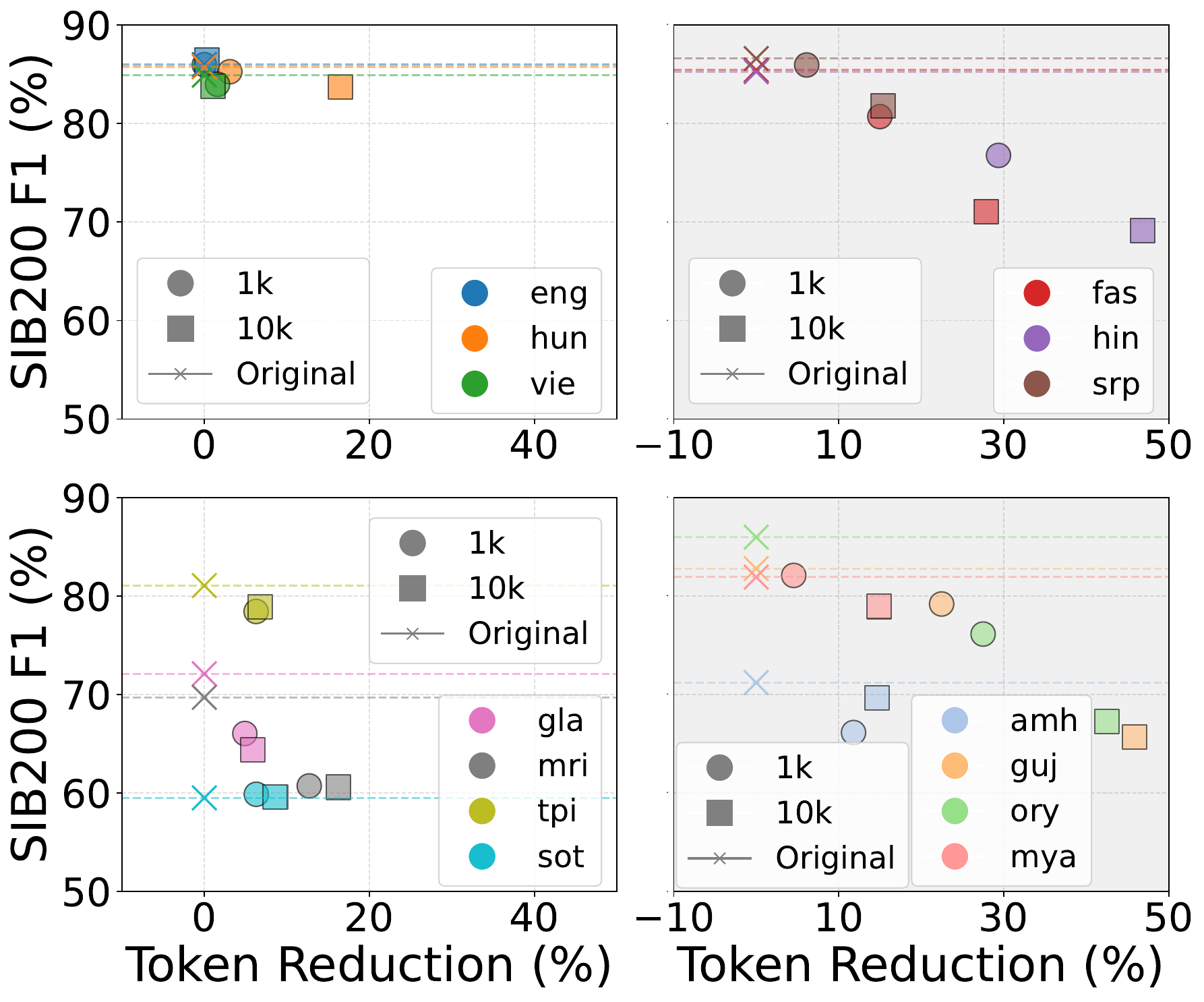}
        \caption{}
        \label{fig:tokreduction_vs_sib200_perf_randomn}
    \end{subfigure}
    \hfill
    \begin{subfigure}[b]{0.48\textwidth}
        %\centering
        \includegraphics[width=\textwidth]{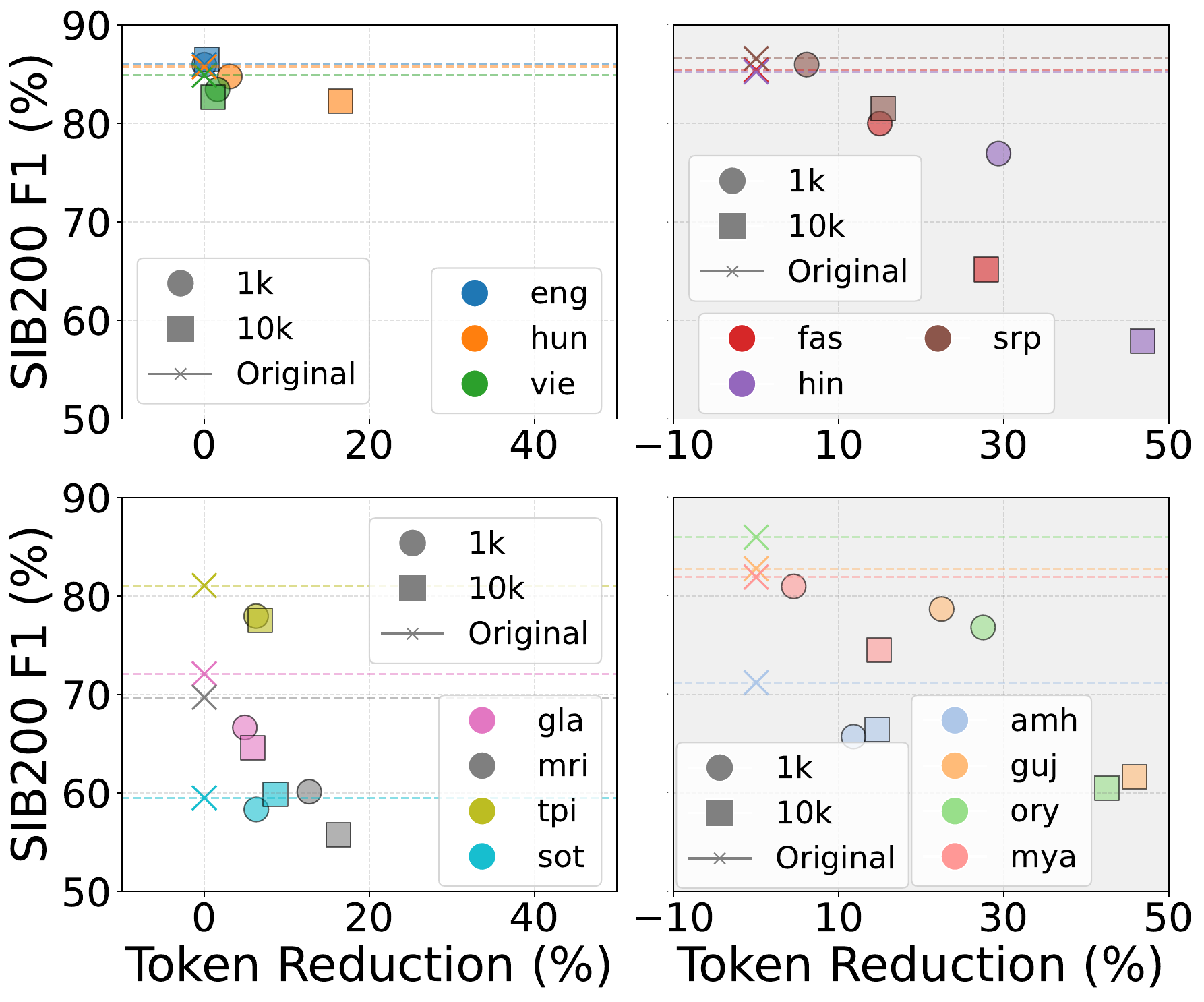}
        \caption{}
        \label{fig:tokreduction_vs_sib200_perf_focus}
    \end{subfigure}
    \caption{Token Reduction vs SIB200 Performance over different corpus sizes. Each subfigure contains results corresponding to different initialization: (a)\patchscopes, (b) FVT, (c) Random, and (d) FOCUS. Each subfigure contains four plots, each showing languages by resourcedness~(\texttt{high} vs. \texttt{low}) and script category~(\texttt{Latin} vs. \texttt{non-Latin}). An increase in the training dataset size~($C_{train}$) has different efficiency-performance effects for different language groups~(with \patchscopes initialization).}
    \label{fig:tokreduction_vs_sib200_perf}
    
\end{figure*}

\begin{figure*}[h]
    \centering
    \begin{subfigure}[b]{0.48\textwidth}
        %\centering
        \includegraphics[width=\textwidth]{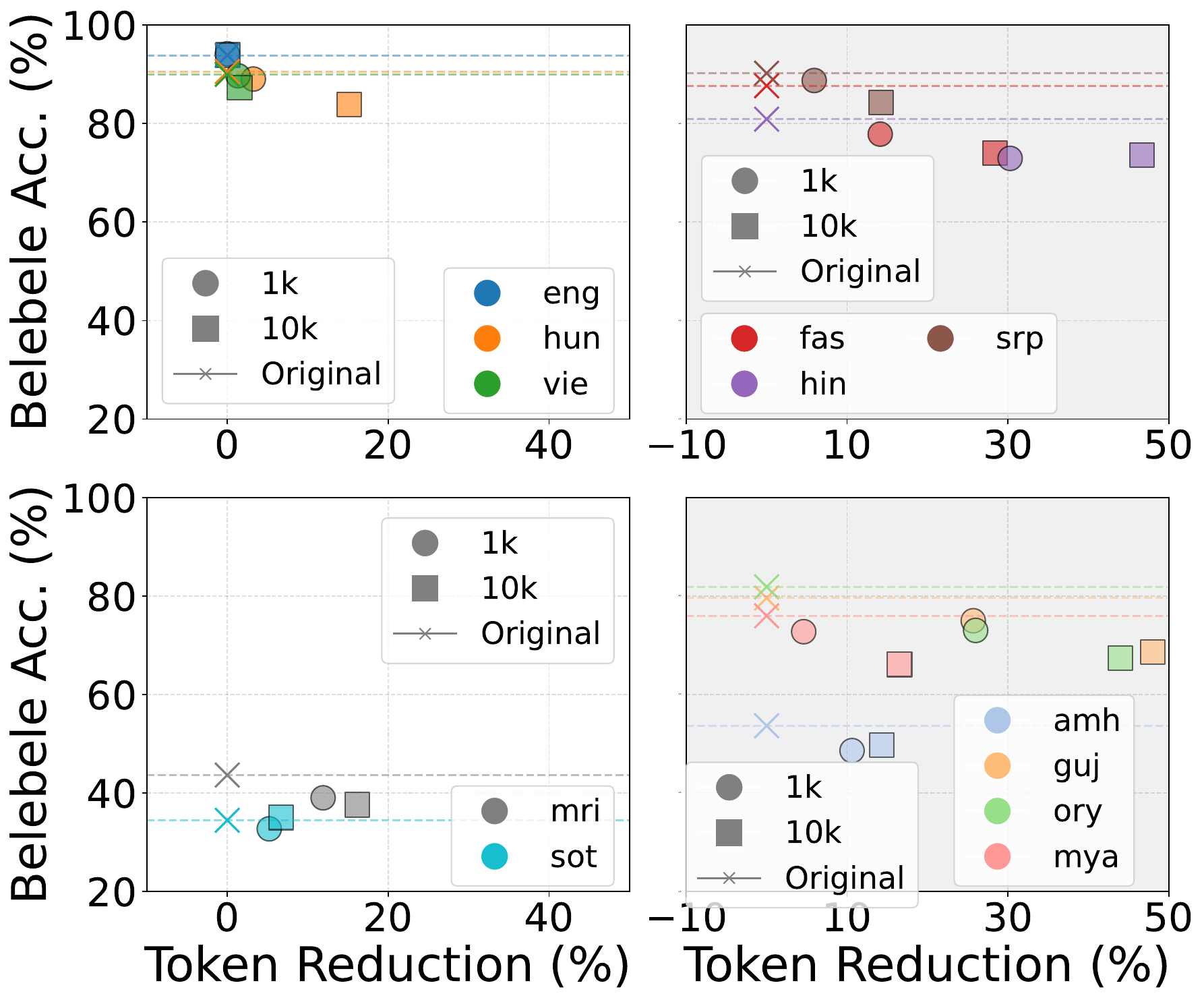}
        \caption{}
        \label{fig:tokreduction_vs_belebele_perf_psc}
    \end{subfigure}
    \hfill
    \begin{subfigure}[b]{0.48\textwidth}
        %\centering
        \includegraphics[width=\textwidth]{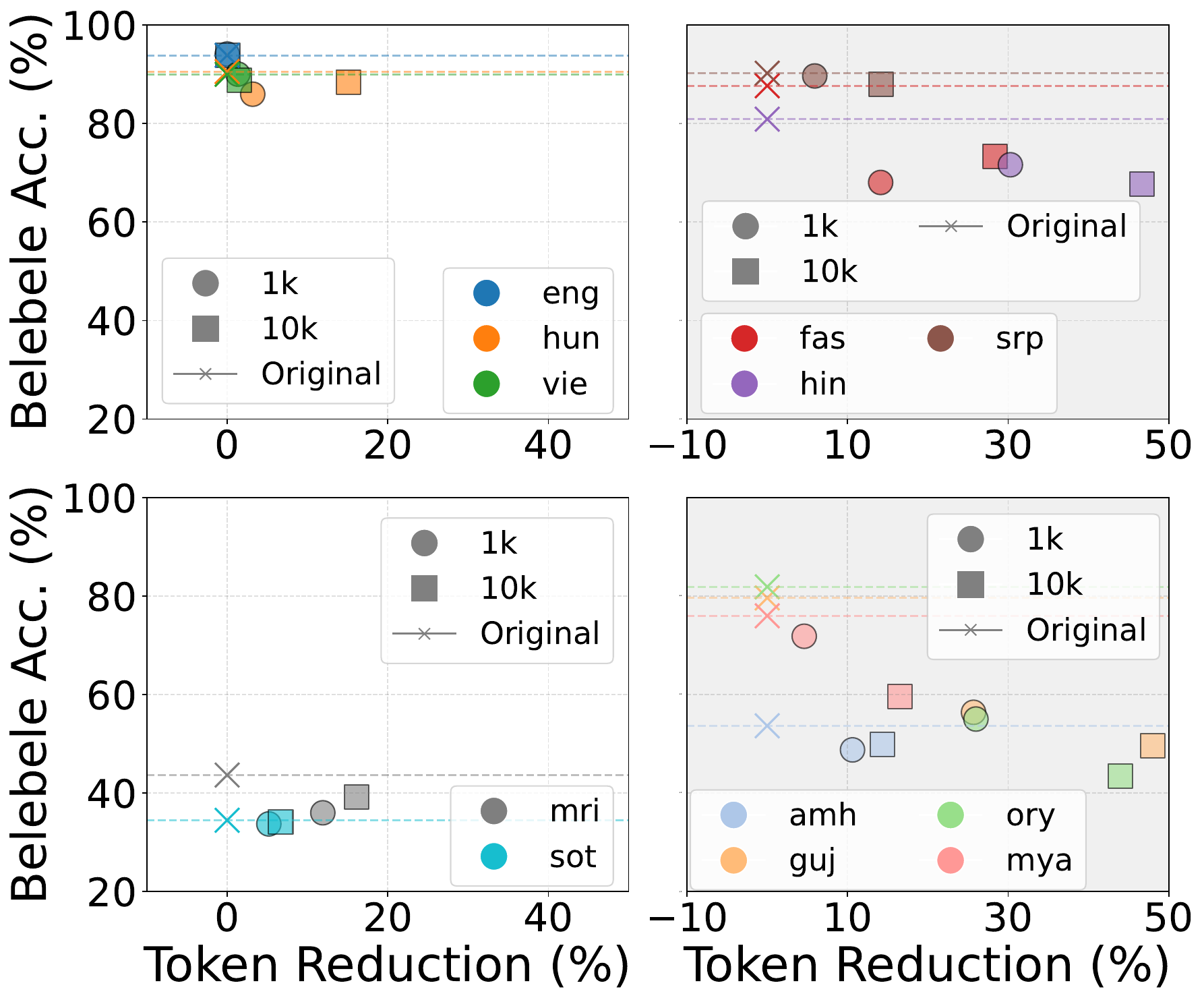}
        \caption{}
        \label{fig:tokreduction_vs_belebele_perf_fvt}
    \end{subfigure}
  
    \vspace{1cm}
    \begin{subfigure}[b]{0.48\textwidth}
        %\centering
        \includegraphics[width=\textwidth]{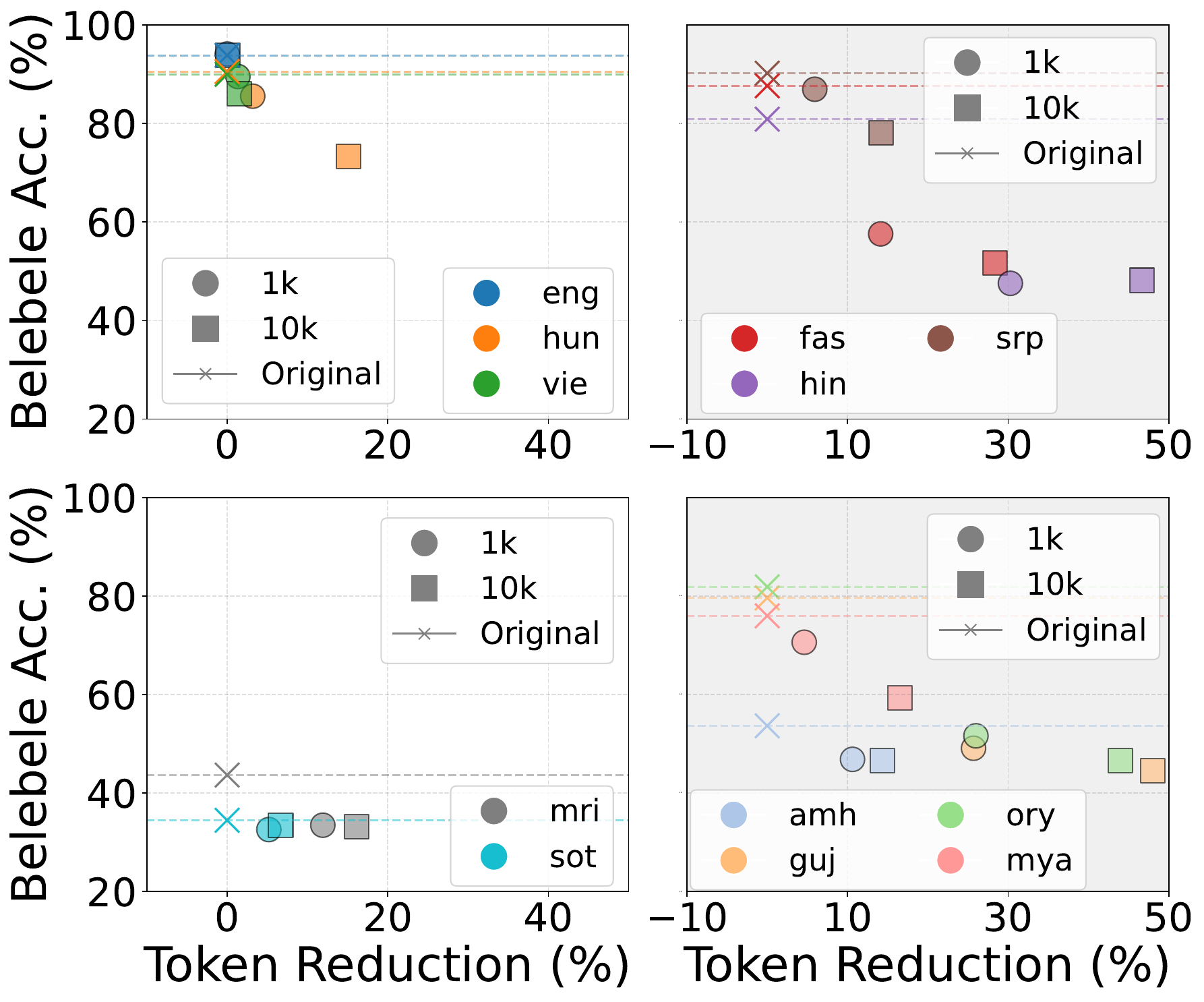}
        \caption{}
        \label{fig:tokreduction_vs_belebele_perf_randomn}
    \end{subfigure}
    \hfill
    \begin{subfigure}[b]{0.48\textwidth}
        %\centering
        \includegraphics[width=\textwidth]{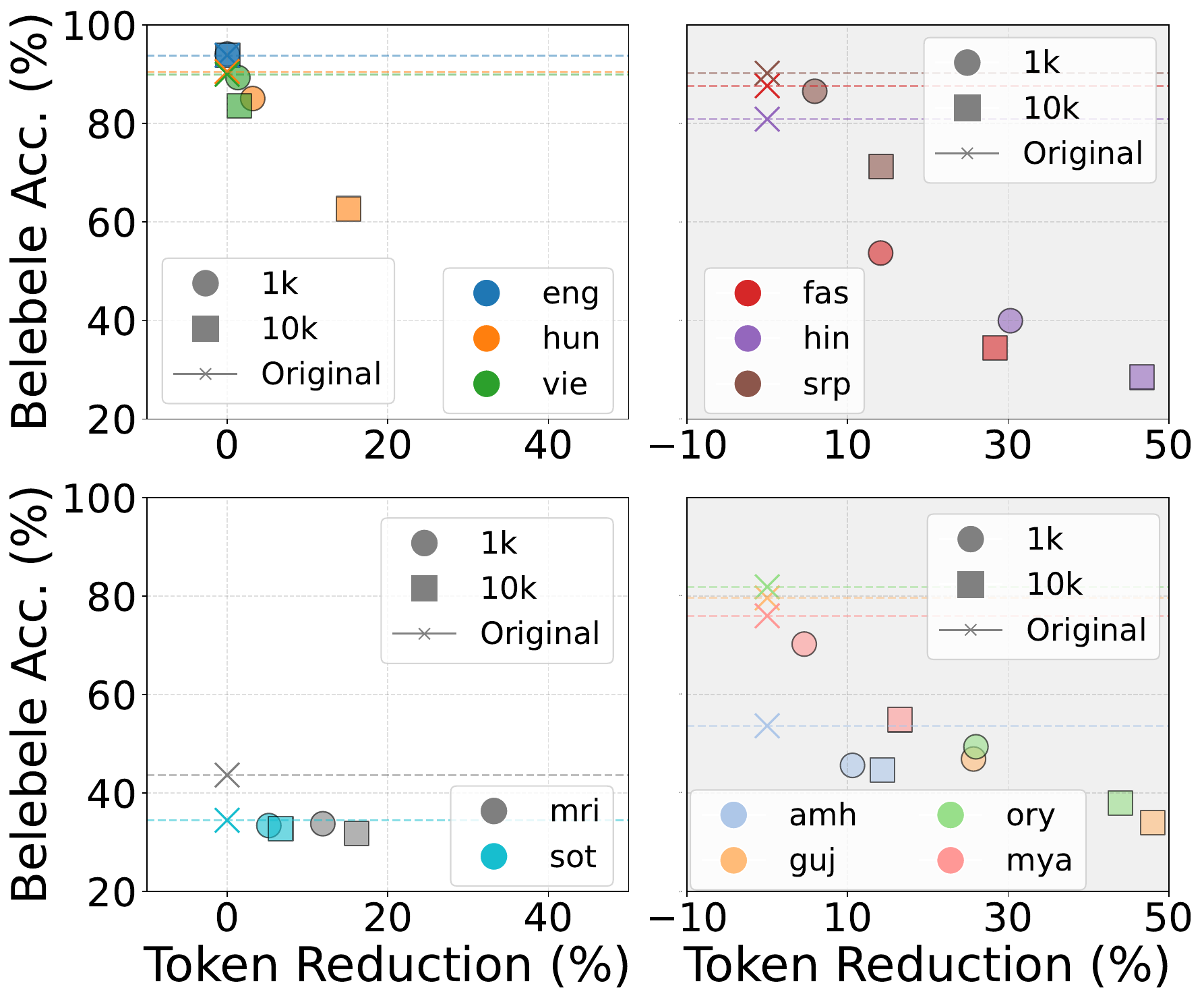}
        \caption{}
        \label{fig:tokreduction_vs_belebele_perf_focus}
    \end{subfigure}
    \caption{Token Reduction vs Belebele Performance over different corpus sizes. Each subfigure contains results corresponding to different initialization: (a)\patchscopes, (b) FVT, (c) Random, and (d) FOCUS. Each subfigure contains four plots, each showing languages by resourcedness~(\texttt{high} vs. \texttt{low}) and script category~(\texttt{Latin} vs. \texttt{non-Latin}).}
    \label{fig:tokreduction_vs_belebele_perf}
    
\end{figure*}

Figures \ref{fig:tokreduction_vs_bpb_perf}--\ref{fig:tokreduction_vs_belebele_perf} show the performance-efficiency tradeoff for different tasks with different initializations. We see that the trends largely follow the discussions in \cref{sec:case_for_interp_init}.

% \begin{figure*}
%     \centering
%     \begin{subfigure}
%         \includegraphics{figures/scatter_sib200_f1_vs_token_reduction_by_language_groups_qwen3_30b_Patchscopes.pdf}
%     \end{subfigure}
%     \caption{An increase in the training dataset size~($C_{train}$) has different efficiency-performance effects for different language groups~(with \patchscopes initialization)}
%     \label{fig:tok_reduct}
% \end{figure*}

% \begin{figure}
%     \centering
%     \includegraphics[width=0.5\linewidth]{figures/scatter_sib200_f1_vs_token_reduction_by_language_groups_qwen3_30b_Patchscopes.pdf}
%     \caption{An increase in the training dataset size~($C_{train}$) has different efficiency-performance effects for different language groups~(with \patchscopes initialization).}
%     \label{fig:corpus_variance_tokreduction_performance}
% \end{figure}

\subsection{\swpatchscopes: Initialization Comparison}
\label{sec:comparing_swpsc_vs_fvt_initialization}
\begin{figure*}[h]
    \centering
    \begin{subfigure}{0.49\textwidth}
        \centering
        \includegraphics[width=\textwidth]{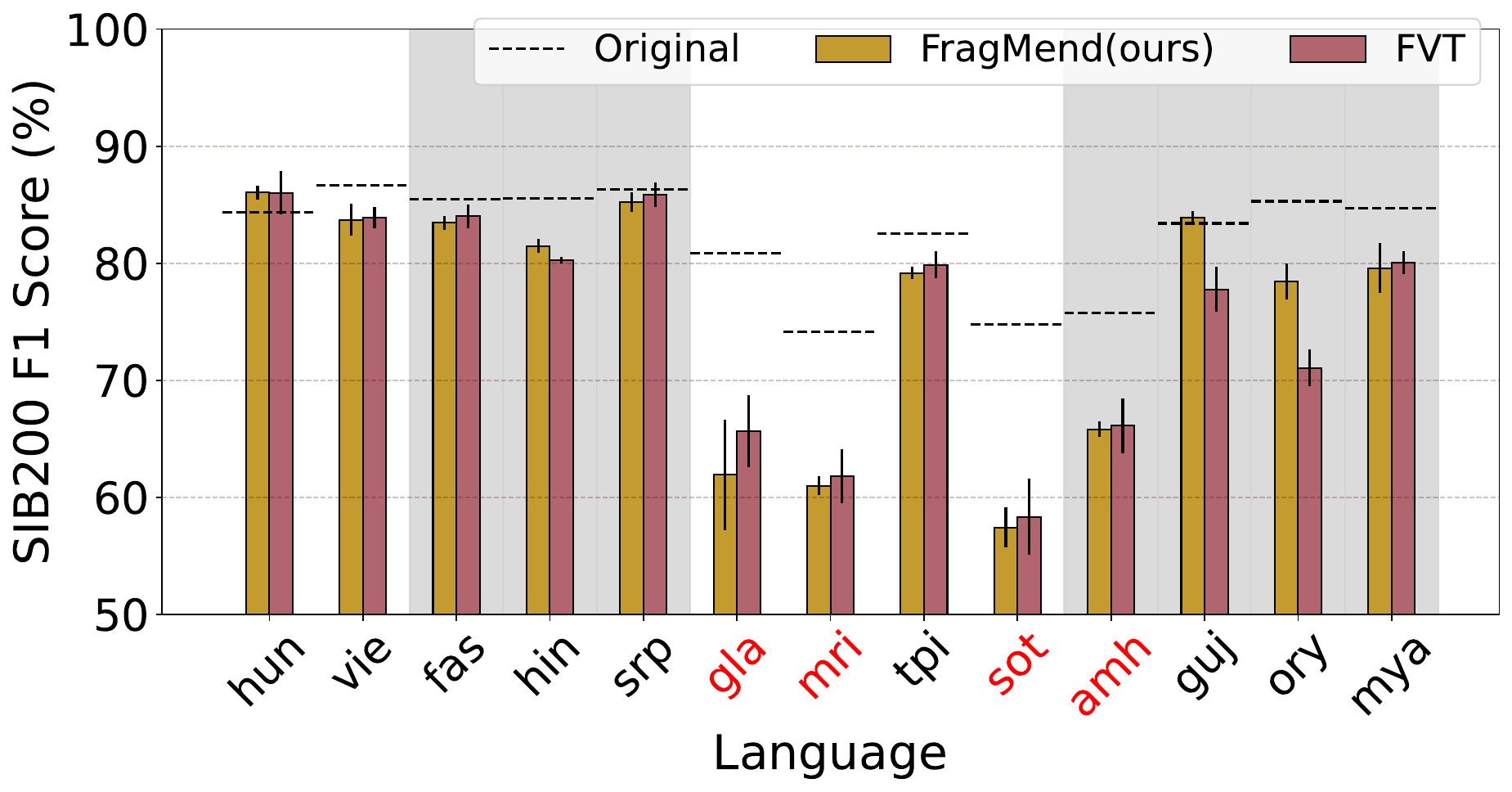}
        \caption{}
        \label{fig:swpsc_vs_fvt_init_qwen3_30b_sib200}
    \end{subfigure}%
    ~
    \begin{subfigure}{0.49\textwidth}
        \centering
        \includegraphics[width=\textwidth]{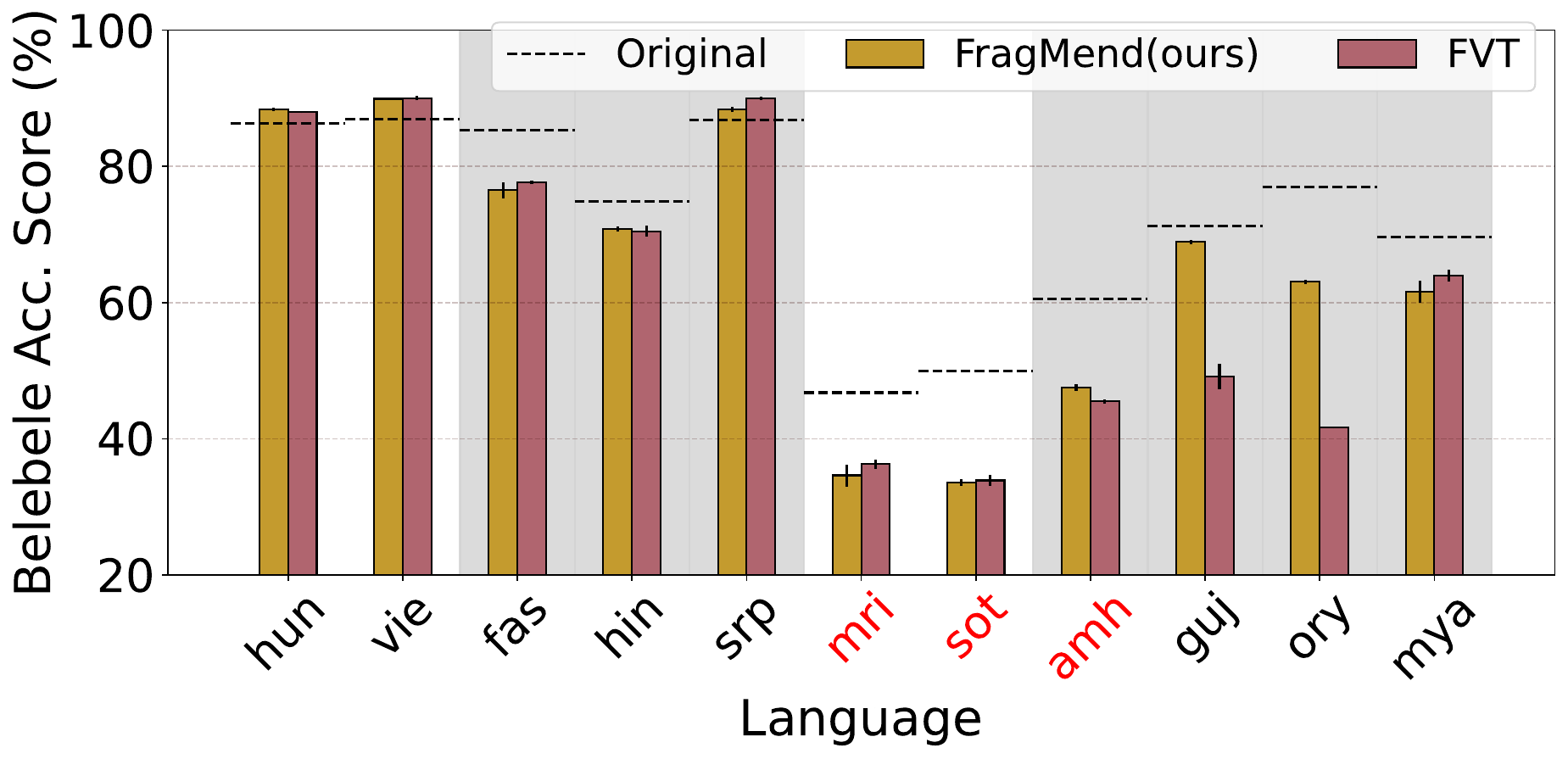}
        \caption{}
        \label{fig:swpsc_vs_fvt_init_qwen3_30b_belebele}
    \end{subfigure}
    \caption{Comparing (a) SIB200 and (b) Belebele performance of \swpatchscopes initialization and FVT for the same vocabulary items. $|C_{train}|=$1k sequences. Model: Qwen3-30B-A3B. Languages in red are not supported by the model.}
    \label{fig:swpsc_vs_fvt_init_comparison_qwen3_30b}
    
\end{figure*}

\begin{figure*}[h]
    \centering
    \begin{subfigure}{0.49\textwidth}
        \centering
        \includegraphics[width=\textwidth]{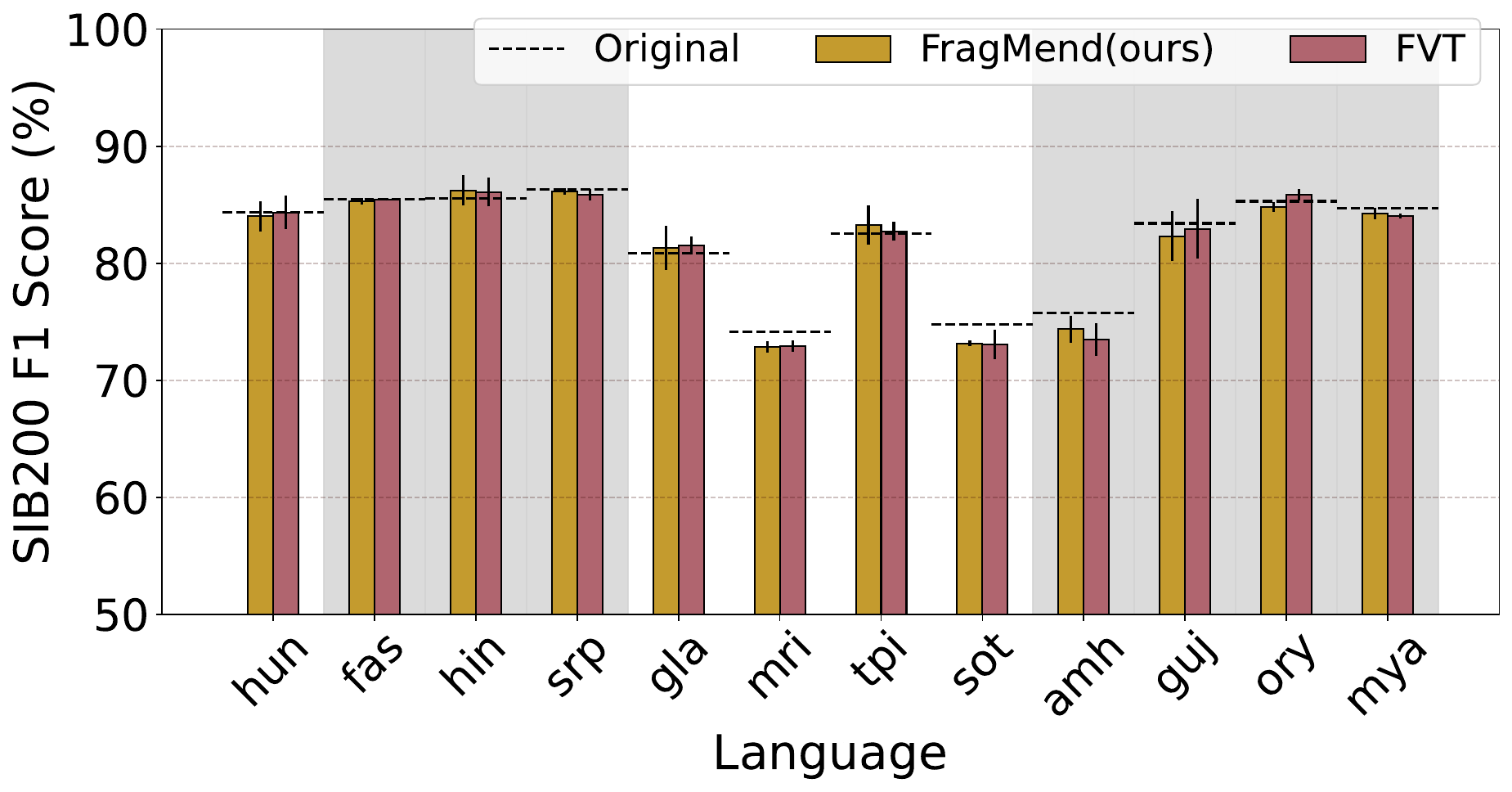}
        \caption{}
        \label{fig:swpsc_vs_fvt_init_qwen35_4b_sib200}
    \end{subfigure}%
    ~
    \begin{subfigure}{0.49\textwidth}
        \centering
        \includegraphics[width=\textwidth]{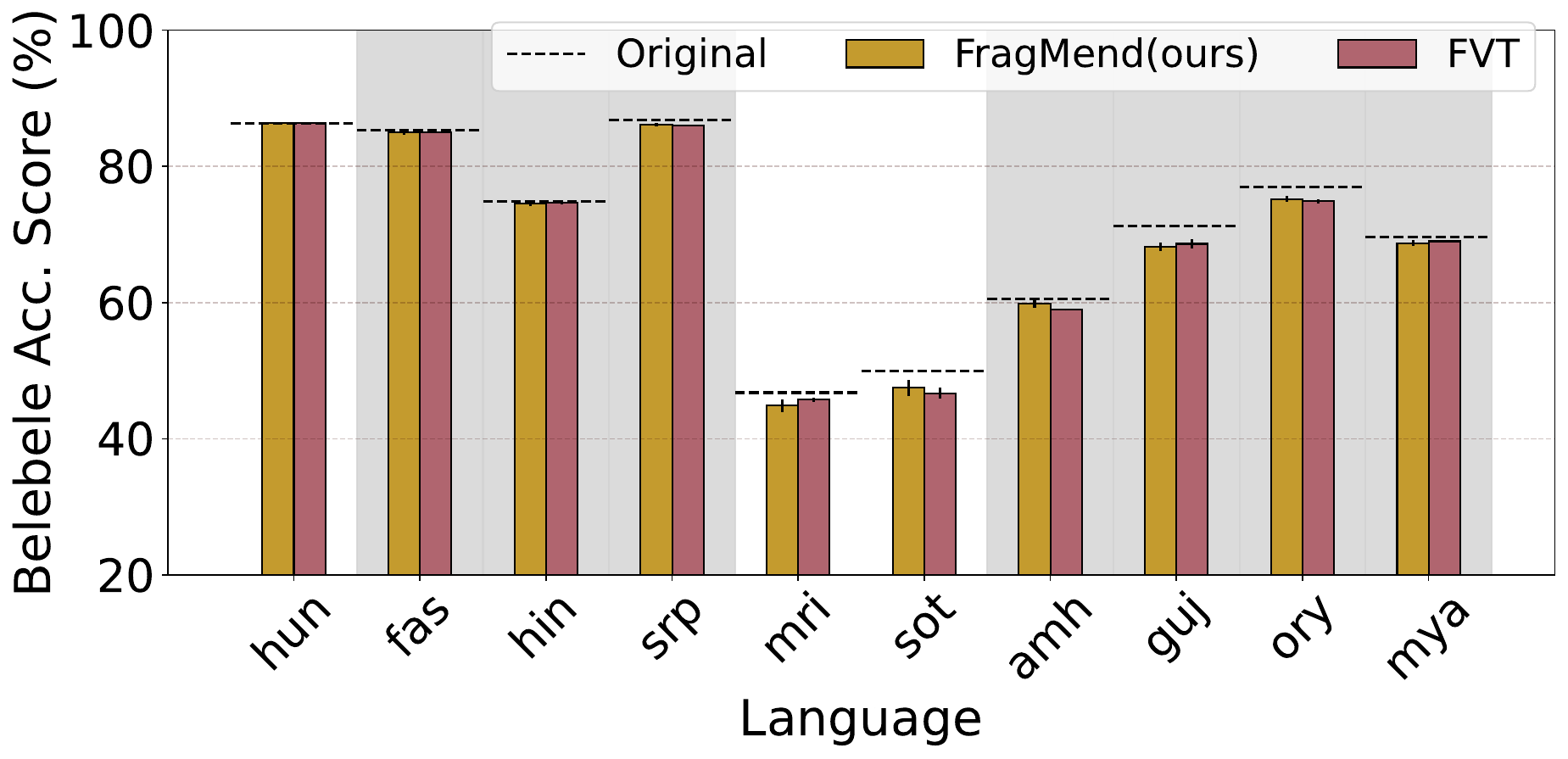}
        \caption{}
        \label{fig:swpsc_vs_fvt_init_qwen35_4b_belebele}
    \end{subfigure}
    \caption{Comparing (a) SIB200 and (b) Belebele performance of \swpatchscopes initialization and FVT for the same vocabulary items. $|C_{train}|=$1k sequences. Model: Qwen3.5-4B.}
    \label{fig:swpsc_vs_fvt_init_comparison_qwen35_4b}
    
\end{figure*}

\begin{figure*}[h]
    \centering
    \begin{subfigure}{0.49\textwidth}
        \centering
        \includegraphics[width=\textwidth]{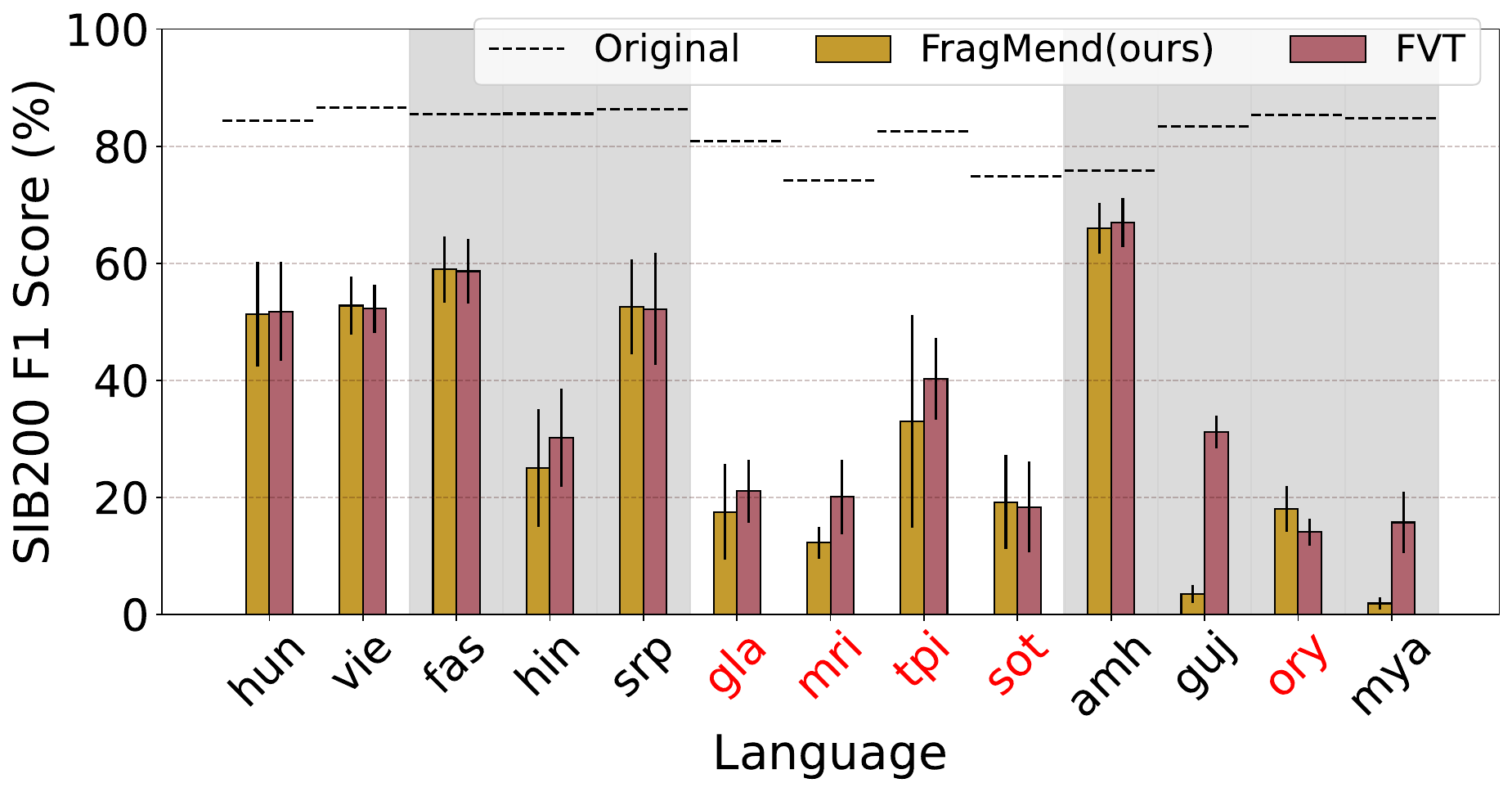}
        \caption{}
        \label{fig:swpsc_vs_fvt_init_tinyaya_sib200}
    \end{subfigure}%
    ~
    \begin{subfigure}{0.49\textwidth}
        \centering
        \includegraphics[width=\textwidth]{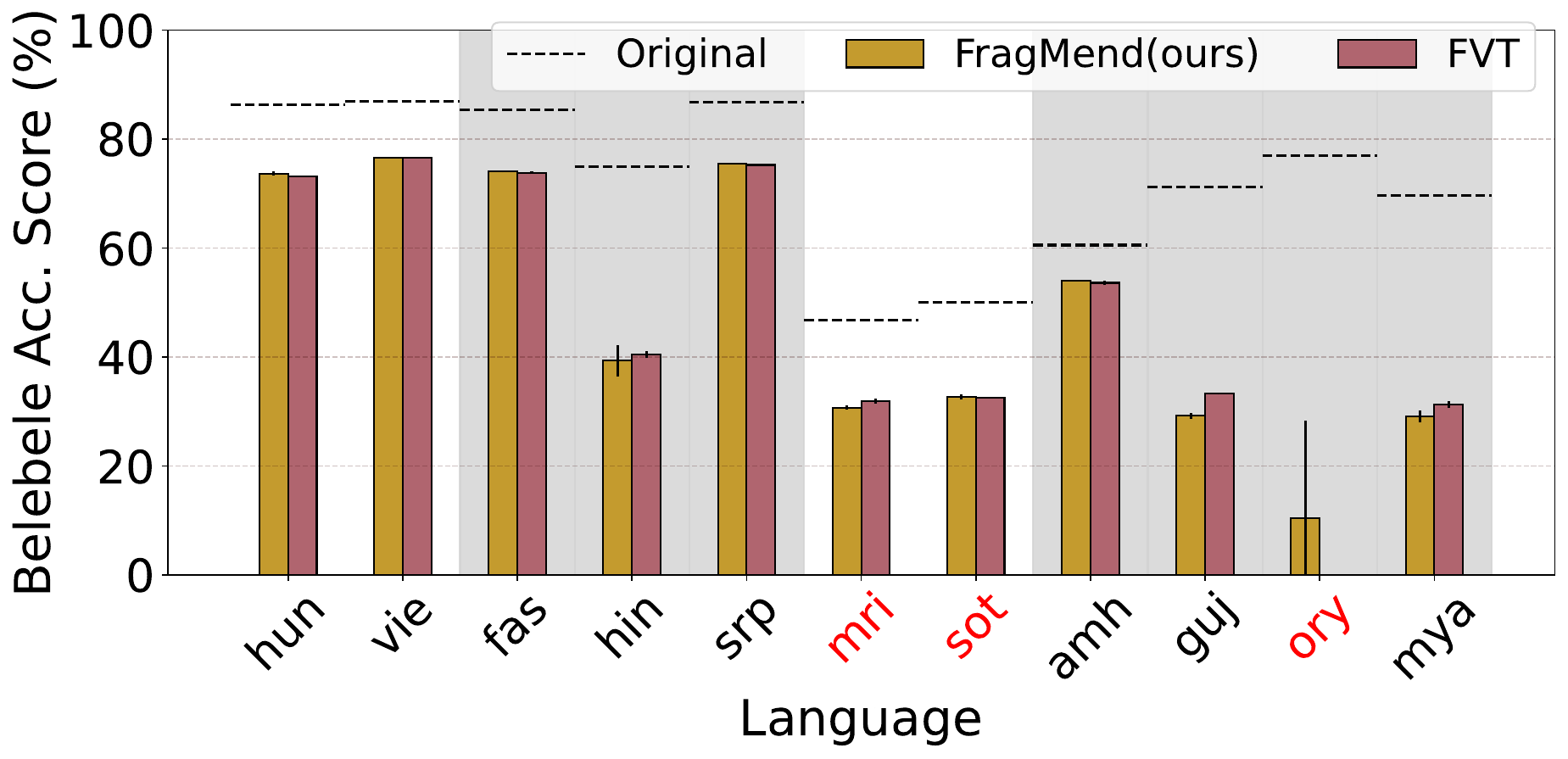}
        \caption{}
        \label{fig:swpsc_vs_fvt_init_tinyaya_belebele}
    \end{subfigure}
    \caption{Comparing (a) SIB200 and (b) Belebele performance of \swpatchscopes initialization and FVT for the same vocabulary items. $|C_{train}|=$1k sequences. Model: TinyAyaGlobal-3B. Languages in red are not supported by the model.}
    \label{fig:swpsc_vs_fvt_init_comparison_tinyayaglobal}
    
\end{figure*}

Figures~\ref{fig:swpsc_vs_fvt_init_comparison_qwen3_30b}--\ref{fig:swpsc_vs_fvt_init_comparison_tinyayaglobal} show the comparison between the \swpatchscopes method and FVT for the same expanded vocabulary as determined by \swpatchscopes. For Qwen-3-30B-A3B~(\ref{fig:swpsc_vs_fvt_init_comparison_qwen3_30b}), the comparisons seem to follow our findings with the \patchscopes initialization where \swpatchscopes outperforms the FVT baseline. However, with Qwen3.5-4B and TinyAya-Global the gains subside, possibly due to poor activations like in \patchscopes.  

\subsection{Evaluation of Generative Tasks}
\label{sec:eval_generative_task}

Our primary evaluation setup considers classification tasks. During our initial experiments, we considered generative tasks~(such as translation: FLORES-200 En-X~\citep{NLLBTeam2022}, discussed here) in addition to our evaluation setup. We performed reference-based evaluation using statistical metrics and LLM-based judgments. However, we found several issues with this setup for low-resource languages:
\begin{enumerate}
    \item We find that the statistical metrics grossly underestimate the similarity in the responses between two expanded models. For the FLORES task, which uses single-reference translation for evaluation, we find that statistical metrics like ChrF++ differ in several orders of magnitude when rating equivalent translations, owing to synonym usage and transliteration~(in languages using a non-Latin script).  With that caveat, we report the results in Table~\ref{tab:gen_tasks_stat_metrics}.
    \item Alternatively, one can propose using LLMs as judges. However, pilot studies found significantly low Cohen’s kappa agreement of 0.35 between a proprietary LLM judge~(GPT-5.5) and human experts for Gujarati, computed using linear-weighted Cohen’s kappa on LLM and human judgments over 150 translation samples (50 translations each from the original, \patchscopes, and FVT initialized models, for the same 50 test instances). Both the LLM judge and the humans used the same 5-point Likert scale for assessing quality. Low agreement among LLM judges in low-resource languages has also been corroborated by prior literature~\citep{fu-liu-2025-reliable}. Moreover, the LLM-as-a-judge approach or human annotations is monetarily infeasible given our experimental scale ($\sim$1500 models). In any case, we conducted an LLM-as-a-judge study on the same dataset used to report the statistical metrics. The results are reported in Table~\ref{tab:gen_stars_llm_judge}.
\end{enumerate}

For these reasons, we strongly believe that classification tasks provide an objective, scalable testbed for evaluating our models.

\begin{table}[t]
\centering

\begin{tabular}{lccc}
\toprule
\textbf{Language} & \textbf{Original} & \textbf{\patchscopes} & \textbf{FVT} \\
\midrule
Persian    & \cellcolor{gray!30} 45.8 & \cellcolor{gray!30}45.2 & \cellcolor{gray!30}44.0 \\
Hindi      & \cellcolor{gray!30}49.7 & \cellcolor{gray!30}45.3 & \cellcolor{gray!30}45.4 \\
Serbian    & \cellcolor{gray!30}48.3 & \cellcolor{gray!30}47.8 & \cellcolor{gray!30}47.0 \\
Gujarati   & \cellcolor{gray!30}41.0 & \cellcolor{gray!30}39.9 & \cellcolor{gray!30}40.0 \\
Odia       & \cellcolor{gray!30}38.0 & \cellcolor{gray!30}35.6 & \cellcolor{gray!30}36.9 \\
Burmese    & \cellcolor{gray!30}31.4 & \cellcolor{gray!30}28.5 & \cellcolor{gray!30}28.9 \\
\bottomrule
\end{tabular}
\caption{Translation performance on FLORES-200 En-X task across different models as measured using ChrF++ score~(for the expanded models, $|C_{train}|=10k$; Original model: Qwen3-30B-A3B). }
\label{tab:gen_tasks_stat_metrics}
\end{table}

\begin{table}[t]
\centering
\begin{tabular}{lccc}
\toprule
\textbf{Language} & \textbf{Original} & \textbf{\patchscopes} & \textbf{FVT} \\
\midrule
Persian    & \cellcolor{gray!30}4.00 & \cellcolor{gray!30}3.99 & \cellcolor{gray!30}3.86 \\
Hindi      & \cellcolor{gray!30}4.16 & \cellcolor{gray!30}3.97 & \cellcolor{gray!30}3.95 \\
Serbian    & \cellcolor{gray!30}3.71 & \cellcolor{gray!30}3.67 & \cellcolor{gray!30}3.65 \\
Gujarati   & \cellcolor{gray!30}3.91 & \cellcolor{gray!30}3.77 & \cellcolor{gray!30}3.74 \\
Odia       & \cellcolor{gray!30}3.63 & \cellcolor{gray!30}3.26 & \cellcolor{gray!30}3.62 \\
Burmese    & \cellcolor{gray!30}3.15 & \cellcolor{gray!30}2.72 & \cellcolor{gray!30}2.94 \\
\bottomrule
\end{tabular}
\caption{Translation performance on FLORES-200 En-X task across different models as measured using an LLM Judge~(GPT 5.5). The translations were evaluated in a reference-less manner on a 5-point Likert scale using the GEMBA-stars~\citep{kocmi-federmann-2023-large} style template~(for the expanded models, $|C_{train}|=10k$; Original model: Qwen3-30B-A3B).} 
\label{tab:gen_stars_llm_judge}
\end{table}

\subsection{Inference Speed-Ups Post Vocabulary Expansion}
\label{sec:inference_speedup}
In this section, we present the impact of vocabulary expansion on generation speed during inference. We compare the generation speed of the model expanded using \swpatchscopes~($|C_{train}|=1k$) with that of the original model on 200 instances from the FLORES-200 En-X translation task. These results are reported in Table~\ref{tab:inference_gains}. Latency, token, and character reduction are computed using the formula given in Equation \ref{eq:reduction}. A negative metric-reduction value indicates that the expanded model has a higher absolute value for that metric than the original model. Note that inference speed may fluctuate across models, as the generation length need not be the same. To normalize for length, we provide comparisons in terms of output characters, in addition to tokens. All metrics are averaged across three runs, and with a batch size of 1 to ensure accurate comparison. 

\begin{table}[t]
\centering
\begin{tabular}{lccccc}
\toprule
\textbf{Language} & \shortstack{\textbf{Original} \\ \textbf{Speed}} & \shortstack{\textbf{Expanded} \\ \textbf{Speed}} & \shortstack{\textbf{Inference Latency}\\\textbf{Reduction}} & \shortstack{\textbf{Token}\\\textbf{Reduction}} & \shortstack{\textbf{Character}\\\textbf{Reduction}} \\
\midrule
Persian  & \cellcolor{gray!30}5.64  & \cellcolor{gray!30}5.46  & \cellcolor{gray!30}3.10  & \cellcolor{gray!30}2.08  & \cellcolor{gray!30}-0.51 \\
Hindi    & \cellcolor{gray!30}9.31  & \cellcolor{gray!30}7.89  & \cellcolor{gray!30}15.25 & \cellcolor{gray!30}13.42 & \cellcolor{gray!30}0.44 \\
Serbian  & \cellcolor{gray!30}5.11  & \cellcolor{gray!30}4.87  & \cellcolor{gray!30}4.69  & \cellcolor{gray!30}2.37  & \cellcolor{gray!30}-2.04 \\
Amharic  & \cellcolor{gray!30}15.73 & \cellcolor{gray!30}15.05 & \cellcolor{gray!30}4.33  & \cellcolor{gray!30}2.65  & \cellcolor{gray!30}-0.39 \\
Gujarati & \cellcolor{gray!30}14.11 & \cellcolor{gray!30}12.34 & \cellcolor{gray!30}12.51 & \cellcolor{gray!30}12.34 & \cellcolor{gray!30}-1.54 \\
Odia     & \cellcolor{gray!30}19.88 & \cellcolor{gray!30}17.24 & \cellcolor{gray!30}13.27 & \cellcolor{gray!30}11.48 & \cellcolor{gray!30}-1.99 \\
Burmese  & \cellcolor{gray!30}18.75 & \cellcolor{gray!30}17.29 & \cellcolor{gray!30}7.76  & \cellcolor{gray!30}5.65  & \cellcolor{gray!30}-15.82 \\
\bottomrule
\end{tabular}
\caption{Inference speed-ups post-expansion. We observe clear latency gains post-expansion. All speeds are measured in seconds per instance. All reductions are in percentage values~(Original model: Qwen3-30B-A3B).}
\label{tab:inference_gains}
\end{table}

We observe a clear correlation (Pearson’s $r$=0.98, Spearman’s $\rho$=0.93) between the latency reduction at generation and the reduction in tokens. Furthermore, expanded models are much faster, even when the generations are slightly longer than those of the original model (e.g., Burmese).

\subsection{Underutilization of Added Items at Generation Time}
\citet{yamaguchi2026howcan} argue that inference speed-ups can theoretically be higher in expanded models. Expanded models do not achieve this theoretical limit because newly added tokens are underutilized at generation time. To analyze whether this phenomenon occurs with our models as well, we perform the following experiment: we take the generated sequences from our experiments in \S\ref{sec:inference_speedup} with the \swpatchscopes model and re-tokenize the generation using the \swpatchscopes tokenizer and the original tokenizer. We denote the number of tokens used at generation as $G$, the number of tokens resulting from re-tokenization using the \swpatchscopes tokenizer as $E$, and the number of tokens from the re-tokenization using the original model tokenizer as $O$. The empirical token reduction at generation time can be found by comparing $G$ with respect to $O$, while the theoretical limit on token reduction can be computed by comparing $E$ with respect to $O$. We report both these numbers for the non-Latin-script languages in Table~\ref{tab:under_utilization}.

We observe that expanded models utilize less than half of the theoretical limit for most languages, echoing \citet{yamaguchi2026howcan}'s findings. Specifically, the utilization of newly added tokens is much lower for languages like Persian, Serbian, and Amharic. This result indicates that latency in the expanded models has the potential to be improved further.

\begin{table}[t]
\centering
\begin{tabular}{lccccccc}
\toprule
\textbf{Metric} & \texttt{pes} & \texttt{hin} & \texttt{srp} & \texttt{amh} & \texttt{guj} & \texttt{ory} & \texttt{mya} \\
\midrule
Empirical Reduction  (\%) & \cellcolor{gray!30}1.5  & \cellcolor{gray!30}12.4 & \cellcolor{gray!30}3.1  & \cellcolor{gray!30}1.7  & \cellcolor{gray!30}13.3 & \cellcolor{gray!30}12.0 & \cellcolor{gray!30}17.5 \\
Theoretical Max. Reduction (\%) & \cellcolor{gray!30}19.2 & \cellcolor{gray!30}37.6 & \cellcolor{gray!30}10.5 & \cellcolor{gray!30}36.7 & \cellcolor{gray!30}39.7 & \cellcolor{gray!30}47.1 & \cellcolor{gray!30}27.8 \\
\bottomrule
\end{tabular}
\caption{Observed empirical token reductions versus maximum token reduction theoretically possible. We observe that new vocabulary items are underutilized at generation time for several languages like Persian, Serbian and Amharic~(Original Model: Qwen3-30B-A3B, Expanded Model: \swpatchscopes).}
\label{tab:under_utilization}
\end{table}
\section{Additional Analysis}

\subsection{Detailed \swpatchscopes Results and Comparisons to \patchscopes}
Detailed analysis for the design choices for \swpatchscopes, and comparison with the original and the corresponding \patchscopes model in shown in \cref{tab:analysis_sw_full}. 
\begin{table}[t]
\centering

\begin{tabular}{@{}llrrrrrrrr@{}}
\toprule
\multirow{2}{*}{\textbf{Benchmark}} & \multirow{2}{*}{\textbf{Lang.}} & \multicolumn{1}{c}{\multirow{2}{*}{\textbf{Orig.}}} & \multicolumn{1}{c}{\multirow{2}{*}{\textbf{T2W}}} & \multicolumn{6}{c}{\textbf{\swpatchscopes}}  \\ \cmidrule(l){5-10} 
                                    &                                    & \multicolumn{1}{c}{}                                   & \multicolumn{1}{c}{}                                      & \multicolumn{1}{c}{\textbf{R$_m$}} & \multicolumn{1}{c}{\textbf{R$_e$}} & \multicolumn{1}{c}{\textbf{R$_m$,T}} & \multicolumn{1}{c}{\textbf{R$_e$,T}} & \multicolumn{1}{c}{\textbf{R$_m$,T,P}} & \multicolumn{1}{c}{\textbf{R$_e$,T,P}} \\ \midrule
% \multirow{14}{*}{SIB200}            & eng                                & 84.69                                                  & 84.58                                                     & 84.58                              & 84.58                             & 84.58                            & 84.58                             & 84.58                              \\
%                           &                                    & Original& Patchscopes    & R_m                   & R_e                  & R_m, T              & R_e, T               & R_m, T, P            & R_e, T, P
 \multirow{13}{*}{SIB200}   & \texttt{hun}                       & 85.8    & 85.9           & {\ul\textbf{86.9}}    & 86.7                 & 86.4                & 86.1                 & 86.4                 & 86.1   \\
                            & \texttt{vie}                       & 84.9    & \textbf{84.1}  & {\ul 83.9}            & 83.7                 & 83.7                & 83.7                 & 83.7                 & 83.7   \\
                            & \cellcolor{gray!30}\texttt{pes}     & \cellcolor{gray!30}85.4    & \cellcolor{gray!30}84.0           & \cellcolor{gray!30}84.2                  & \cellcolor{gray!30}{\ul \textbf{84.3}}  & \cellcolor{gray!30}83.3                & \cellcolor{gray!30}83.5                 & \cellcolor{gray!30}82.9                 & \cellcolor{gray!30}83.5   \\
                            & \cellcolor{gray!30}\texttt{hin}     & \cellcolor{gray!30}85.3    & \cellcolor{gray!30}81.8           & \cellcolor{gray!30}78.3                  & \cellcolor{gray!30}79.0                 & \cellcolor{gray!30}81.7                & \cellcolor{gray!30}{\ul\textbf{82.1}}   & \cellcolor{gray!30}81.7                 & \cellcolor{gray!30}81.5   \\
                            & \cellcolor{gray!30}\texttt{srp}     & \cellcolor{gray!30}86.6    & \cellcolor{gray!30}85.4           & \cellcolor{gray!30}85.1                  & \cellcolor{gray!30}{\ul \textbf{85.8}}  & \cellcolor{gray!30}85.0                & \cellcolor{gray!30}85.3                 & \cellcolor{gray!30}84.6                 & \cellcolor{gray!30}85.2   \\
                            & \texttt{gla}                       & 72.1    & \textbf{66.8}  & 62.9                  & 63.2                 & 61.3                & 61.9                 & {\ul63.4}            & 62.0   \\
                            & \texttt{mri}                       & 69.7    & \textbf{61.9}  & 60.3                  & 59.2                 & 59.3                & {\ul 61.6}           & 59.2                 & 61.0   \\
                            & \texttt{tpi}                       & 81.1    & 78.4           & 78.0                  & 77.9                 & 78.4                & 77.9                 & 78.8      & {\ul \textbf{79.2}}\\
                            & \texttt{sot}                       & 59.5    & \textbf{59.7}  & {\ul57.5}             & 57.0                 & 57.0                & {\ul57.5}            & 57.0                 & {\ul57.5}   \\
                            & \cellcolor{gray!30}\texttt{amh}     & \cellcolor{gray!30}71.2    & \cellcolor{gray!30}\textbf{70.6}  & \cellcolor{gray!30}62.0                  & \cellcolor{gray!30}62.9                 & \cellcolor{gray!30}66.2                & \cellcolor{gray!30}{\ul66.4}            & \cellcolor{gray!30}65.5                 & \cellcolor{gray!30}65.8   \\
                            & \cellcolor{gray!30}\texttt{guj}     & \cellcolor{gray!30}82.8    & \cellcolor{gray!30}\textbf{84.5}  & \cellcolor{gray!30}75.4                  & \cellcolor{gray!30}77.5                 & \cellcolor{gray!30}82.0                 & \cellcolor{gray!30}83.1                & \cellcolor{gray!30}82.4                 & \cellcolor{gray!30}{\ul83.9}   \\
                            & \cellcolor{gray!30}\texttt{ory}     & \cellcolor{gray!30}86.0    & \cellcolor{gray!30}\textbf{82.3}  & \cellcolor{gray!30}63.8                  & \cellcolor{gray!30}71.2                 & \cellcolor{gray!30}78.1                & \cellcolor{gray!30}78.6                 & \cellcolor{gray!30}{\ul 78.9}           & \cellcolor{gray!30}78.5   \\
                            & \cellcolor{gray!30}\texttt{mya}     & \cellcolor{gray!30}81.9    & \cellcolor{gray!30}\textbf{82.6}  & \cellcolor{gray!30}76.8                  & \cellcolor{gray!30}79.1                 & \cellcolor{gray!30}78.7                & \cellcolor{gray!30}79.6                 & \cellcolor{gray!30}{\ul 79.7}           & \cellcolor{gray!30}79.6   \\ \midrule
% \multirow{12}{*}{Belebele}          & eng                                & 93.89                                                  & 93.89                                                     & 93.89                              & 93.89                             & 93.89                            & 93.89                             & 93.89                              \\
 \multirow{11}{*}{Belebele} & \texttt{hun}                       & 93.9    & \textbf{89.0}  & 87.9                  & 88.0                 & 88.0                & {\ul 88.4}           & 88.0                 & {\ul88.4}\\
                            & \texttt{vie}                       & 90.6    & 89.7           & {\ul \textbf{89.9}}   & {\ul \textbf{89.9}}  & {\ul \textbf{89.9}} & {\ul \textbf{89.9}}  & {\ul \textbf{89.9}}  & {\ul \textbf{89.9}}     \\
                            & \cellcolor{gray!30}\texttt{pes}     & \cellcolor{gray!30}87.7    & \cellcolor{gray!30}\textbf{77.9}  & \cellcolor{gray!30}72.8                  & \cellcolor{gray!30}73.4                 & \cellcolor{gray!30}{\ul 76.9}          & \cellcolor{gray!30}{\ul 76.9}           & \cellcolor{gray!30}76.6                 & \cellcolor{gray!30}76.5   \\
                            & \cellcolor{gray!30}\texttt{hin}     & \cellcolor{gray!30}80.9    & \cellcolor{gray!30}\textbf{72.9}  & \cellcolor{gray!30}62.4                  & \cellcolor{gray!30}63.3                 & \cellcolor{gray!30}70.2                & \cellcolor{gray!30}{\ul 70.8}           & \cellcolor{gray!30}70.4           & \cellcolor{gray!30}{\ul 70.8}   \\
                            & \cellcolor{gray!30}\texttt{srp}     & \cellcolor{gray!30}90.2    & \cellcolor{gray!30}\textbf{88.7}  & \cellcolor{gray!30}86.0                  & \cellcolor{gray!30}86.1                 & \cellcolor{gray!30}88.0                & \cellcolor{gray!30}88.0                 & \cellcolor{gray!30}{\ul 88.4}  & \cellcolor{gray!30}88.3   \\
                            & \texttt{mri}                       & 43.7    & \textbf{39.0}  & {\ul 35.3}            & 34.5                 & 35.2                & 34.9                 & 35.2                 & 34.6   \\
                            & \texttt{sot}                       & 34.5    & 32.7           & 32.5                  & 32.9                 & 33.5                & {\ul \textbf{33.6}}  & 33.5         &{\ul\textbf{33.6}} \\
                            & \cellcolor{gray!30}\texttt{amh}     & \cellcolor{gray!30}53.7    & \cellcolor{gray!30}\textbf{48.6}  & \cellcolor{gray!30}44.8                  & \cellcolor{gray!30}44.7                 & \cellcolor{gray!30}47.3                & \cellcolor{gray!30}47.1                 & \cellcolor{gray!30}{\ul47.7}            & \cellcolor{gray!30}47.6   \\
                            & \cellcolor{gray!30}\texttt{guj}     & \cellcolor{gray!30}79.6    & \cellcolor{gray!30}\textbf{75.0}  & \cellcolor{gray!30}50.1                  & \cellcolor{gray!30}55.0                 & \cellcolor{gray!30}65.4                & \cellcolor{gray!30}68.2                 & \cellcolor{gray!30}67.0                 & \cellcolor{gray!30}{\ul 68.9}  \\
                            & \cellcolor{gray!30}\texttt{ory}     & \cellcolor{gray!30}81.9    & \cellcolor{gray!30}\textbf{73.0}  & \cellcolor{gray!30}44.7                  & \cellcolor{gray!30}49.0                 & \cellcolor{gray!30}58.9                & \cellcolor{gray!30}62.5           & \cellcolor{gray!30}62.0                      & \cellcolor{gray!30}{\ul 63.0} \\
                            & \cellcolor{gray!30}\texttt{mya}     & \cellcolor{gray!30}76.0    & \cellcolor{gray!30}\textbf{72.8}  & \cellcolor{gray!30}50.2                  & \cellcolor{gray!30}54.4                 & \cellcolor{gray!30}61.2                & \cellcolor{gray!30}62.1                 & \cellcolor{gray!30}{\ul 62.2}           & \cellcolor{gray!30}61.6  \\\bottomrule
\end{tabular}
\caption{\swpatchscopes ~results with different design choices. We present comparisons of the original model's performance with that of the expanded and initialized model using \patchscopes. }
\label{tab:analysis_sw_full}
\end{table}

\subsection{Model Scale Analysis: SIB200}
\label{sec:model_Scale_sib200_appx}
\begin{figure}
    \centering
    \begin{subfigure}{0.49\textwidth}
        \centering
        \includegraphics[width=\textwidth]{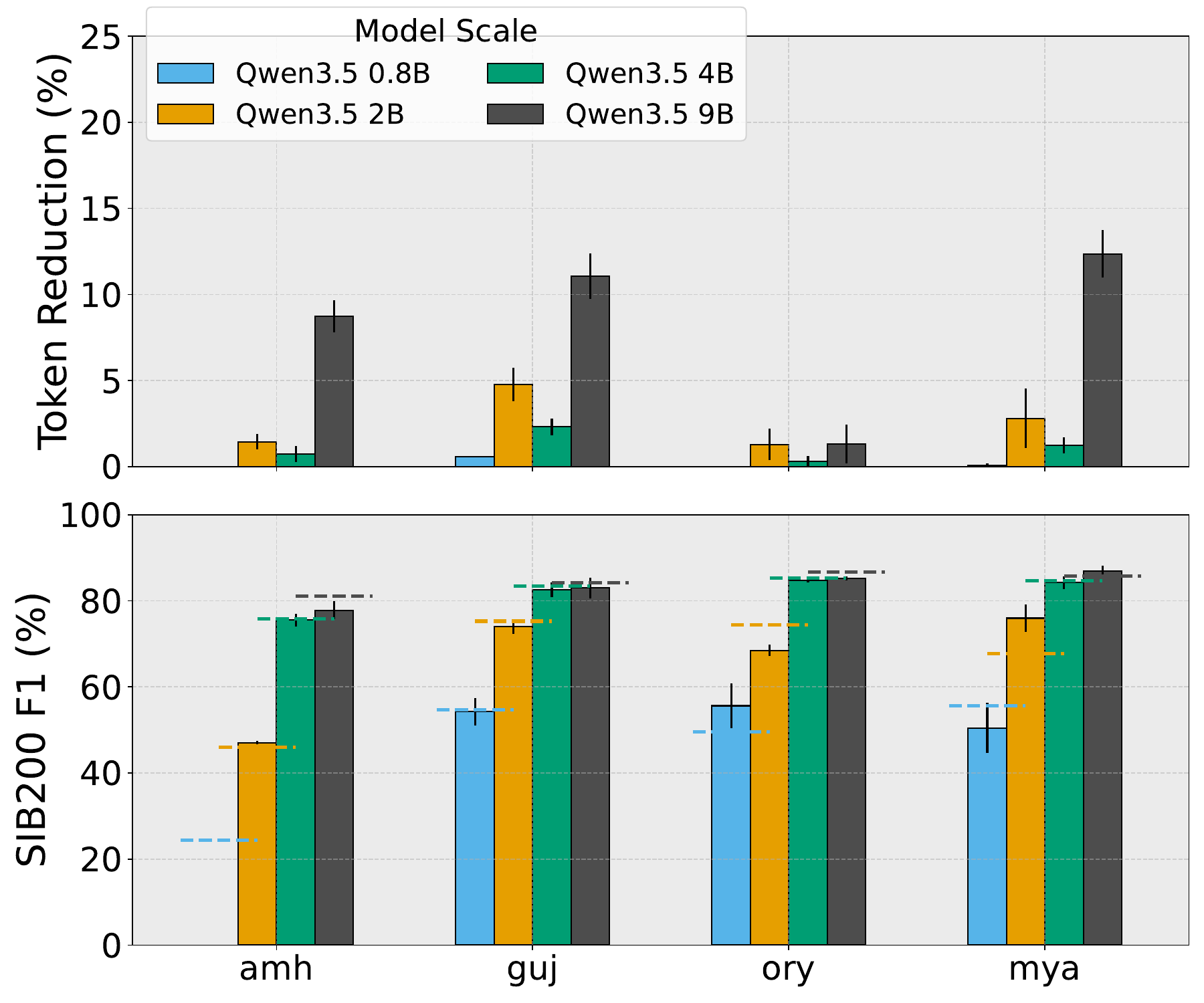}
        \caption{}
        \label{fig:model_scale_exp_sib200_psc}
    \end{subfigure}%
    ~
    \begin{subfigure}{0.49\textwidth}
        \centering
        \includegraphics[width=\textwidth]{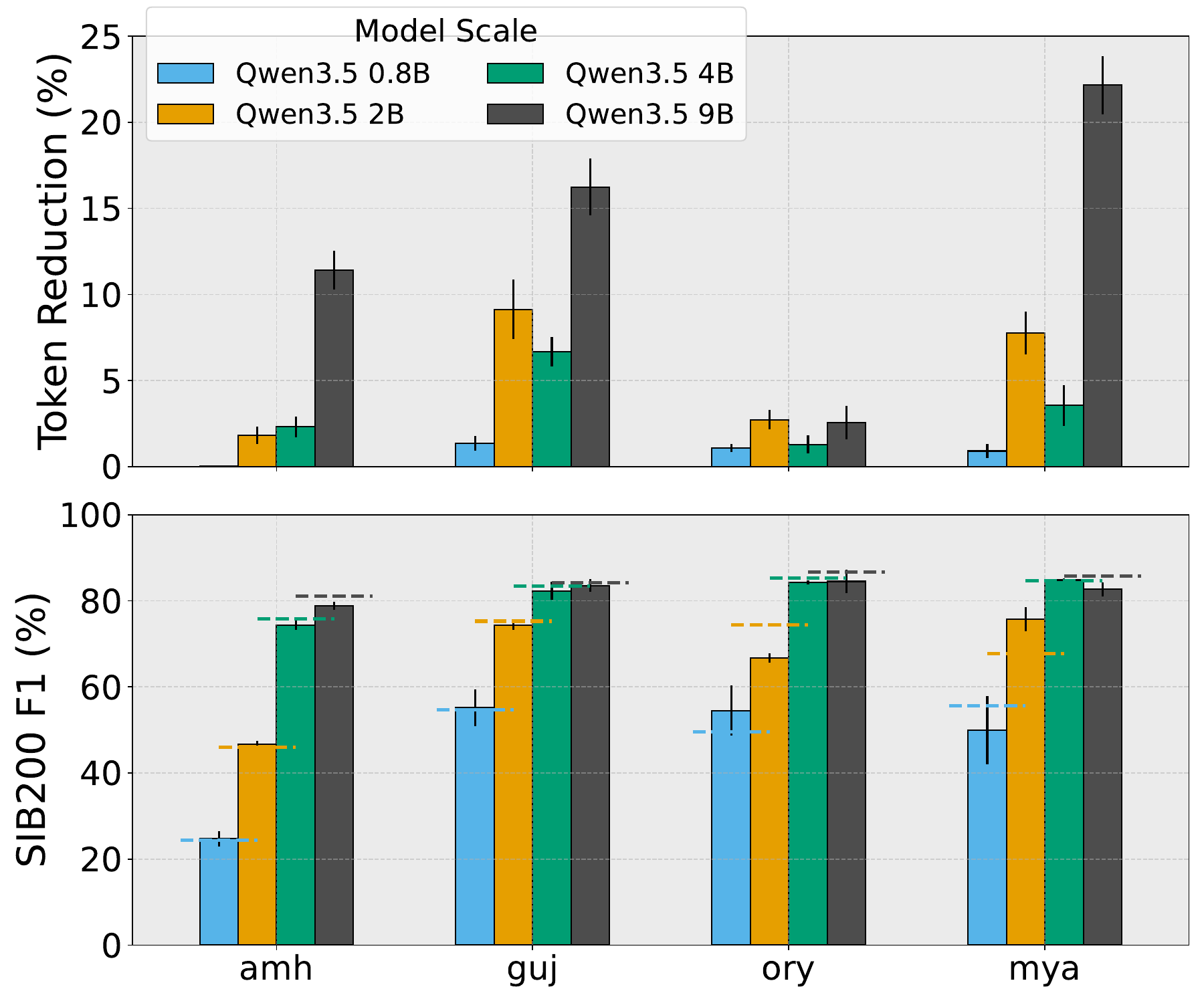}
        \caption{}
        \label{fig:model_scale_exp_sib200_swpsc}
    \end{subfigure}
    \caption{Token Reduction~(top) and corresponding SIB200 performance~(bottom) with the (a) \patchscopes and (b) \swpatchscopes method. All four languages are low-resource languages written in a non-Latin script. $|C_{train}|=$1k. }
    \label{fig:model_scale_exp_sib200}
\end{figure}

Figure \ref{fig:model_scale_exp_sib200} show token reduction and SIB200 performance across model sizes for \patchscopes and \swpatchscopes. We observe that token reduction generally improves with scale. Also, \swpatchscopes affords much higher reduction as compared to \patchscopes while showing performance robustness.  

\subsection{Model Scale Analysis: Belebele}
\begin{figure}
    \centering
    \begin{subfigure}{0.49\textwidth}
        \centering
        \includegraphics[width=\textwidth]{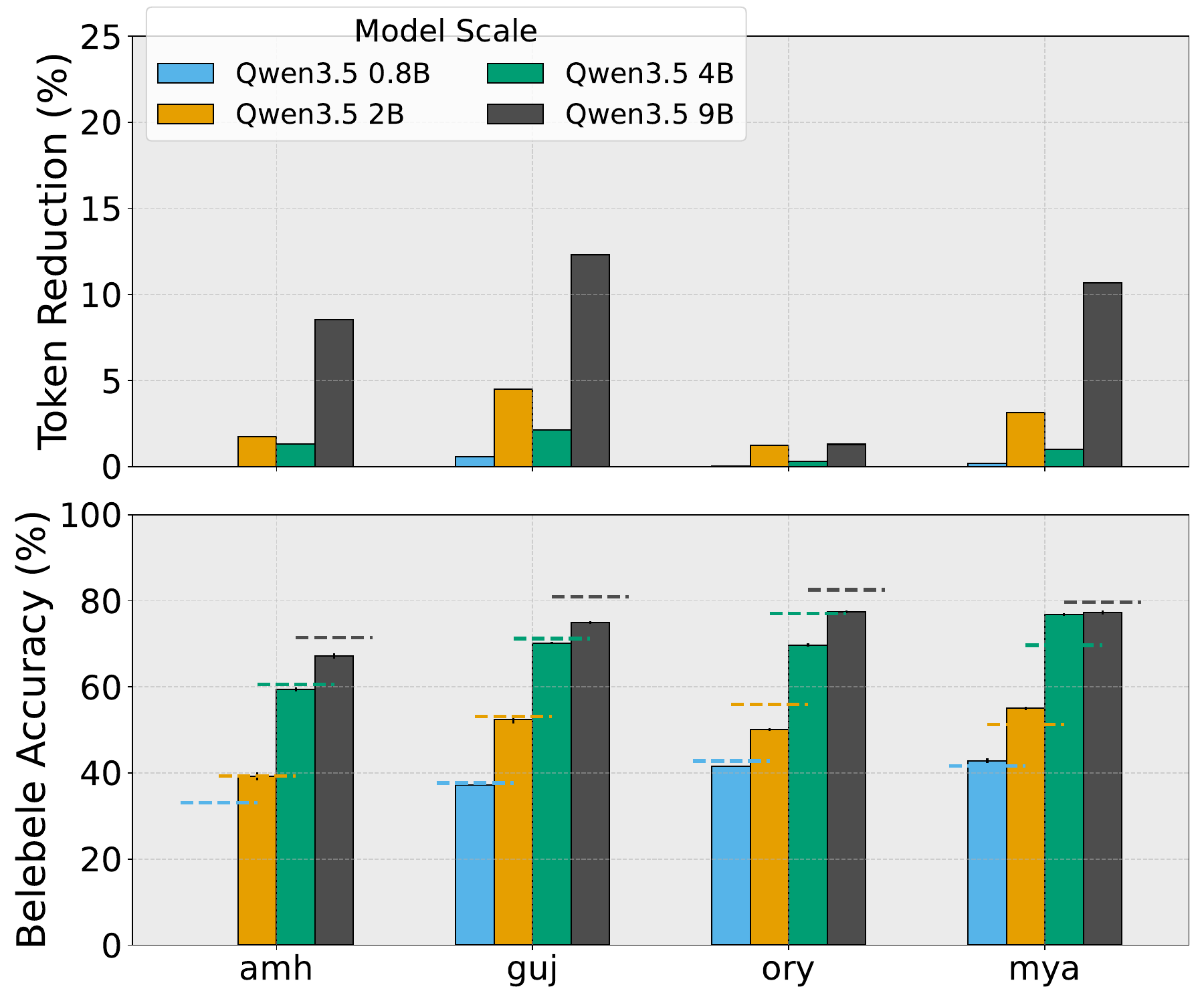}
        \caption{}
        \label{fig:model_scale_exp_belebele_psc}
    \end{subfigure}%
    ~
    \begin{subfigure}{0.49\textwidth}
        \centering
        \includegraphics[width=\textwidth]{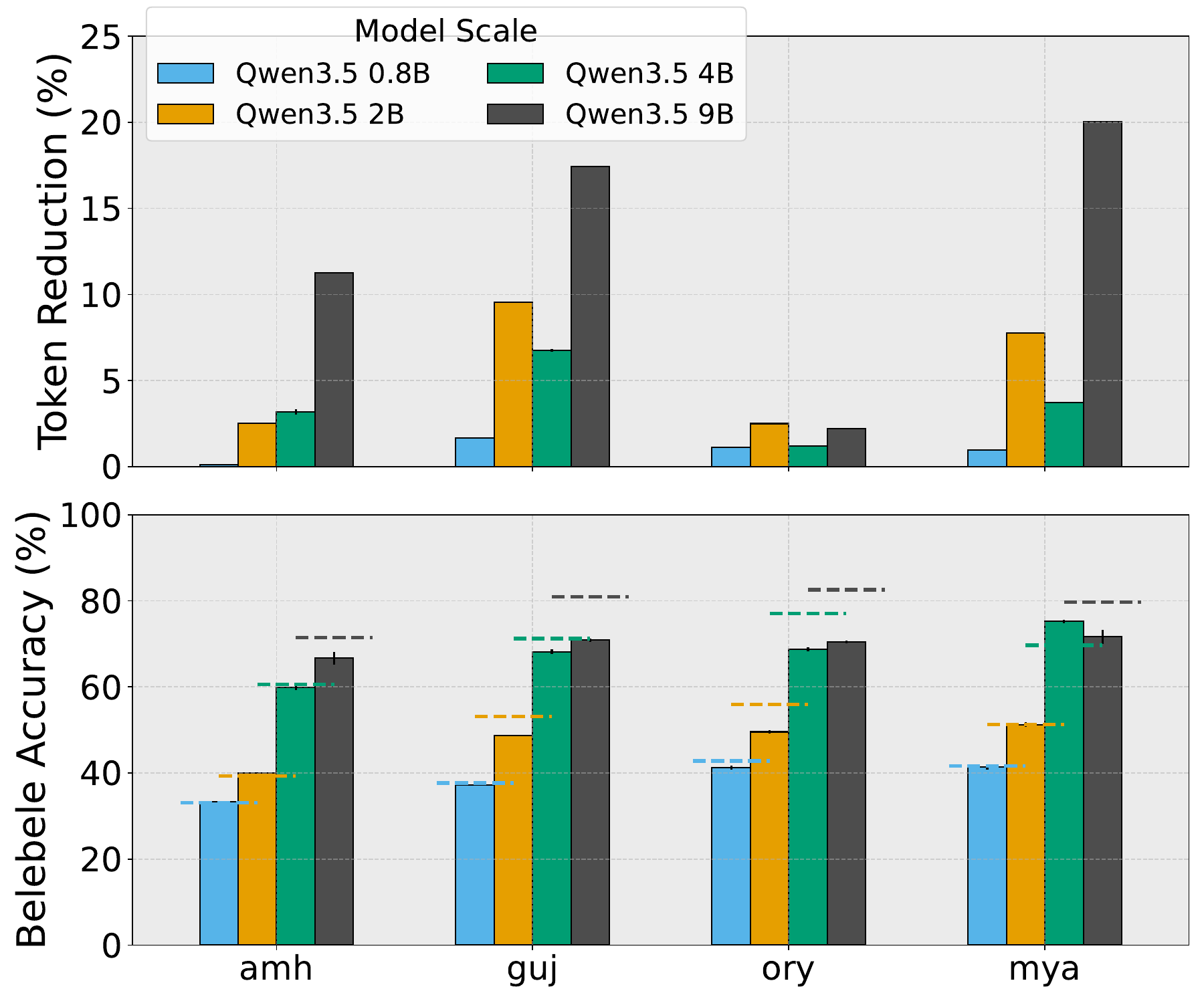}
        \caption{}
        \label{fig:model_scale_exp_belebele_swpsc}
    \end{subfigure}
    \caption{Token Reduction~(top) and corresponding Belebele performance~(bottom) with the (a) \patchscopes and (b) \swpatchscopes method. All four languages are low-resource languages written in a non-Latin script.$|C_{train}|=$1k.}
    \label{fig:model_scale_exp_belebele}
\end{figure}

Figure \ref{fig:model_scale_exp_belebele} shows token reduction and Belebele performance across model sizes for \patchscopes and \swpatchscopes. All results follow a similar trend as mentioned in \cref{sec:model_Scale_sib200_appx}.

\subsection{Token Reduction Dip for Qwen3.5-4B}

In \S\ref{sec:model_scale_analysis_sec} and \S\ref{sec:model_Scale_sib200_appx}, we compared token reduction rates across model sizes and found a dip in reduction for Qwen3.5-4B. We hypothesize that this dip can be attributed to the propensity of the 4B model to generate thinking tokens, as the thinking fine-tuning phase was only introduced in the 4B model~(and is not present in the 2B and 0.8B variants). We believe that thinking tokens can impede detokenization and hence lead to less meaningful vocabulary additions. To pursue this hypothesis further, we qualitatively analyzed the detokenization traces for Gujarati and Odia.

\paragraph{Setup.} We sampled 20 words that were successfully detokenized by the 2B model but not by the 4B model~(let’s denote it as `\textit{Set A}'). Further, we sample 20 words that were successfully detokenized by the 4B model but not by the 2B model~(denoted as `\textit{Set B}'). For both sets, we study their detokenization traces.

\paragraph{What proportion of words contain thinking tokens in their detokenization traces?} First, we report the proportion of words for which the detokenization traces contain thinking tokens. We categorize explicit \texttt{<think>} blocks and instructions not explicitly mentioned in the input (e.g., ‘answer in ten words’) as ‘\textit{thinking tokens}’. Note that since instructions can be free-form text and not necessarily formulaic, we perform a manual analysis. For the 4B model, we find that nearly 50$\%$ of the words in Gujarati (across both \textit{Sets A and B)} have detokenization traces that contain thinking tokens at some layer. Similarly, 30$\%$ of the words in Odia have traces that contain thinking tokens at some layer (across both sets). However, for the 2B model, none of the detokenization traces across both sets and languages contain thinking tokens. This result provides preliminary evidence for the presence of thinking tokens in detokenization traces corresponding to the 4B model.

\paragraph{How do detokenization traces differ between the 2B and 4B models?} Next, we manually inspect the differences in the detokenization traces between the two models. We find that the 2B model is far more compliant with the repetition structure (as shown in Prompt Template D.1) than the 4B model. The 4B model's detokenization traces often contain random free text, in addition to thinking tokens. To quantify this hypothesis, we annotated detokenization traces across all layers for the words in Sets \textit{A }and \textit{B}. We find that nearly 90$\%$ of traces corresponding to the 2B model across the two languages contain a repetition or repetition-structure-compliant output. This figure nears only 65$\%$ for the 4B model, showing a higher presence of free-form text in the traces. We argue that this observation may be due to two reasons. First, the 4B model is worse at following instructions and hence cannot adhere to the input prompt. Second, the model’s reliance on test-time reasoning increases the chances of generating free-form text before restoring the repetition structure. However, the first hypothesis directly contradicts our downstream performance results in Figure~\ref{fig:model_scale_exp_sib200}, making the second hypothesis the likelier cause.

\paragraph{Are there any differences between the traces pertaining to Set A and Set B for the 4B model?} We do not find significant differences in either study that distinguish set A and set B traces. The only exception is that, in Odia, 62$\%$ of the Set A traces contain a repetition or a repetition-structure-compliant output. However, the number jumps to 72$\%$ when we measure for Set B. This result suggests that a higher rate of repetition in traces might improve the chances of successful detokenization. However, given this single point of evidence, we refrain from drawing any concrete conclusions.

While this analysis is not definitive proof of the hypothesis, the study does provide useful signals. We leave a conclusive study for future work.
\section{Limitations and Future Directions}
In this work, we carefully analyze the merits of interpretability-based approaches for vocabulary expansion, first by comparing the existing \patchscopes method against conventional approaches, and then by addressing its limitations through the proposal of \swpatchscopes. Note that \swpatchscopes and ``\textit{subword detokenization}'' present one prospective direction of improvement under this agenda. Consequently, in this section, we structure the discussion around further improving detokenization-based methods. 

Detokenization-based approaches, such as the ones discussed in this work, can be improved in the following four stages: 
\begin{enumerate}
    \item \textbf{Selecting Better Candidate Items.} With \swpatchscopes, we explore this direction by increasing the number of candidate items for vocabulary expansion by considering subwords, in addition to the full-word approach taken by \patchscopes. Here, `\textit{better}' refers to the number of candidate items. While our approach considers the theoretical maximum candidate set possible for white-space separated words, this approach requires detokenization checks at every subword token of the candidate word at every model layer. This approach works for the relatively small corpus sizes considered in this work. However, this approach may become slow as the corpus size scales up. Future work can aim to identify specific subword units \textit{a priori}~(e.g., valid morphological units) that may have a higher chance of detokenization. In our work, we considered subwords of white-space separated tokens. An alternate approach could consider candidate `\textit{superwords}'~\cite{liu2025superbpe} spanning across word boundaries, which could further boost the number of candidate items.   
    \item \textbf{Improving Detokenization Success.} The second direction of improvement pertains to improving the detokenization success rate. In \S\ref{sec:detok_successrate_psc}, we observe that at least a third of the candidate items in low-resource languages are not successfully detokenized with \patchscopes, with the problem being far more acute in smaller models. \textit{Can we improve the detokenization success rate?} We see two directions of improvement:
    \begin{inparaenum}[a)]
        \item identifying a better approach for detecting detokenization. In the current approach, both methods use Patchscopes. However, one can use alternative approaches~(e.g., LatentQA techniques such as Activation Oracles~\citep{karvonen2025activation}), or
        \item using auxiliary signals, such as bilingual counterparts in higher-resource languages or transliterations, as stand-ins for detokenization. 
    \end{inparaenum}
      
    \item \textbf{Performing Better Transformation.} Both methods learn a layer-wise transformation between layer activations and embeddings over the original model vocabulary. Once detokenization is successful, both methods use the appropriate transformation to produce initialization embeddings for the detokenized item. However, \textit{what is the best activation~(or combination thereof) that should be fed to this transformation?} In \S\ref{sec:swpsc_design_choices}, we answer this question in part. We show that transforming the activation at the earliest layer at which detokenization is successful produces better results than averaging all activations with detokenization success. However, one could adopt a more nuanced approach and use model confidence-based heuristics to determine the best activation for transformation. Another aspect to consider is the learned transformation itself. In our approach, we learn for all activations across all in-vocabulary items. Yet, one could filter out weaker activations used for this training~(perhaps by using detokenization again on the single-token item). Finally, the method used to learn the transformation merits further exploration.   
    
    \item \textbf{Enhancing Tuning Strategies.} This point corresponds to \texttt{Stage 3} of the vocabulary expansion process, as described in \S\ref{sec:intro}. In this work, we do not explore this stage, and keep it fixed across all our experiments. Prior work~\citep{kim2024efficienteffectivevocabularyexpansion,yamaguchi2026howcan} has studied this stage in the context of vocabulary expansion with conventional methods. However, this stage could benefit from further exploration in the context of interpretability-based methods. 
\end{enumerate}

The discussion above pertains to detokenization-based approaches. However, interpretability-based vocabulary expansion is a broader research topic that encompasses various other methods. \citet{dobler2026token}'s token distillation is one such approach. Several other lines of research, potentially using other interpretability-based resources such as Sparse Autoencoders~\citep{huben2024sparse} and Cross-Layer Transcoders~\citep{ameisen2025circuit}, are promising directions for future vocabulary-expansion methods.  

\end{document}